\documentclass[journal]{IEEEtran}
\usepackage[T1]{fontenc}
\usepackage{graphicx}
\usepackage{amsfonts}
\usepackage{amsthm}
\usepackage{amsmath}
\usepackage{amssymb}
\usepackage{algpseudocode}
\usepackage{algorithm}
\usepackage{url}
\usepackage{array}
\usepackage{subfigure} 
\usepackage{mathtools}
\usepackage{mathtools}
\usepackage{bm}
\usepackage{tabularx}
\usepackage{epstopdf}
\usepackage{color} 
\usepackage[numbers,sort&compress]{natbib}
\usepackage{multirow,booktabs}
\usepackage{alltt}
\usepackage{bigstrut}
\usepackage[T1]{fontenc}
\usepackage[caption=false]{subfig}
\usepackage{lipsum}

\newcommand\blfootnote[1]{%
	\begingroup
	\renewcommand\thefootnote{}\footnote{#1}%
	\addtocounter{footnote}{-1}%
	\endgroup
}
\def \l {{\boldsymbol{\ell}}}
\def \r {{\mathbf{r}}}
\def \k {{\mathbf{k}}}
\def \s {{\textbf{s}}}
\def \p {\textbf{\emph{p}}}
\def \res {\textbf{\emph{res}}}
\def \res {\upsilon}
\def \f {\boldsymbol{f}}
\def \fb {\mathbf{f}}
\def \g {\boldsymbol{g}}
\def \q {\textbf{q}}
\def \Z {\mathbf{Z}}

\def \L {{\mathcal{L}}}
\def \z {{\mathbf{z}}}
\def \e {{\mathbf{e}}}
\def \U {{\mathcal{U}}}
\def \D {{\mathcal{D}}}
\def \V {{\mathcal{V}}}
\def \db {{\boldsymbol{\delta}}}

\def \Z {{\mathcal{Z}}}
\def \E {{\mathcal{E}}}
\def \W {{\mathbf{W}}}
\def \mub {{\boldsymbol{\mu}}}
\def \K {{\mathbf{K}}}
\def \Rn {{\mathbb{R}}}
\def \Sig {{\boldsymbol{\Sigma}}}
\def \O {{\mathcal{O}}}
\def \x {{\boldsymbol{x}}}
\makeatletter
\newcommand*\bigcdot{\mathpalette\bigcdot@{1}}
\newcommand*\bigcdot@[2]{\mathbin{\vcenter{\hbox{\scalebox{#2}{$\m@th#1\bullet$}}}}}

\makeatother

\begin{document}
	\title{A Generalized Framework for Autonomous Calibration of Wheeled Mobile Robots}
	\author{\IEEEauthorblockN{Mohan Krishna Nutalapati,~\IEEEmembership{Student Member,~IEEE},
			Lavish Arora,~\IEEEmembership{Student Member,~IEEE},
			Anway Bose,\newline
			Ketan Rajawat,~\IEEEmembership{Member,~IEEE} and 
			Rajesh M Hegde,~\IEEEmembership{Senior Member,~IEEE} }
	\thanks{\IEEEauthorblockA{The authors are with the Department of Electrical
				Engineering, Indian Institute of Technology Kanpur, Kanpur 208016, India,
				(e-mail: $\left\{\text{nmohank, lavi, anwayb, ketan, rhegde} \right\}$@iitk.ac.in).}}}
	\IEEEtitleabstractindextext{
		\begin{abstract}
			Robotic calibration allows for the fusion of data from multiple sensors such as odometers, cameras etc., by providing appropriate transformational relationships between the corresponding reference frames. For wheeled robots equipped with exteroceptive sensors, calibration entails learning the motion model of the sensor or the robot in terms of the odometric data, and must generally be performed prior to performing tasks such as simultaneous localization and mapping (SLAM). Within this context, the current trend is to carry out simultaneous calibration of odometry and sensor without the use of any additional hardware. Building upon the existing simultaneous calibration algorithms, we put forth a generalized calibration framework that can not only handle robots operating in 2D with arbitrary or unknown motion models but also handle outliers in an automated manner. We first propose an algorithm based on the alternating minimization framework applicable to two-wheel differential drive. Subsequently, for arbitrary but known drive configurations we put forth an iteratively re-weighted least squares methodology leveraging an intelligent weighing scheme. Different from the existing works, these proposed algorithms require no manual intervention and seamlessly handle outliers that arise due to both systematic and non-systematic errors. Finally, we put forward a novel Gaussian Process-based non-parametric approach for calibrating wheeled robots with arbitrary or unknown drive configurations. Detailed experiments are performed to demonstrate the accuracy, usefulness, and flexibility of the proposed algorithms. 
			\blfootnote{This paper has supplementary downloadable material (available at \url{http://tinyurl.com/simultaneous-calibration}) that includes raw-data of all the experiments and implementation codes for the proposed methodologies.}
		\end{abstract}
		}
	\maketitle
	\IEEEdisplaynontitleabstractindextext
	\begin{table}
		\centering
		\captionsetup{font=scriptsize}
		\caption{Nomenclature used in the paper}
		\label{notation}
		\begin{tabular}{@{}l@{}}
			\toprule
			Parameters that are to be estimated                                                                                                                                                                 \\ \midrule
			$\begin{matrix} \p = (\underbrace{\ell_x,\ell_y,\ell_\theta}_\l,\r) &  \text{parameters to be estimated} \end{matrix}$ \\
			$\begin{matrix} \l  &  \text{position of extrinsic sensor w.r.t robot frame} \end{matrix}$ \\
			$\begin{matrix} \r  &  \text{robot instrinsic parameters} \end{matrix}$ \\
			$\left. \begin{matrix}r_L,r_R & \hspace{-0.4cm}\text{left and right wheel radii (m)}\\\hspace{-0.5cm}b& \text{distance between two wheels (m)} \end{matrix}\right\}\text{Two-wheel drive}$ 
			\\
			$\left.\begin{matrix}L_x &\text{half of axle length along x-axis of the robot (m)}&  \\ L_y & \text{half of axle length along x-axis of the robot (m)} & \\ r&\text{radius of each wheel}& \end{matrix}\right\} \text{Mecanum drive}$ \\
			\midrule
			Measurements \\ \midrule
			$\begin{matrix}
			\mathcal{U}  & \text{raw data  log of odometry sensor} & \\
			\mathcal{V}  & \text{Measurememts from exteroceptive sensor} & \\ 
			\db(t) & \text{vector of wheel angular velocities at time instant $t$} \\
			\Z(t) & \text{exteroceptive sensor measurement at time instant $t$} \\
			\q(t)\ \   = &[q_x(t)\ q_y(t)\ q_\theta(t)]^T\ \text{Pose of robot at any time }\emph{t} \\ 
			\hat{\s}_{jk}  & \text{sensor displacement estimate for time interval}\ \emph{$[t_j,t_k)$} 
			\end{matrix}$ \\
			\midrule 
			Operators \\
			\midrule
			$\begin{matrix}
			\oplus &\text{Roto-translation operator}  & \\ 
			\circleddash &\text{inverse of} \oplus \text{operator}   & \\
			& \begin{bmatrix}
			a_x\\ 
			a_y\\ 
			a_\theta
			\end{bmatrix} \oplus \begin{bmatrix}
			b_x\\ 
			b_y\\ 
			b_\theta
			\end{bmatrix} \overset{\Delta}{=} \begin{bmatrix}
			a_x + b_x \cos a_\theta - b_y \sin a_\theta\\ 
			a_y + b_x \sin a_\theta + b_y \cos a_\theta \\ 
			a_\theta + b_\theta
			\end{bmatrix} & \\ \\ 
			& \circleddash \begin{bmatrix}
			a_x\\ 
			a_y\\ 
			a_\theta
			\end{bmatrix} \overset{\Delta}{=} \begin{bmatrix}
			-a_x \cos a_\theta - a_y \sin a_\theta\\ 
			a_x \sin a_\theta - a_y \cos a_\theta\\ 
			-a_\theta
			\end{bmatrix}
		\end{matrix}$ \\
		\midrule
		\end{tabular}
		\end{table}
		\section{Introduction}
		\IEEEPARstart{R}{obotic} calibration is an important first step necessary for carrying out various sophisticated tasks such as simultaneous localization and mapping (SLAM) \cite{SLAM1, SLAM2}, object detection and tracking \cite{objectdetec}, and autonomous navigation \cite{Autonav}. For most wheeled robot configurations that comprise encoders and exteroceptive sensors, the calibration process entails learning a mathematical model that can be used to fuse odometry and sensor data. For instance, calibrating a robot with a two-wheel differential drive robot involves learning various \emph{intrinsic} parameters, namely the wheel radii and the axle length, and the \emph{extrinsic} parameters, namely the pose of the sensors \cite{intrinsic1,intrinsic2,extrinsic2,extrinsic3}. More generally, however, when the motion model of the robot is unavailable, calibration involves learning the relationships that describe the sensor motion in terms of the odometry measurements. Precise calibration is imperative since calibration errors are often systematic and tend to accumulate over time \cite{new}. Conversely, an accurately specified odometric model complements the exteroceptive sensor, e.g. to correct for measurement distortions if any \cite{LOAM}, and continues to provide motion information even in featureless or geometrically degenerate environments \cite{degraded}. 
		
		\begin{figure} 
			\centering  
			\subfigure[\textit{Configuration} \textbf{T1}]{\includegraphics[width=0.48\linewidth,trim={0.5cm 2cm 0.3cm 0.8cm},clip]{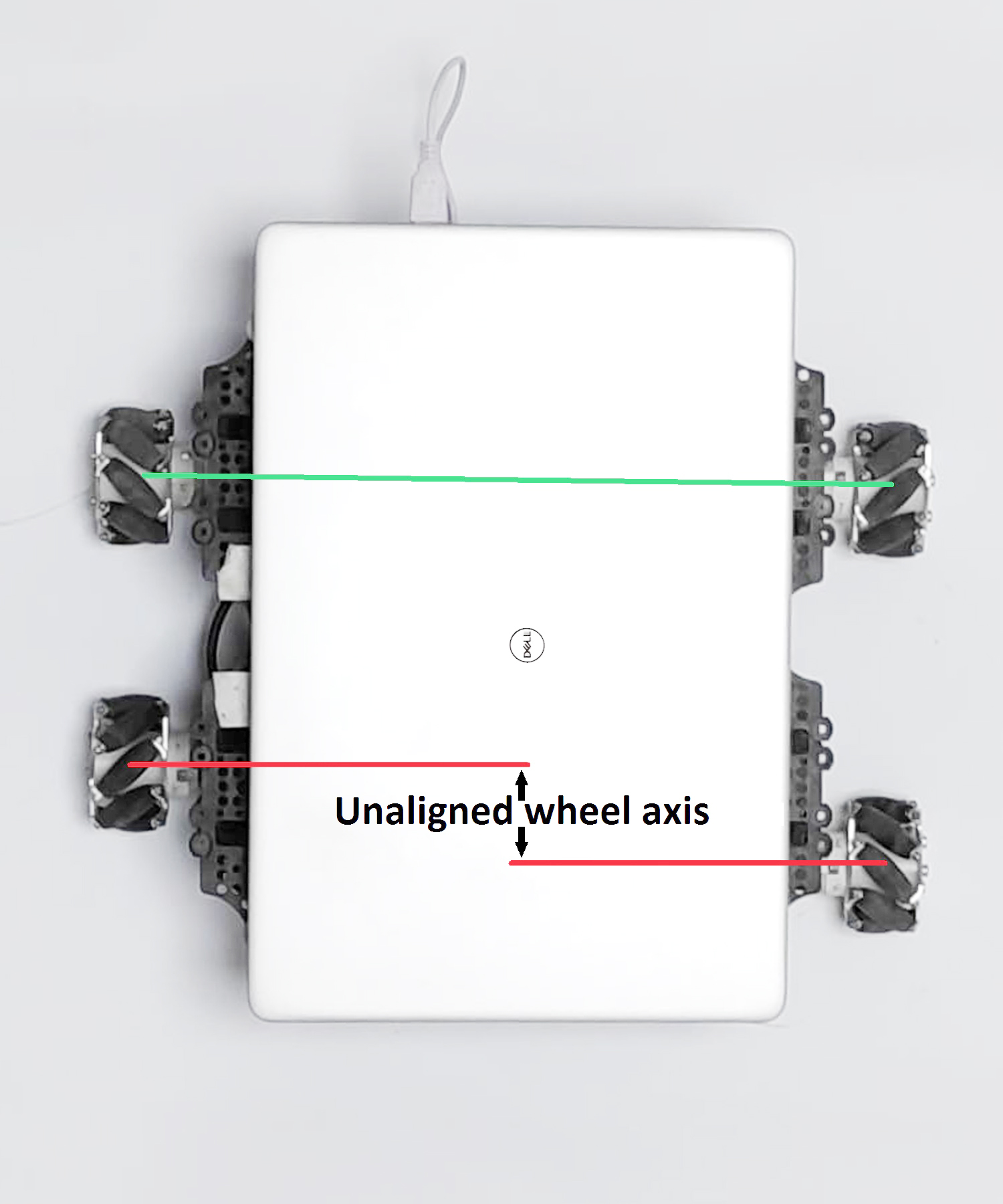}}
			\subfigure[\textit{\textit{Configuration}} \textbf{T2}]{\includegraphics[width=0.48\linewidth,trim={0.5cm 2cm 0.3cm 0.8cm},clip ]{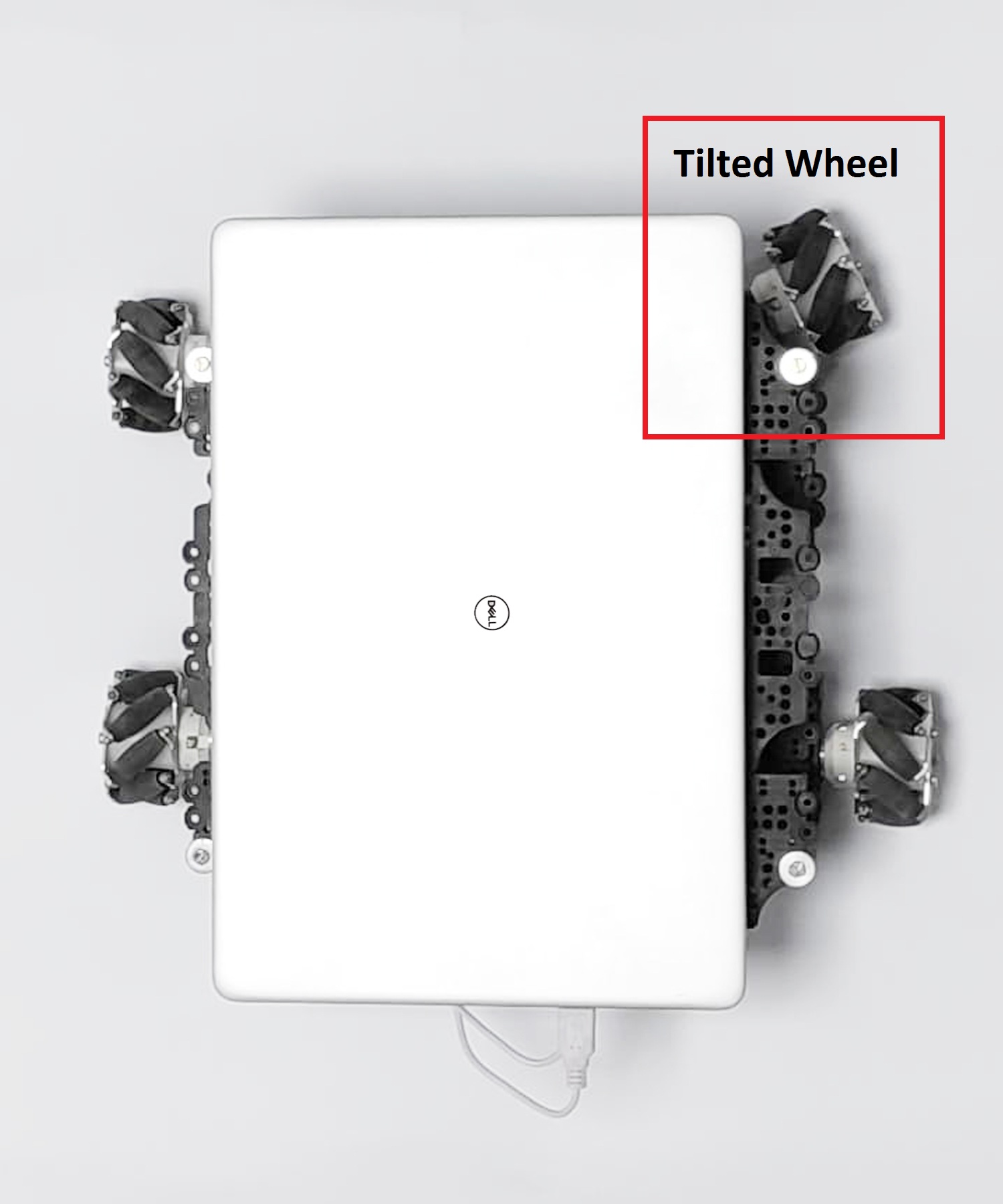}}
			\captionsetup{font=scriptsize}
			\caption{Deformed Turtlebot3 Mecanum drive robot used for experimental evaluations. (a) Unaligned wheel axis deformation, (b) Tilted wheel deformation.}\label{meca}
			\vspace{-0.6cm}
		\end{figure}
		\begin{figure} 
			\centering  
			\subfigure[\textit{\textit{Fire Bird VI} robot}]{\includegraphics[width=0.49\linewidth,trim={0cm 4cm 0cm 0cm},clip]{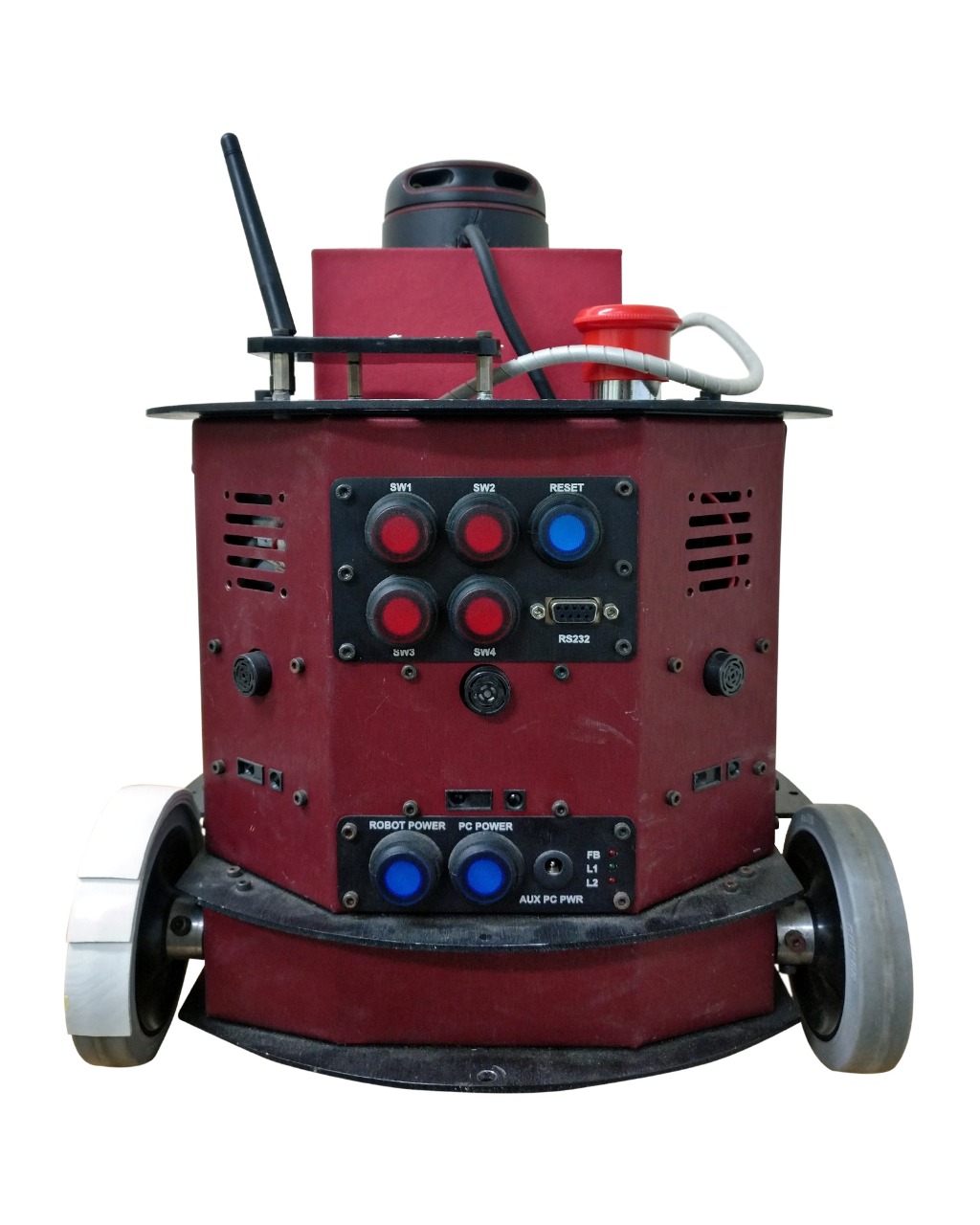}}
			\subfigure[\textit{Configuration} \textbf{F1}]{\includegraphics[width=0.40\linewidth,trim={2cm 0cm 0cm 4cm},clip ]{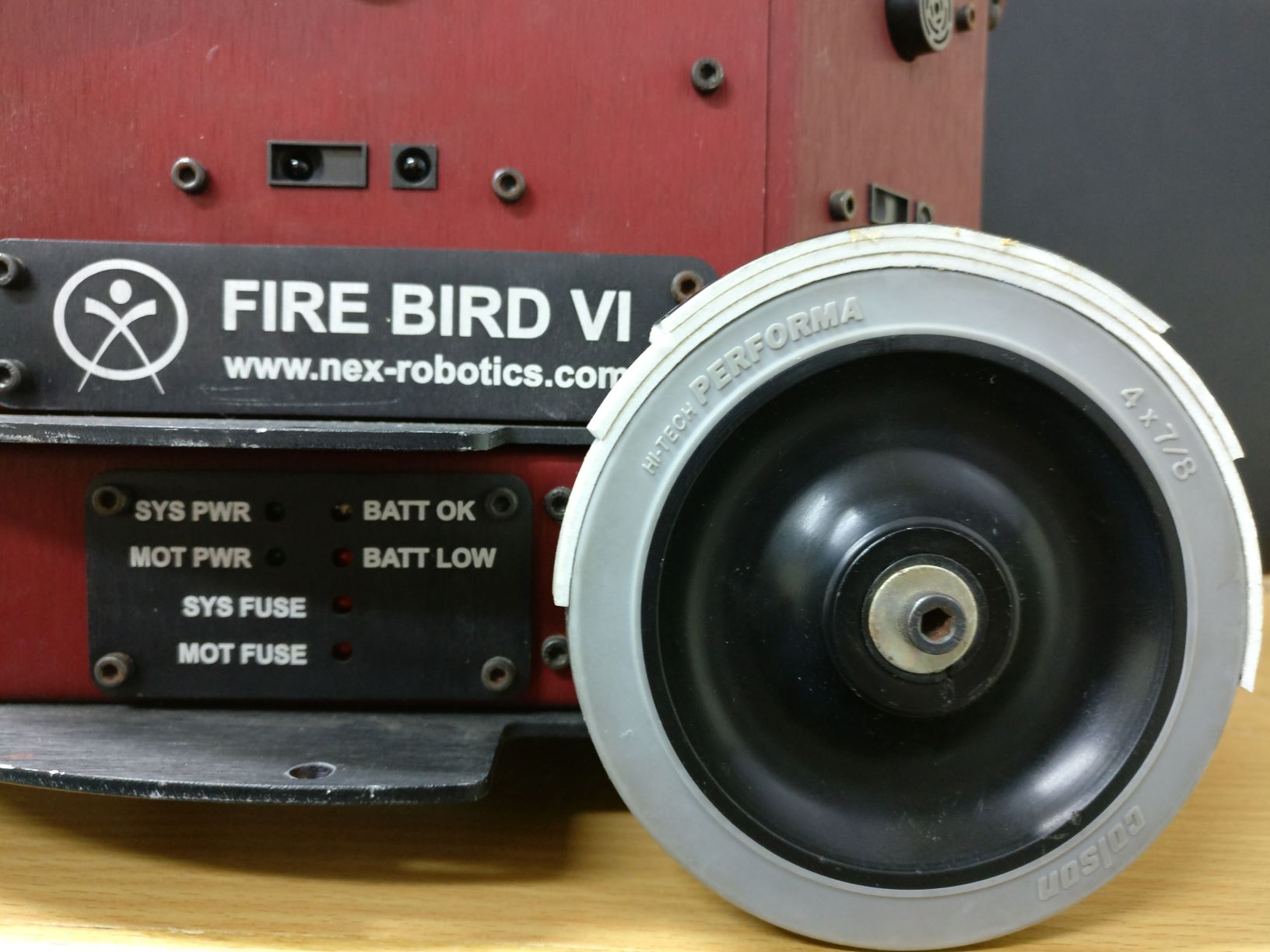}}
			\captionsetup{font=scriptsize}
			\captionsetup{justification = centering}
			\caption{One of the wheels (left) of \textit{Fire Bird VI} robot is deformed such that the wheel looses its notion of circularity.}\label{fig_def}
		\end{figure}  
		As the robot undergoes wear-and-tear during its course of operation, calibration must also be performed regularly. Such a requirement motivates the need for calibration approaches that work without any additional hardware, require no prior information, and do not need to disrupt the operation of the robot. A joint calibration approach for a two-wheel differential drive that is independent of specialized hardware was first proposed in \cite{censi08}, with the formal analysis presented in \cite{censi} and has since been extended to other settings such as the tricycle robot \cite{tricycle}. The idea here is to find a maximum likelihood estimate of the various intrinsic and extrinsic parameters using the ego-motion estimates provided by the exteroceptive sensor. As long as these ego-motion estimates are sufficiently accurate, calibration can be carried out without any additional hardware or a specialized environment. The present work also utilizes such a joint calibration approach. Fully automatic and simultaneous extrinsic-odometric calibration algorithms exist, but require artificial landmarks to properly handle outliers \cite{trad2}.

		Henceforth, existing algorithms for joint calibration suffer from two key issues (a) applicability to specific drive configurations due to the use of customized algorithms; and (b) outlier rejection mechanisms requiring either manual intervention or specialized hardware. Besides, extending the approaches in \cite{censi,tricycle} to other robot configurations may be possible but not straightforward. More importantly, existing methods do not apply to robots for which full kinematic models are not available. Examples include complex multi-wheel robots, robots with a misaligned axis, other unknown offsets or deformations (see Fig. \ref{fig_def} and Fig. \ref{meca}), and robots suffering from excessive wear-and-tear, such as a punctured wheel \cite{nomodel}. As the calibration is performed in operating environments, the ego-motion estimates provided by the sensor are often corrupted with outliers. A fully automated outlier removal mechanism is necessary to ensure that the calibration routine runs by itself.

		This work puts forth a generalized calibration framework applicable to robots with arbitrary or unknown models and is capable of rejecting outliers automatically. We begin with describing joint odometry and sensor calibration algorithm for two-wheel differential drive robot that is capable of handling outliers in an automated manner. Notably, the proposed approach handles various non-systematic errors such as those arising from sensor malfunctions and wheel slippages. The method is in contrast with existing approaches such as \cite{censi} where outlier removal requires specific environment-dependent parameters to be carefully tuned. Subsequently, an iteratively re-weighted least-squares algorithm is proposed that is capable of calibrating any robot with a known kinematic model while also automatically handling outliers. We remark here that the proposed algorithm can be used with a variety of robots with complicated drive configurations and is the first such general-purpose simultaneous calibration algorithm. Finally, a completely generic Gaussian process (GP) regression based calibration algorithm is proposed that can calibrate any robot from scratch and without knowing its kinematic model. Experiments are carried out to assess the performance of the proposed algorithms. As compared to the state-of-the-art algorithms, the first two algorithms yield better performance while also handling outliers automatically. The third algorithm is tested on robots with minor deformations to wheels and axes (see Fig. \ref{meca} and Fig. \ref{fig_def}), and it is understood that existing algorithms are not applicable to handle such deformations. In contrast, the proposed GP-based algorithm continues to yield accurate odometric motion estimates of the extrinsic sensor.
		
		The rest of the paper is organized as follows. Sec. \ref{rel} briefly reviews some related literature. Sec. \ref{probfor} details the system setup and the problem formulation. The proposed algorithms are described in Sec. \ref{sec4}. Detailed experimental evaluations are carried out to validate the performance of the proposed methods, and the results are discussed in Sec. \ref{expsec}. Finally, Sec. \ref{consec} concludes the paper. The notation used in the paper is summarized in Table \ref{notation}.

		\section{Related Work} \label{rel}
		Wheeled robots with exteroceptive sensors require both intrinsic and extrinsic calibration before the odometry data can be fused with the sensor data. Initial works on intrinsic calibration utilized specially crafted trajectories to estimate the imperfections in the kinematic model of the robot and hence correct for systematic odometry errors \cite{earlier}. Similar approaches were later proposed for generalized trajectories in \cite{extend} as well as other configurations such as car-like \cite{intrinsic3} and tricycle \cite{AGV9}. A linear identification problem to estimate odometry calibration parameters is formulated and solved within the least-squares framework in  \cite{liniden}. A common issue among these approaches was the need for external measurement systems such as calibrated video cameras or motion capture systems. Towards simplifying the calibration problem, \cite{accaliber} proposed an algorithm that exploited the redundant information from multiple sensors mounted on the robot. Likewise, filtering-based calibration techniques were proposed in \cite{censi5, censi6, censi7, censi8, AGV10,AGV11}, some of these approaches involve incorporating the systematic parameters in the state vector and estimating them within the extended Kalman filter (EKF)-based SLAM framework. However, filtering techniques such as EKF are not designed to handle outliers, that must be eliminated separately rendering the entire process suboptimal. Finally, the calibration problem has also been studied within the aegis of optimization theory, and relevant works include \cite{intrinsic4,linanal} where systematic and random errors are analyzed and modelled for vehicle odometry.

		Methods for extrinsic calibration generally assume that the intrinsic parameters of the robot are already available. The problem of determining the transformation between a camera and an IMU was considered in \cite{censi19} within the EKF framework. In \cite{20censi}, the maximum likelihood formulation using observations from a mirror surface is considered to estimate the six-degrees-of-freedom transformation between a camera and the body of the robot. Extrinsic calibration of a lidar sensor attached to a ground vehicle using EKF is considered in \cite{AGV15}. Both extrinsic and intrinsic calibration approaches may perform poorly if the underlying parameter estimation problem is ill-conditioned. Observability analysis, in general, provides valuable insights towards practical considerations under which all the intended parameters are observable. Observability properties for different combinations of sensors are described in \cite{obs1}. Later \cite{obs2,obs3} solved the extrinsic calibration problem along with formal observability analysis. Also, \cite{censi17} calibrate a bearing sensor and theoretically validate through an observability analysis, taking into account the system nonlinearities. Misalignment error if any, was explicitly modelled and corrected in the multi-sensor calibration algorithms in \cite{censi21,censi22}.

		Simultaneous calibration of the intrinsic and extrinsic parameters has been considered before, but with the help of external hardware; see, e.g., \cite{ipec}. As already discussed, the problem of simultaneous calibration without any special equipment was first considered in  
		\cite{censi08}, with the formal analysis presented in \cite{censi} for two-wheel differential drive robots and later for tricycle robots in \cite{tricycle}. More generic calibration routines for arbitrary robot configurations were presented in \cite{agv25} without any convergence guarantees.
		A joint SLAM and calibration problem was considered in \cite{SLAMC} and involved solving a non-linear optimization problem using the Gauss-Newton method. Note however that while such an algorithm allows us to track the calibration parameters closely, it incurs a significantly higher overall complexity as opposed to batch calibration methods. Most of the current approaches are not robust to outliers in the sensor measurements. However, some approaches handle outliers either through a manual trimming procedure in \cite{censi} or by installing special hardware such as reflective markers \cite{tricycle}. 
		Generalizing the existing schemes, the proposed algorithm allows calibration of arbitrary robot configurations while automatically handling outliers.

		Thus far, model-free calibration of robots has not been widely studied. A few exceptions include \cite{nomodel} which entailed learning the inverse kinematics of the robot using instance-based learning techniques. The present work is inspired by \cite{new} where GP regression is used to complement the existing model by accounting for unmodeled errors. The proposed approach is more general and allows us to directly learn the full kinematic model.

		\begin{figure}
			\centering  
			\subfigure{\includegraphics[width=0.90\linewidth]{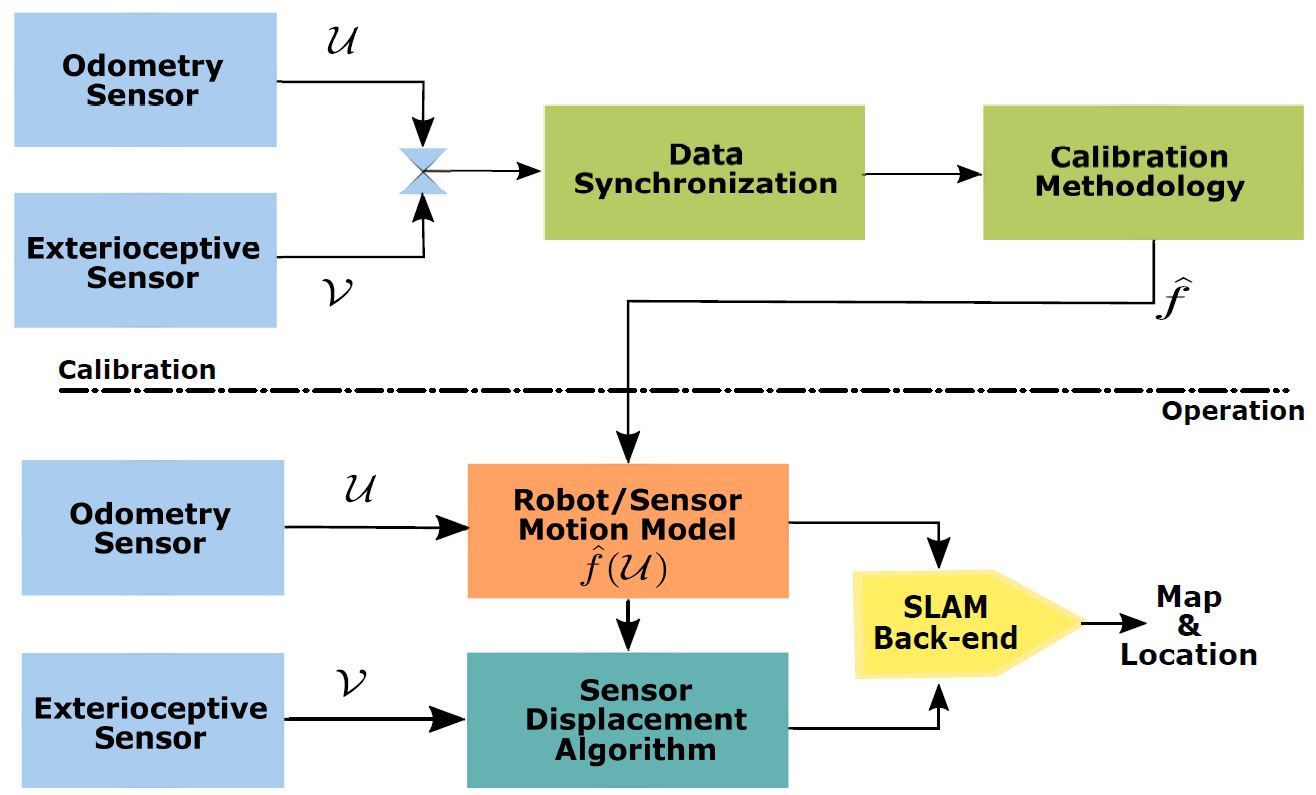}}
			\captionsetup{font=scriptsize}
			\caption{Illustrating the calibration methodology and its usage in performing high level autonomous tasks such as SLAM}\label{meth}
		\end{figure}
		
		\section{Problem Statement} \label{probfor}
		We begin with a brief description of the system model adopted here and depicted in Fig.\ref{meth}. In the calibration phase the foremost goal is to estimate the robot/sensor motion model making use of synchronized data from odometry and extrinsic sensors, where as in the operational phase the estimated model is used to perform high level autonomous tasks such as 
		SLAM etc. The relevant notation is summarized in Table \ref{notation}. Consider a general robot with an arbitrary drive configuration, equipped with $m$ rotary encoders on its wheels and/or joints and an exteroceptive sensor such as a lidar or a camera. The exteroceptive sensor can sense the environment and generate scans or images $\V=\{\Z(t)\}_{t\in \mathcal{T}}$ that can be used to estimate its ego motion. Here, $\mathcal{T}:=\{t_1, t_2, \ldots, t_n\}$ denotes the set of discrete time instants at which the measurements are made. The rotary encoders output raw odometry data in form of a sequence of wheels angular velocities $\U = \{\db(t)\}_{t\in \mathcal{T}}$.  Given two time instants $t_j$ and $t_k$ such that $\Delta t_{jk}:=t_k-t_j > 0$ is sufficiently small, it is generally assumed that $\db(t) = \db(t_j):=\db_{jk}$ for all $t_j \leq t < t_k$. Traditionally, the odometry data is pre-processed to yield translation motion and orientation information, in the form of the robot pose $\{\q_j:=\q(t_j)\}_{j=1}^n$, and is subsequently fused with the ego motion estimates from exteroceptive and other sensors. Following the notation in Table \ref{notation}, the relative pose of the robot at time $t_k$ with respect to that at time $t_j$ is given by $\q_{jk}:=\circleddash \q_j  \oplus \q_k$. The pre-processing step necessitates the use of the motion model $\f_r$ of the robot that acts upon the odometry data $\db_{jk}$ to yield the relative pose of the robot  $\q_{jk} = \f_r(\db_{jk})$. Note that if the exteroceptive sensor is mounted exactly on the robot frame of reference, the sensor motion model $\f$ is the same as the robot motion model $\f_r$. In general however, if the pose of the exteroceptive sensor with respect to the robot is denoted by $\l$, the sensor motion model is given by $\f(\db_{jk}) = \circleddash \l \oplus \f_r(\db_{jk}) \oplus \l$, where generally $\l$ is also unknown.

		As shown in Fig.\ref{meth}, the goal of the calibration phase is to estimate the function $\f$, given $\mathcal{U}$ and $\mathcal{V}$ collected during the training phase. The estimated motion model, denoted by $\hat{\f}$, is subsequently used in the operational phase to augment or even complement the motion estimates provided by the exteroceptive sensor. More importantly, accurate odometry can be used to correct distortions in the sensor measurements \cite{LOAM}. We begin with studying the special case when the function $\f$ takes the form $\f(\bigcdot) = \g(\bigcdot\ ;\ \p)$ where $\g$ is a known function, and $\p$ is the set of unknown parameters, such as the dimensions of the wheel, sensor position w.r.t robot frame of reference etc. Knowing the form of $\g$ allows us to consider a simpler parametric problem that entails estimation of the relevant parameters $\p$. Two distinct scenarios are considered: the simpler case where the robot has a two-wheel differential drive that allows for joint scan-matching and calibration, and the more general case where sensor displacement is calculated a priori and provided as input to the calibration routine. A significantly more challenging scenario occurs when the form of $\f$ is not known, e.g. due to excess wear-and-tear, or is difficult to handle, e.g. due to non-differentiability. For such cases, the parametric approach is no longer feasible, and the unknown variable $\f$ is generally infinite dimensional. Towards this end, a low-complexity approach is proposed, wherein a simple but generic (e.g. linear) model for $\f$ is postulated. A more general and fully non-parametric Gaussian process framework is also put forth that is capable of handling more complex scenarios and estimate a broader class of motion models $\f$. It is remarked that in this case, unless the exteroceptive sensor is mounted on the robot axis, additional information may also be required to estimate the robot motion model $\f_r$.

		An interesting feature of the proposed class of algorithms is that they do not require the ground truth of the robot motion. Therefore, the calibration phase may be repeated as often as required depending on the rate of wear-and-tear of the robot wheels or sensor pose changes. More generally, the calibration phase can be integrated within the operational phase itself and may, therefore, be carried out without interrupting the robot operation. However, as will be shown later, the calibration phase does require the robot motion to comprise of both, rotation and translation at each step.

		While the formulation and techniques developed here can be applied to arbitrary exteroceptive sensors, for ease of exposition and testing, we will restrict ourselves to 2D Lidar that is mounted in such a way that it makes zero pitch and roll with robot axis. Likewise, although most of the calibration techniques are general, the fomulae and algorithms presented here will only consider a robot operating in a planar environment. 
		
		The 2D pose of the robot at time $t$ is denoted by $\q(t):=[q_x(t) ~q_y(t) ~ q_\theta(t)]^T$, consisting of the location coordinates and orientation of the robot with respect to a fixed frame of reference. The robot motion is governed by the following differential equation for $t\in \mathbb{R}$:
		\begin{equation}\label{robotode}
		\dot{\q}(t) := \frac{\mathrm{d}\q(t)}{\mathrm{d}t} =  
		\begin{pmatrix}
		v_x(t)\\
		v_y(t) \\ 
		\omega(t)  
		\end{pmatrix}= 
		\begin{pmatrix}
		v(t)\cos(q_\theta(t)) \\
		v(t)\sin(q_\theta(t)) \\ 
		\omega(t)  
		\end{pmatrix} 
		\end{equation}   
		where $v(t)$ and $\omega(t)$ are respectively the translational and rotational velocities of the robot and $v_x(t)$, $v_y(t)$ being the components of $v(t)$ along x-axis and y-axis respectively. Having introduced the preliminary notation, we discuss the three relevant examples that will subsequently be considered. 
		\begin{figure} 
			\centering  
			\subfigure[]{\includegraphics[width=0.49\linewidth,trim={0cm 0cm 0cm 0cm},clip]{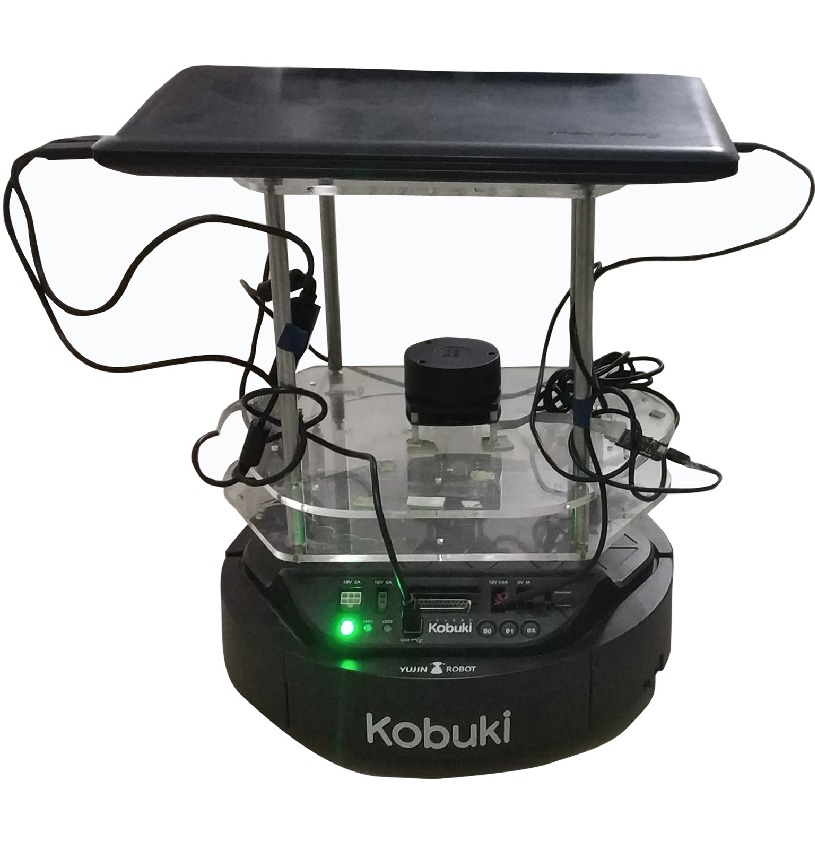}}
			\subfigure[]{\includegraphics[width=0.40\linewidth,trim={0cm 0cm 0cm 0cm},clip ]{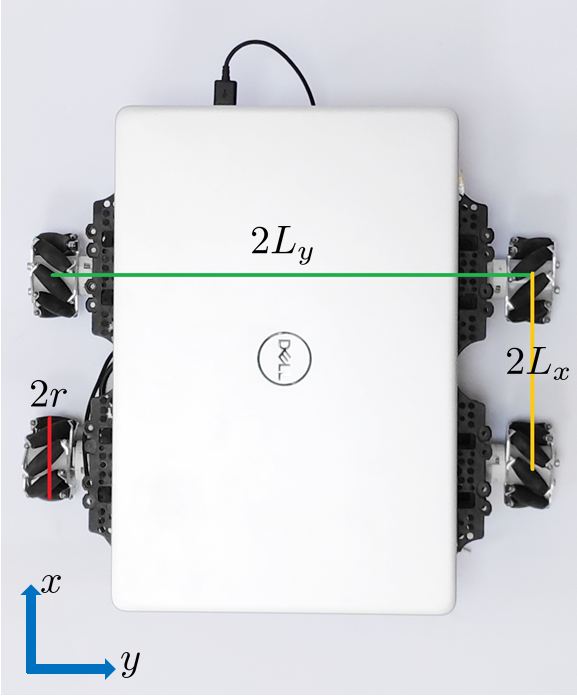}}
			\captionsetup{font=scriptsize}
			\caption{Robots used for experimental evaluations of proposed algorithms along with the state-of-the-art. (a) \emph{Kobuki} robot equipped with a 2D-lidar, wheel encoders and an on-board computer. (b) Similarly \emph{Fire Bird VI} robot.} \label{fig7}
		\end{figure}

		\noindent \textbf{Example 1:} Consider a \textbf{two-wheel differential drive} robot (see Fig. \ref{fig7}(a)) operating in a planar environment. Calibrating such a robot entails estimation of two sets of parameters (a) kinematic model parameters $\r:=(r_L, r_R, b) \in \mathbb{R}^3$, where $r_L$ and $r_R$ denote the radii of the left and right wheels respectively, and $b$ denotes the axle length; and (b) the pose of the exteroceptive sensor with respect to the robot frame $\l:=(\ell_x, \ell_y, \ell_\theta)\in \mathbb{R}^3$. For such a robot, the left and right wheel odometry sensors output a sequence of ticks, that can generally be converted into left- and right-wheel angular velocities, denoted by  $\omega_L(t)$ and $\omega_R(t)$ respectively. The robot translation and rotation velocities used in \eqref{robotode} are related to $\db(t):=[\omega_L(t)~\omega_R(t)]^T$ as follows, 
		\begin{align}\label{eq2}
		\begin{pmatrix}
		v(t)\\ 
		\omega(t)
		\end{pmatrix} &= \mathbf{J_\r}\db(t)
		\shortintertext{where }
		\mathbf{J_\r} &:= \begin{pmatrix}
		r_L/2 &r_R/2 \\ 
		-r_L/b & r_R/b
		\end{pmatrix}\label{eq3}
		\end{align}
		Henceforth, we denote the wheel angular velocities by $\omega_i(t)$ where $i = \{L, R\}$. 
		
		In order to derive the motion model, consider the time interval $[t_j,t_k)$ such that $\Delta t_{jk}$ is sufficiently small and the wheel velocities $\{\omega_i(t)\}_{i = L, R}$ remain approximately constant for $t\in [t_j, t_k)$. Such an approximation allows us to use the notation $\omega_i^{jk}:=\omega_i(t)$ for $t\in [t_j, t_k)$. In practice, the wheel angular velocities are obtained by counting the number of ticks recorded in the interval $[t_j,t_k)$, scaling it with the manufacture specified radians per tick factor, and dividing the result by $\Delta t_{jk}$. Consequently, from \eqref{eq3} we have that
		\begin{align}
		v(t) &:= \frac{1}{2} r_L \omega_L^{jk} + \frac{1}{2}r_R \omega_R^{jk} = v_{jk} \label{eq4}  \\ 
		\omega(t) &:= -\frac{1}{b}r_L \omega_L^{jk} + \frac{1}{b}r_R \omega_R^{jk} = \omega_{jk} \label{eq5}
		\end{align}
		for $t\in [t_j, t_k)$. Finally, the relative pose $\q_{jk}$ can be obtained by integrating  \eqref{robotode} over the interval $[t_j,t_k)$, that yields
		\begin{subequations}\label{qjk2w}
			\begin{align}
			q_{jk}^x &= v_{jk}\Delta t_{jk}(\sin \hspace{1mm} \omega_{jk}\Delta t_{jk})/(\omega_{jk} \Delta t_{jk}) \\
			q_{jk}^y &= v_{jk}\Delta t_{jk}(1-\cos \hspace{1mm} \omega_{jk}\Delta t_{jk})/(\omega_{jk} \Delta t_{jk}) \\
			q_{jk}^{\theta} &= \omega_{jk} \Delta t_{jk}.
			\end{align}    
		\end{subequations}
		The robot motion model $\f_r$ can be obtained by substituting \eqref{eq4}-\eqref{eq5} into \eqref{qjk2w}. It can be observed that the for the two-wheel differential drive robot, the sensor motion model $\f(\db_{jk}) = \circleddash \l \oplus \f_r(\db_{jk}) \oplus \l$ is completely specified, but for the parameters $\p = (\l, \r)$.

		\noindent \textbf{Example 2:} Consider an arbitrary complex wheel drive robot operating in a planar environment, for example, a \textbf{four-wheel Mecanum drive} robot as shown in Fig. \ref{fig7}(b). Calibrating such a robot entails estimation of kinematic model parameters $\r:=(r, L_x, L_y) \in \mathbb{R}^3$, where $r$ denotes the fixed radius of all the four wheels, $L_x$ denotes the half of axle length along x-axis of the robot, and $L_y$ denotes the same along the y-axis. For such a robot, the left-rear, right-rear, left-front and right-front wheel odometry sensors output a sequence of ticks, that can generally be converted into corresponding wheel angular velocities denoted by  $\omega_{Lr}(t)$ , $\omega_{Rr}(t)$ , $\omega_{Lf}(t)$, and $\omega_{Rf}(t)$, respectively. The robot translation and rotation velocities used in \eqref{robotode} are related to $\db(t):=[\omega_{Lr}(t)~\omega_{Rr}(t)~\omega_{Lf}(t)~\omega_{Rf}(t)]^T$ as follows, 
		\begin{align}\label{Mecanum}
		&\begin{pmatrix}
		v_x(t)\\ 
		v_y(t)\\
		\omega(t)
		\end{pmatrix} = \mathbf{J_\r}\ \db(t)
		\shortintertext{where }
		\mathbf{J_\r} := r  \hspace{1mm} 
		&{\begin{pmatrix}
			1 & 1 & 1 & 1 \\ 
			-1 & 1 & 1 &-1 \\
			-\dfrac{1}{L_x + L_y} & \dfrac{1}{L_x + L_y} & -\dfrac{1}{L_x + L_y} & \dfrac{1}{L_x + L_y}
			\end{pmatrix}}.
		\end{align} 
		Following the short hand notations used in Example 1, the wheel angular velocities are denoted by $\omega_i(t)$ where $i \in \{L_r, R_r, L_f, R_f\}$. Likewise, assume that $\Delta t_{jk}$ is small, so that $v_x(t) = v^x_{jk}$ and $v_y(t) = v^y_{jk}$ for $t \in[t_j,t_k)$.  Consequently, the relative poses are given by    
		\begin{subequations}\label{Mecanum_model}
			\begin{align}
			q_{jk}^x &= v^x_{jk} \Delta t_{jk} \\
			q_{jk}^y &= v^y_{jk} \Delta t_{jk} \\
			q_{jk}^{\theta} &= \omega_{jk} \Delta t_{jk}.
			\end{align}    
		\end{subequations}
		We remark here that for the four-wheel Mecanum drive robot, the sensor motion model $\f(\db_{jk}) = \circleddash \l \oplus \f_r(\db_{jk}) \oplus \l$ is completely specified, but for the parameters $\p = (\l, \r)$.

		\noindent \textbf{Example 3:} Finally, consider a two-wheel differential drive and a four-wheel Mecanum drive robots suffering from hardware malfunctions such as deformation of one of the wheels (see Fig. \ref{fig_def}), unaligned wheel axis, or tilted wheel (see Fig. \ref{meca}). The original kinematic model of the respective robots with such deformations no longer captures the true motion behaviour, and therefore cannot be used. In general, such distortions introduce new parameters that are difficult to model.  Despite the distortions, the robots continue to output angular velocities $\db(t)$ (of dimension $m$) that can be used to determine $\f:\mathbb{R}^m \rightarrow \mathbb{R}^3$ using the ego-motion estimates of the exteroceptive sensors. Indeed, for such cases, it only makes sense to talk about the sensor motion model and not the robot motion model, which is unidentifiable unless the position of the sensor (parameterized by $\l$) is known. Defining constant angular wheel velocities as in Example 1 for a small interval $[t_j,t_k)$, the relative pose of the sensor would be given by $\q_{jk} = \f(\db_{jk})$ and the goal is to estimate $\f$. 
		
		In summary, the goal is to learn the function $\f$, given raw odometry $\U$ and sensor measurements $\V$ while being robust to outliers in an automated manner. For the case when the parametric form of the function $\f$ is known, the problem entails estimating the associated parameters $\p$ involving robot intrinsic parameters $\r$ and exteroceptive sensor position $\l$. On the other hand, when the kinematic model of the robot is not known, the aim is to learn the non-parametric form of the function $\f$.

		\section{Calibration Methodology}\label{sec4}
		This section details the algorithms for various cases discussed in Sec.\ref{probfor}, namely, calibration of robots with two-wheel differential drive configuration (Sec. \ref{twowheel}), robots with generic but known drive configurations (Sec. \ref{anywheel}), and finally, robots with unknown or unmodeled drive configurations (Sec. \ref{anybot}). The algorithms for the more general cases can always be used in the special cases, e.g., a robot with two-wheel differential drive can be calibrated by any of the techniques presented in this section. However, using a more general method incurs a higher computational complexity and may result in lower calibration accuracy.

		\subsection{Autonomous calibration of the two-wheel differential drive robots} \label{twowheel}
		Consider a two-wheel differential drive robot equipped with a Lidar. Recall from Example 1 that for this case, the parameters of interest are $\p=(\l,\r)$ and the kinematic model of the robot is given by \eqref{eq4}-\eqref{qjk2w}. The simple form of the kinematic model for the two-wheel drive motivates the formulation of a joint scan-matching and parameter estimation problem that can be solved via the alternating minimization algorithm. Different from the existing unsupervised calibration algorithms, e.g. \cite{censi}, the proposed algorithm builds upon the ICP framework \cite{besl,tricp} and allows automated outlier rejection. However, the overall approach here is general and can be used with other scan matching algorithms such as the point-to-line ICP (PLICP) \cite{P2LICP}; see \ref{appB}.

		Let $\{\z^{(i)}(t)\}_{i=1}^{\left | \Z(t) \right |}$ denote the Cartesian coordinates of the scan points expressed with respect to the Lidar frame and collected at time $t \in \mathcal{T}$. Here, $\Z(t)$ denotes the set of all scan points collected at time $t$. Following the notation introduced in Sec. \ref{probfor}, the scan points can be transformed into the robot frame and written as $\{\l \oplus \z^{(i)}(t)\}_{i=1}^{\left | \Z(t) \right |}$. We drop the subscript $i$, also $t$ and denote the scan points collected at times $\{t_j\}$ by $\{\z^{(i)}_j\}$. For any point $\z^{(i)}_j$, the distance to the closest point in $\Z(t_k)$ is given by 
		\begin{align} \label{distclose}
		d^{(i)}_{jk} := \min_{\z \in \Z(t_k)} \left\|\q_{jk} \oplus \l\oplus\z - \l\oplus\z^{(i)}_j\right\|^2_2
		\end{align}
		Within the ICP framework, the correspondence error is given by 
		\begin{align}\label{fobj}
		h(\r,\l) = \sum_{(j,k) \in \E}\sum_{i\in\Z_{jk}} d^{(i)}_{jk}
		\end{align}
		where $\E$ represents set of all chosen scan pairs and $\Z_{jk} \subset \Z(t_k)$ consists of points for which the distances $d^{(i)}_{jk}$ are the smallest. In order to allow partial overlap between the scan pairs $\Z(t_j)$ and $\Z(t_k)$, we utilize a trimming procedure inspired from \cite{tricp}. If the distances $\{d^{(i)}_{jk}\}$ are sorted and the $\iota$-th smallest distance is denoted by $d^{[\iota]}_{jk}$, then the trimmed correspondence error may be written as 
		\begin{align}\label{fobj2}
		h(\r,\l) = \sum_{(j,k) \in \E}\sum_{\iota = 1}^{I_{jk}} d^{[\iota]}_{jk}.
		\end{align}
		Here, the number of points under consideration $I_{jk}$ is generally decided on the basis of the overall trimmed correspondence error value. Trimming, by choosing $I_{jk} < |\Z(t_k)|$, not only allows partial overlaps between scan pairs but also handles erroneous measurements and shape defects. The overlap parameter $\zeta = I_{jk}/|\Z(t_k)|$ is determined automatically by solving a simple optimization problem as detailed in \cite[Sec. 3]{tricp}. Recall that in \eqref{distclose}, $\q_{jk}$ is a function of $\r$ and is given by \eqref{qjk2w} for the two-wheel drive. As in the classical ICP algorithm, for each pair of scans in $\E$, the scan points are first transformed into a common frame of reference and then compared.  The parameters are recovered by solving the following optimization problem
		\begin{align}\label{hmin}
		(\r^{\star},\l^\star) = \arg\min_{\r,\l} h(\r,\l). 
		\end{align}
		Note that while in theory, one could impose constraints of the form $\r\geq 0$, such constraints would not be useful in practice. Indeed, a solution on the boundary (e.g. with $r_L = 0$) is also not acceptable, and would generally necessitate collecting more measurements for calibration. 
		
		The problem in \eqref{hmin} can be viewed as a generalization of the trimmed ICP framework, where the parameter $\l$ does not appear. The inclusion of $\l$, however, complicates the optimization problem, which is already non-convex and difficult to solve. To this end, we propose the following algorithm: starting from an initial estimate $(\r^{0},\l^{0})$, perform the two steps iteratively (a) establish correspondences $\Z^{\alpha}_{jk}$ based on $(\r^{\alpha},\l^{\alpha})$; and (b) update $(\r^{\alpha+1},\l^{\alpha+1}) = \arg\min_{\r,\l} h^{\alpha}(\r,\l)$ where $h^{\alpha}$ is obtained from $h$ while using the given correspondences $\Z^{\alpha}_{jk}$. Of these, the first step is straightforward and implements the trimming procedure, as outlined in \cite{tricp}. The step (b) is however still complicated, and we solve it via an alternating minimization approach. Also referred to as the block-coordinate descent (BCD) method \cite{bertsekasNLP}, the approach entails alternating minimization of $h^\alpha$ with respect to $\l$ and $\r$. While the algorithm is not guaranteed to converge except under specific conditions \cite{tseng}, it is known to work well in practice and has been used to solve a wide variety of problems in communications and signal processing. 
		
		In the present context, the algorithm admits a natural interpretation, namely alternate intrinsic and extrinsic calibration, summarized in the following Lemma.
		
		\textbf{\emph{Lemma 1.}} \emph{Given correspondences, the following problems:
			\begin{align}
			\text{Extrinsic calibration given $\r'$: } &\min_{\l\in\L} h^\alpha(\r',\l) \label{hminl}\\
			\text{Intrinsic calibration given $\l'$: } &\min_{\r\geq 0} h^\alpha(\r,\l') \label{hminr}
			\end{align}
			can both be solved efficiently. Specifically,\eqref{hminl} can be solved in closed-form while \eqref{hminr} can be solved via two-dimensional grid search. }
		
		Note that the subproblems in \eqref{hminl}-\eqref{hminr} are both non-convex. Interestingly, however, Lemma 1 establishes that the global minima of these subproblems can be readily found. The full calibration via alternating minimization (CAM) algorithm is summarized in  Algorithm \ref{alg1}. The parameters are initialized either from manufacture supplied values or from the values obtained via manual measurements. The development of the closed form solutions to \eqref{hminr}-\eqref{hminl} is carried out in \ref{appA}. Having developed the joint scan matching and calibration algorithm, discussion on various implementation related issues is due. 
		
		\begin{algorithm}
			\caption{CAM algorithm for autonomous calibration of two wheel differential drive robots} 
			\label{alg1}
			\begin{algorithmic}[1]
				\State  Collect scans and wheel odometry measurements 
				\State  Initialize parameters i.e $\p^{\alpha} = (\r^\alpha, \l^\alpha)$, with $\alpha=0$
				\State  Select scan pairs based on the criteria detailed in \ref{ss}  
				\Repeat 
				\State Establish correspondences with current best estimate of parameters  
				\State  \textbf{Perform Alternating Minimization :} \\ 
				\hspace{1cm}$\beta = 0,\ \p^\beta = (\r^\beta,\l^\beta) = (\r^\alpha, \l^\alpha) = \p^{\alpha} $  
				\Repeat  
				\State   \textbf{Extrinsic calibration of Lidar :} \\
				\hspace{2cm}$\l^{\beta+1} \leftarrow  \arg \min_{\l} h^\alpha(\r^\beta,\l)$ 
				\State    \textbf{Intrinsic calibration of wheel odometry :}  \\   
				\hspace{2cm}$\r^{\beta+1} \leftarrow \arg \min_{\mathbf{r}} h^\alpha(\r,\l^{\beta+1})$  
				\State  $\beta \leftarrow  \beta +1$     
				\Until Convergence  $\left \| \p^{\beta +1} - \p^\beta\right \| \leq \epsilon$
				\State   $\alpha \leftarrow \alpha +1$, update $\p^\alpha$
				\Until Convergence
			\end{algorithmic}
		\end{algorithm}

		\subsubsection{Observability and uniqueness} As in classical parameter estimation settings, it is necessary to explicate the limitations of the proposed algorithm. Observability analysis seeks to identify the conditions and constraints under which the estimates obtained from Algorithm 1 are reasonably close to the actual robot parameters. The idea is to generate a set of guidelines for the generation of measurements that make the system parameters observable. 
		
		As a simple example, consider a robot that always moves along the x-axis in a straight line. Then, for any two time points $t_j$ and $t_k$, it holds that $q^x_{jk} = v_{jk}\Delta t_{jk}$ while $q^y_{jk} = q^\theta_{jk} = 0$. Recall from \eqref{eq4} that $v_{jk}$ depends only on $r_L$ and $r_R$ but not on the axle length $b$. Substituting these into \eqref{fobj}, it can be seen that in this case, $h$ would also not depend on $b$, i.e., $\nabla_b h(\r,\l) = 0$ regardless of the value of $b$. Indeed, if the robot only moves along the x-axis, the parameter $b$ would remain unobservable irrespective of the calibration method used.
		
		More generally, the following proposition specifies the guidelines for robot motion during the calibration process. 
		
		\textbf{\emph{Proposition 1.}} The sensor parameters are unobservable if the robot motion comprises of either pure translations or pure rotations alone. 
		
		The proof of Proposition 1 is provided in \ref{appC}. The observability analysis reveals the conditions under which certain parameters would remain unobserved. For example, it can be seen from \eqref{fobj} that for pure rotation (i.e. when  $q_{jk}^{x} = q_{jk}^y = 0$), the function $h$ does not depend on intrinsic parameters $r_L$ and $r_R$. In other words,  $r_L$ and $r_R$ are not observable if all the scan pairs in $\E$ correspond only to pure rotations. In other words, if the training phase is comprised of pure rotations alone, inferring the radii $r_L$ and $r_R$ would be impossible.  The requirement for the robot to be sufficiently mobile during the training phase is generally always required for passive calibration schemes \cite{censi}.
		
		Before concluding, we remark here that the solution to \eqref{hmin} is not necessarily unique, even when all the parameters are observable. In particular, it can be observed from \eqref{hmin} that both $(r_L,r_R,b,\ell_x,\ell_y,\ell_{\theta}    )$ and $(-r_L,-r_R,-b,-\ell_x,-\ell_y,\ell_{\theta}+\pi)$ yield the same value of $h(\r,\l)$ and $\nabla h(\r,\l)$. Such parameter ambiguity is often unavoidable in calibration algorithms; see, e.g. \cite{censi}. In the present case, however, the ambiguity can be resolved by selecting the solution that has positive values of $(r_L, r_R, b)$. In general, if the initial values are not too far from the actual values, the proposed Algorithm was found to converge to the vicinity of the correct solution. 
		
		\subsubsection{On the choice of scan pairs and trajectory}\label{ss}
		While the proposed calibration algorithm allows for the calibration to be carried out while the robot is operating, the observability analysis does impose certain restrictions on the robot trajectory. Specifically, for the parameters to be observable with sufficiently high accuracy, the overall motion of the robot during the calibration phase should involve both translation and rotation. For instance, the estimated parameters would be highly inaccurate if the robot continues to move along an almost straight line or continues to turn around without moving. 
		
		A typical calibration routine consists of a large number of scans and including every possible scan pair in $\E$ is inefficient and computationally demanding. Given that the robot trajectory adheres to such a restriction, the efficiency and accuracy of the calibration phase can both be improved by intelligently selecting the scan pairs. Specifically, based on robot odometry with nominal parameters, we choose scan pairs that have non-zero translation and rotation between them. Here, it is essential to ensure that scan pairs still correspond to sufficiently close robot locations, lest our assumption regarding small $\Delta t_{jk}$ is violated. The simulations utilize a heuristic upper bound on the translation distance to ensure this. 
		
		\subsubsection{Automatic outlier rejection} 
		Outliers enter into the system either through scan points that belong to non-overlapping regions or due to wheel slippages. The proposed CAM algorithm automatically removes both kinds of outliers; specifically, the trimmed correspondence error in \eqref{fobj2} allows for partial overlaps and the Huber loss function incorporated within the subproblem \eqref{hminr} eliminates outliers due to wheel slippages (see \ref{appA}). The automatic outlier rejection approach contrasts existing calibration algorithms that require manual trimming \cite{censi} or additional hardware \cite{tricycle}.     
		
		\subsubsection{Tuning parameters for CAM} Recall that the CAM overlap parameter $\zeta$ is found by solving an optimization problem as detailed in \cite{tricp}. The corresponding preset parameter for the trimming procedure is set to be unity since the scan selection criteria yielded scans with considerable overlap. For this choice of preset parameter, the performance of the CAM algorithm was relatively robust to the choice of the Huber parameter $c$, which was also selected to be unity.

		\subsection{Autonomous calibration of robots with arbitrary but known drive configurations}\label{anywheel}
		This section develops a general-purpose calibration algorithm that can be applied to a robot with any given drive configuration. As in Sec. \ref{twowheel}, the parameters of interest are denoted by $\p=(\r,\l)$, where $\r$ collects the intrinsic robot parameters while $\l$ denotes the pose of the exteroceptive sensor in the robot frame. Unlike Sec. \ref{twowheel} however, the exteroceptive sensor need not be a 2D Lidar, but any sensor capable of estimating its ego-motion, e.g., a camera. 
		
		As in Sec. \ref{twowheel}, let $\mathcal{T}$ denote the set of times at which the sensor observations are made. 
		Consider a pair of times $[t_j,t_k) \in \mathcal{T}, \ni t_k>t_j$ and $\Delta t_{jk} = t_k-t_j$ is not too large. 
		Recall that the relative pose of the robot $\q_{jk}(\r)$ is a function of the intrinsic robot parameters. Given $\l$, the sensor displacement between $t_j$ and $t_k$ can be calculated as 
		\begin{equation} \label{sjk}
		\s_{jk}(\r,\l) = \circleddash\  (\q_{j} \oplus \l) \oplus (\q_{k} \oplus \l) = \circleddash\ \l \oplus \q_{jk}(\r) \oplus \l 
		\end{equation}
		where $\q_{jk}(\r)$ encodes the robot motion model, as detailed in Sec. \ref{probfor}. For certain pair of times $[t_j,t_k)$, the exteroceptive sensor may generate an estimate of its ego motion, henceforth denoted by $\hat{\s}_{jk}$. For most sensors, such estimates are also accompanied by error variances $\Sig_{jk}:=\text{diag}((\sigma^x_{jk})^2,(\sigma^y_{jk})^2,(\sigma^\theta_{jk})^2)$ that quantify the estimation error variances in $x$, $y$, and $\theta$ estimates in $\hat{\s}_{jk}$. Defining the set of all scan pairs $\E:=\{(j,k) \mid \hat{\s}_{jk}$ \text{ is available and $\Delta t_{jk}$ small}\}, the calibration problem can be posed within the non-linear least squares framework as 
		\begin{align}\label{anyobj}
		(\hat{\r},\hat{\l}) &= \arg\min_{\r,\l} h(\r,\l):=\!\!\sum_{(j,k)\in\E} \left\|\hat{\s}_{jk}-\s_{jk}(\r,\l)\right\|^2_{\Sig_{jk}^{-1}}
		\end{align}
		Define the residual vector $\boldsymbol{\res}_{jk}(\r,\l) := \hat{\s}_{jk}-\s_{jk}(\r,\l)$ and denote its $i$-th element by $\res^i_{jk}(\r,\l)$ for  $i\in\{x,y,\theta\}$. Observe that the objective function in \eqref{anyobj} is also the negative log-likelihood $-\log p(\{\hat{\s}_{jk}\} ; \r,\l)$ if the uncertainty in $\hat{\s}_{jk}$ is modeled as independent Gaussian distributed with zero mean and co-variance matrix $\Sig_{jk}$. Such an interpretation allows \eqref{anyobj} to be interpreted as the maximum likelihood (ML) estimator \cite{censi, tricycle}. 
		
		The exteroceptive sensor output is well known to be prone to outliers, e.g., due to the scan-matching failures. On the other hand, the ordinary least squares estimator tends to fail catastrophically even in the presence of a single outlier \cite{robustls}. In order to  handle such outliers, we resort to the class of robust M-estimators. Rewriting the objective function in \eqref{anyobj} in terms of $\res^i_{jk}$, we obtain
		\begin{align}\label{anyobj2}
		h(\r,\l) = \sum_{(j,k)\in\E} \sum_{i\in \{ x,y,\theta\}} \rho\left(\frac{\res^i_{jk}(\r,\l)}{\sigma^i_{jk}}\right)
		\end{align} 
		where $\rho(u) = \frac{u^2}{2}$. The robust counterpart of \eqref{anyobj2} is obtained by replacing the squared loss function $\rho$ with the Huber function $\rho_c$ defined as \cite{huber}
		\begin{align}\label{rhoc}
		\rho_c(u) = \begin{cases} \frac{u^2}{2} & |u| \leq c \\
		c(|u|-\frac{c}{2})    & |u| > c.
		\end{cases}
		\end{align}
		It can be seen that the Huber function is the same as the squared law function for $|u| \leq c$ but becomes linear in $|u|$ for $|u| > c$. The Huber estimator obtained by plugging in $h_c(\r,\l):=\sum_{(j,k)\in\E} \sum_{i\in \{ x,y,\theta\}} \rho_c\left(\frac{\res^i_{jk}(\r,\l)}{\sigma^i_{jk}}\right)$ in \eqref{anyobj} is robust to outliers in $\hat{\s}_{jk}$. Alternatively, the Huber estimator can be viewed as a special case of a sparsity controlling outlier rejection framework \cite{sparsity}. Note that the Huber estimator is inefficient when the noise is actually Gaussian distributed. For instance, the choice $c = 1.345$ results in about 95\% efficiency \cite{huber}. 
		
		In the following subsections, we discuss the methodology to solve \eqref{anyobj} with the Huber loss function, followed by observability analysis.
		
		\begin{figure}
			\centering  
			\subfigure[]{\includegraphics[width=0.51\linewidth]{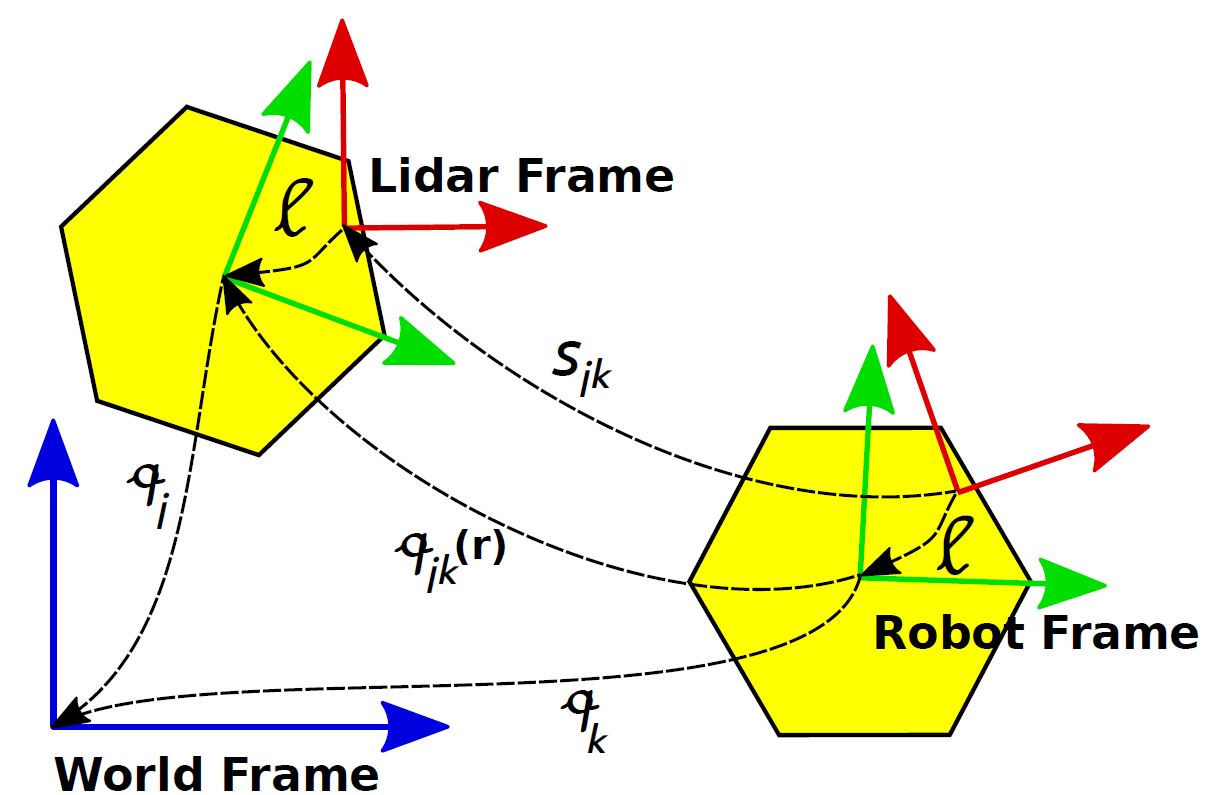}} 
			\subfigure[]{\includegraphics[width=0.45\linewidth,height=3.5cm]{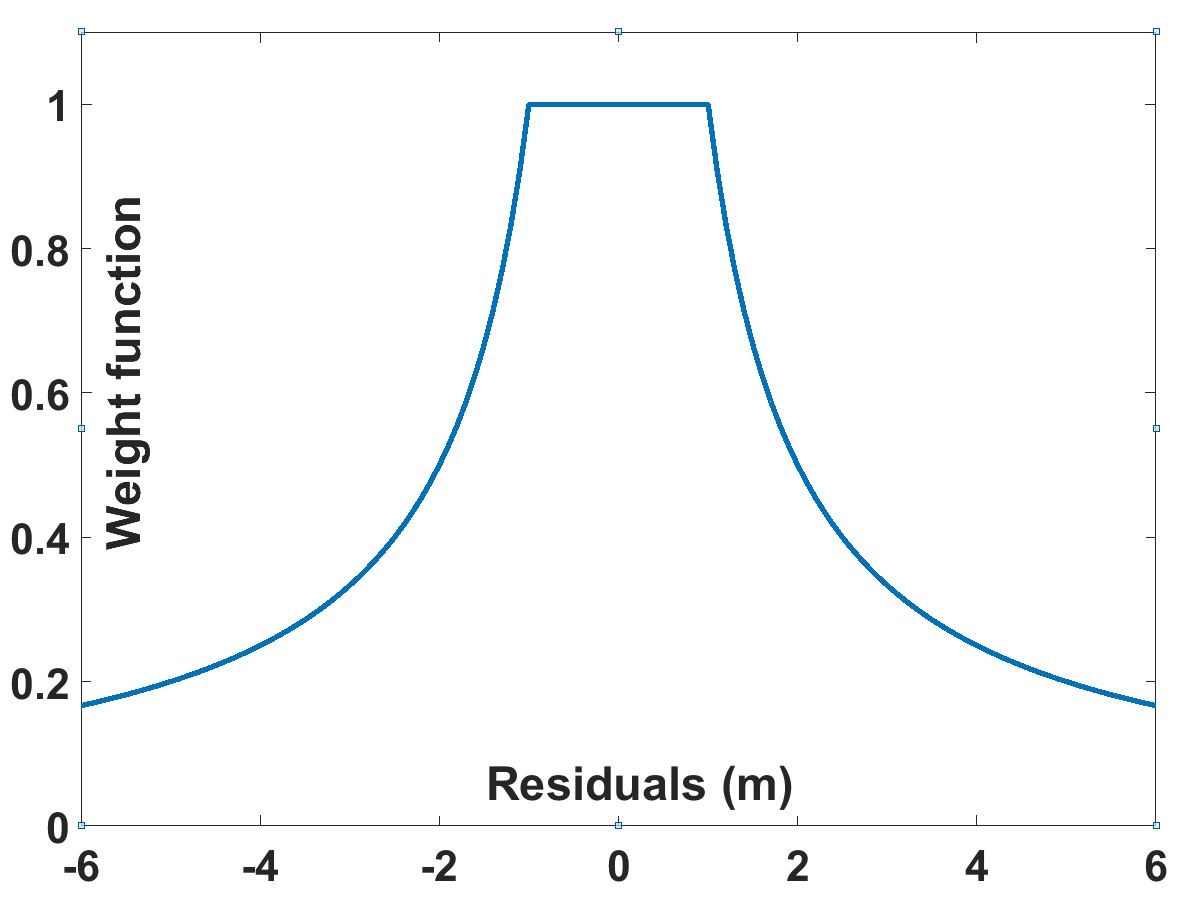}}
			\captionsetup{font=scriptsize}
			\caption{ (a) Illustration of robot poses $\q_j$,$\q_k$ w.r.t to world frame; $\l$ represents pose of Lidar sensor w.r.t robot frame; $\q_{jk}$ is the relative pose of $\q_k$ w.r.t $\q_j$. (b) Weight function plotted against residuals for Huber loss.} \label{fig1} 
		\end{figure}

		\subsubsection{Iteratively Re-weighted Least Squares (IRLS) Algorithm} In general, both $h$ and $h_c$ are non-convex functions of $(\r,\l)$ and are difficult to work with. Moreover, the function $\rho_c$ is not twice differentiable so classical second order methods (e.g. Newton method) cannot be directly applied. Instead, we put forth an IRLS variant that solves a non-linear least squares problem at each step. Recalling that $\p = (\r,\l)$, the equation $\nabla h_c(\p) = 0$ can be equivalently written as
		\begin{align}\label{lsdiff}
		\sum_{(j,k)\in\E} \sum_{i\in \{ x,y,\theta\}} w_{jk}^i(\res_{jk}^i(\p)) \nabla_\p \left(\frac{1}{2}\left[\res_{jk}^i(\p) \right]^2\right) = 0
		\end{align}
		where the weight function is defined as
		\begin{align}\label{wjk}
		w_{jk}^i(u) := \begin{cases} \frac{1}{\left(\sigma_{jk}^i\right)^2} & |u| \leq c \\
		\frac{c}{|u|\left(\sigma_{jk}^i\right)^2} & |u| \geq c
		\end{cases}
		\end{align}
		and shown in Fig.\ref{fig1}(b). Observe that given the weights $\{w_{jk}^i\}$, solving \eqref{lsdiff} is equivalent to solving a non-linear weighted least squares (NLS) problem. Assuming that such a problem can be solved readily, the IRLS algorithm starts with an initial guess $\hat{\p}^{(0)}$ and entails carrying out the following iterations for $\alpha \geq 0$:
		\begin{align}\label{irls}
		\hat{\p}^{(\alpha+1)} &= \arg\min_{\p} \!\!\sum_{(j,k)\in\E} \left\|\boldsymbol{\res}_{jk}(\p)\right\|^2_{\W_{jk}^{(\alpha)}}
		\end{align}
		where $\W_{jk}^{(\alpha)}$ is a diagonal matrix whose $(i,i)$-th entry is given by  $[\W_{jk}^\alpha]_{ii} = w_{jk}^i(\res_{jk}^i(\p^{(\alpha)}))$. Note that the non-linear least squares problem in \eqref{irls} is similar to that in \eqref{anyobj} except for the changing weights. Classical approaches for solving the NLS include (a) Newton method that can be applied when the motion model $\q_{jk}(\r)$ is twice differentiable in $\r$; and the (b) Levenberg Marquardt (LM) algorithm otherwise. In general, LM algorithm is preferred when a lower computational complexity is desired while the Newton method is more suited when the Hessian can be calculated easily.

		Based on empirical observations, we propose two enhancements to the IRLS algorithm. First, when the number of outliers is large, there may be a large number of residual terms with very small weights. While the contribution of each of these residual terms may be small, their combined contribution may still be significant and may bias the estimate. Ideally, the residuals for which the corresponding weights are very small should be completely eliminated from the optimization process. Towards this end, we propose to trim the weights at every iteration as follows:
		\begin{align}\label{eq47}
		\tilde{w}_{jk}^i(u) &= \begin{cases} 0 & w_{jk}^i(u) \leq \gamma \\
		w_{jk}^i(u) & w_{jk}^i(u) > \gamma
		\end{cases}
		\end{align}  
		where $\gamma = 1- \frac{1}{n}\sum\limits_{k=1}^{n} w_{jk}^i(u)$. To understand the trimming process, observe that when there are few outliers in the data, several of the weights $w_{jk}^i(u)$ are close to 1, and consequently, $\gamma \approx 0$. As a result, no trimming occurs in such a scenario, and all the residual terms contribute to the optimization. On the other hand, when the number of outliers is very large, $\gamma$ is larger, and a number of residuals having small weights may get trimmed. As a result, even when the level of corruption is high, the contribution of the outlier residual terms still stays small. The justification on the effectiveness of the trimming procedure will be substantiated in the experimental section. Second, the residuals are adjusted using leverage, as suggested in \cite{adjust}. The adjusted residuals are denoted by $\tilde{\upsilon}_{jk}^i(\p)$ and full IRLS algorithm is summarized in Algorithm \ref{alg2}. 
		
		\begin{algorithm}
			\caption{CIRLS algorithm for autonomous calibration of arbitrary drive robots} 
			\label{alg2}
			\begin{algorithmic}[1]
				\State  Collect measurements from wheel odometry and extereoceptic sensors over any sufficiently exciting trajectories
				\State   Now run the corresponding sensor displacement algorithm for each selected interval, to get the estimates  $\{\hat{\s}_{jk}\}$ 
				\State  Initialize parameters  $\p^{\alpha} = (\l^{\alpha},\r^{\alpha}) $, with $\alpha = 0$
				\State   Also initialize the weight matrix to $\textbf{W} = \textbf{I}$ 
				\Repeat 
				\State  Using current weight matrix ($\textbf{W}$) and available best estimate of  parameters ($\p^{\alpha}$), solve the following:\\  
				\hspace*{0.4cm} $ \hat{\p} = \arg\min_{\p} \!\!\sum_{(j,k)\in\E} \left\|\boldsymbol{\res}_{jk}(\p)\right\|^2_{\W_{jk}^{(\alpha)}}
				$
				\State This can be done by employing either \textbf{Gauss Newton} or \textbf{ Levenberg Marquardt algorithm}
				\State Update ${\p^\alpha} $, also update subsequent weights using equation \eqref{eq47}, thus $\textbf{W}$  
				\Until Convergence
			\end{algorithmic}
		\end{algorithm} 
		Note that in the context of two-wheel differential drive and the tricycle drive, the weighted least squares problem in \eqref{irls} can be solved in closed-form if the following condition holds \cite{censi, tricycle}
		\begin{align}
		w_{jk}^x = w_{jk}^y.\label{sigxy}
		\end{align}
		While such a condition would not generally hold since the weights in \eqref{wjk} depend on the residuals, the availability of closed-form solution simplifies the overall algorithm significantly. Specifically, the proposed CIRLS algorithm reduces to alternatively solving \eqref{irls} in closed-form and updating the weights as in \eqref{wjk} or \eqref{eq47}. It is possible to explicitly force \eqref{sigxy} to hold by setting 
		\begin{equation}
		w^i_{jk} = \max\{w_{jk}^x,w_{jk}^y\} \hspace{ 5mm} for \hspace{3mm} i \in \{x,y\} 
		\end{equation}
		for all $(j,k) \in \E$. In this case, equal weight is applied to all the residuals in the first iteration. We refer to this case as CIRLS CF (CIRLS with closed forms).

		\subsubsection{Observability and Covariance analysis of estimated parameters} The observability analysis for $\l$ is similar to that in Sec. \ref{twowheel}. That is, pure translations alone, make $\ell_\theta$ unobservable while pure rotations alone make $\ell_x$ and $\ell_y$ unobservable. Therefore the robot trajectory in the calibration phase should not consist entirely of pure translations or pure rotations solely. The observability of the intrinsic parameters depends on the motion model of the robot and must be explicitly analyzed for a given drive configuration. It can be seen for instance that if one or more entries of $\nabla \q_{jk}(\r)$ are zero, the NLS problem may become ill-conditioned. 
		
		The uncertainty in the estimated parameters can be found as follows: 
		\begin{equation}
		\boldsymbol{\Sigma} = \textbf{\emph{mse}} \times (\textbf{J}^T\textbf{J})^{-1}
		\end{equation}  
		where $\textbf{\emph{mse}}$ is the mean of squared weighted residual terms evaluated at the converged solution and the $e$-th row of $\textbf{J}$ is $\nabla_\p \s_e(\r,\l)$ for all $e\in\E$.
		
		\subsection{Autonomous calibration of robots with un-modeled wheel deformations}\label{anybot}
		When no information about the kinematic model of the robot is available, it becomes necessary to estimate $\f$ directly. As in Sec. \ref{anywheel}, let $\{t_j\}$ be the set of time instants at which measurements are made. For certain pairs of times $[t_j,t_k) \in \E$ for which $\Delta t_{jk}$ is not too large, the exteroceptive sensor generates motion estimates $\hat{\s}_{jk}$. Given data of the form $\D:=(\db_{jk}, \hat{\s}_{jk})_{(j,k)\in\E}$ and $n:=|\E|$, the goal is to learn the function $\f:\Rn^m\rightarrow \Rn^3$ that adheres to the model
		\begin{align}
		\hat{\s}_{jk} = \f(\db_{jk}) + \boldsymbol{\varepsilon}_{jk}
		\end{align}
		for all $(j,k)\in\D$, where $\boldsymbol{\varepsilon}_{jk} \in \Rn^{3}$ models the noise in the measurements, and  $\db_{jk}$ now represents the wheel ticks recorded in the time interval $\Delta t_{jk}$. As in Sec. \ref{anywheel}, it is assumed that the noise variance $\Sig_{jk}$ is known. Given an estimated $\hat{\f}$ of the sensor motion model, new odometry measurements $\db_\star$ can be used to directly yield sensor pose changes $\s_\star = \hat{\f}(\db_\star)$. As remarked earlier, it may be possible to obtain the robot pose change $\q_\star$ from $\s_\star$ if the sensor pose $\l$ is known a priori. Since the functional variable $\f$ is infinite dimensional in general, it is necessary to postulate a finite dimensional model that is computationally tractable. Towards solving the functional estimation problem, we detail two methods, that are very different in terms of computational complexity and usage flexibility.
		
		\subsubsection{Calibration via Gaussian process regression}
		The GP regression approach assumes that the measurement noise is Gaussian distributed and assumes that the $\hat{\s}_{jk}$ also follows normal distribution with mean $\f(\db_{jk})$. Specifically, we have the likelihood as 
		\begin{align}
		p(\hat{\s}_{jk}|\f(\db_{jk})) = \mathcal{N}(\hat{\s}_{jk}|\f(\db_{jk}), \Sig_{jk})
		\end{align}
		or equivalently, $ \bm{\varepsilon_{jk}} \sim \mathcal{N}(0,\Sig_{jk})$. Given inputs $\{\db_{jk}\}$, let $\fb$ denote the $\{3n \times 1\}$ vector that collects $\{\f(\db_{jk})\}$ for $\{(j,k)\in\E \}$. Defining $\Sig \in \Rn^{3n \times 3n}$ as the block diagonal matrix with entries $ \Sig_{jk}$ and $\hat{\s} \in \Rn^{3n}$ as the vector that collects all the measurements $\{\hat{\s}_{jk}\}_{(j,k)\in\E}$. Having this we can equivalently write the joint likelihood as
		\begin{align}
		p(\hat{\s}|\fb) = \mathcal{N}(\hat{\s}|\fb, \Sig)
		\end{align}

		Unlike the parametric model in Sec. \ref{anywheel}, we impose a Gaussian process prior on $\fb$ directly. Equivalently, we have that
		\begin{align}
		p(\fb) = \mathcal{N}(\fb|\mathbf{\bar{\mub}}, \K)
		\end{align}
		where $\bar{\mub} \in \Rn^{3n}$ is the mean vector with stacked entries of $\mub(\db_{jk}) \in \Rn^3$ and the covariance matrix  $\K \in \Rn^{3n \times 3n}$ has entries $[\K_{jk,j'k'}] = \boldsymbol{\kappa}(\db_{jk},\db_{j'k'})$ for $(j,k)$ and $(j',k') \in \E$ . The choice of the mean function $\mub:\Rn^m \rightarrow \Rn^3 $ and kernel function $\boldsymbol{\kappa} :\Rn^{m} \times \Rn^m \rightarrow \Rn^{3\times3}$ is generally important and application specific. Popular choices include the linear, squared exponential, polynomial, Laplace, and Gaussian, among others. Next, the predictive posterior for a new test input $\db_\star$ is defined as follows,
		\begin{align}
		p(\hat{\s}_\star|\hat{\s}) = \int p(\hat{\s}_\star|\hat{\f}(\db_\star))\cdot p(\hat{\f}(\db_\star)|\hat{\s})\cdot d\hat{\f}(\db_\star)
		\end{align}
		where $p(\hat{\f}(\db_\star)|\hat{\s}) = \int p(\hat{\f}(\db_\star)|\fb) \cdot p(\fb|\hat{\s})\cdot d\fb$ and also note $p(\hat{\f}(\db_\star)|\fb)$ must be Gaussian. With a Gaussian prior and noise model, the posterior distribution of $\f$ given $\D$ is also Gaussian. For a new odometry measurement $\db_\star$ with noise variance $\Sig_{\star}$, let $\k_\star \in \Rn^{3n \times 3}$ be the vector that collects $\{\boldsymbol{\kappa}(\db_\star, \db_{jk})\}_{(j,k)\in\E}$. Then the distribution of $\hat{\f}(\db_\star)$ for given $\hat{\s}$ is
		\begin{align}
		p(\hat{\f}(\db_\star)|\hat{\s}) = \mathcal{N}(\hat{\f}(\db_\star)\ | \ \hat{\mub}_\star, \hat{\Sig}_\star)
		\end{align}
		where $\hat{\mub}_\star = \k_e^T(\K+\Sig)^{-1}(\hat{\s}-\bar{\mub}) + \mub(\db_\star) $ and the covariance $\hat{\Sig}_\star = \boldsymbol{\kappa}(\db_\star,\db_\star) - \k_\star^T(\K+\Sig)^{-1}\k_\star$.
		It is remarked that while the computational complexity of calculating the required inverse matrix $(\K+\Sig)^{-1}$ is $\mathcal{O}(n^3)$, it is required to be computed only once at the end of the training phase. The full CGP algorithm is summarized in Algorithm  \ref{algo3}.
		
		\begin{figure*}
			\subfigure[X]{\includegraphics[width=0.31\linewidth]{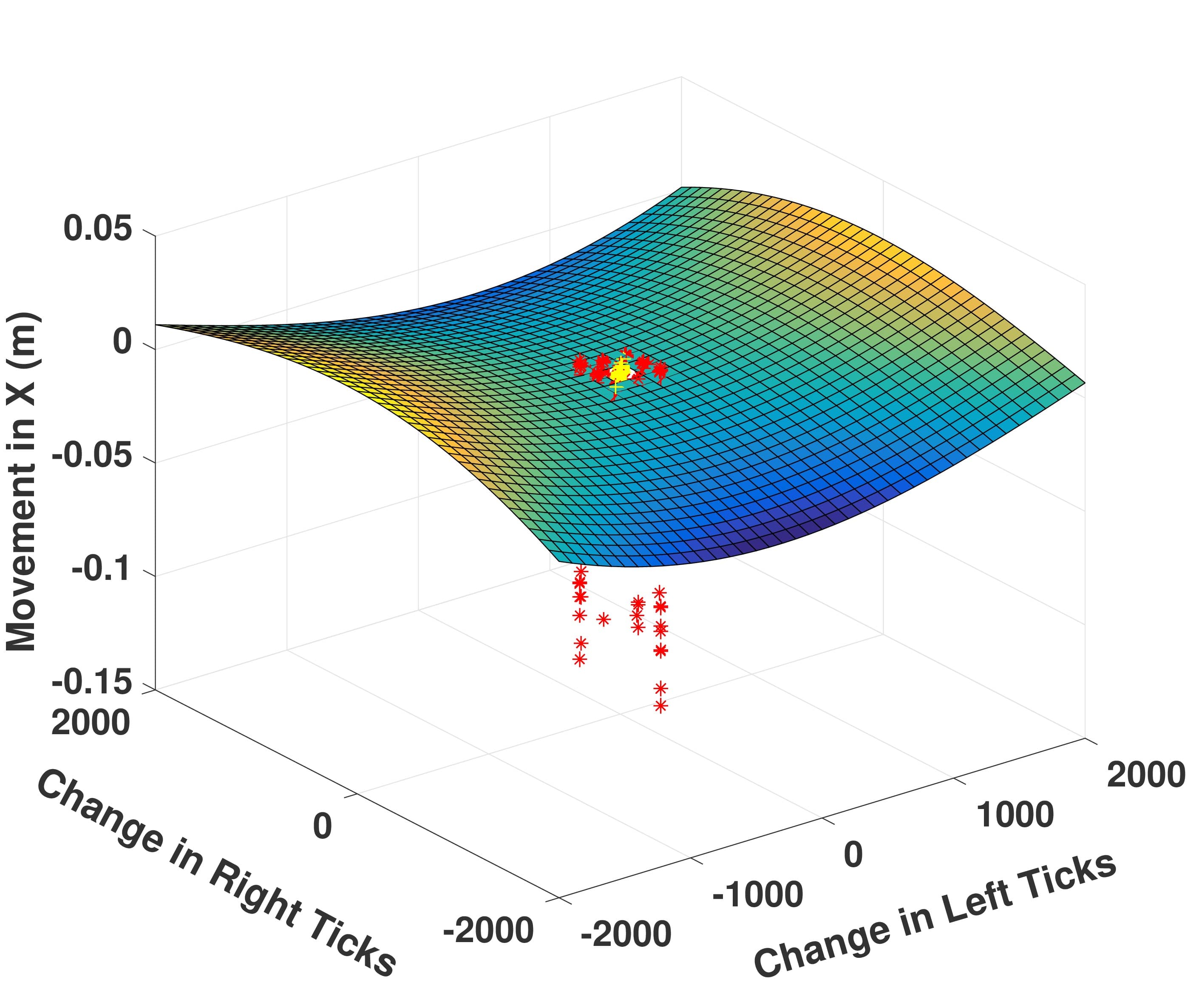}}
			\subfigure[Y]{\includegraphics[width=0.31\linewidth]{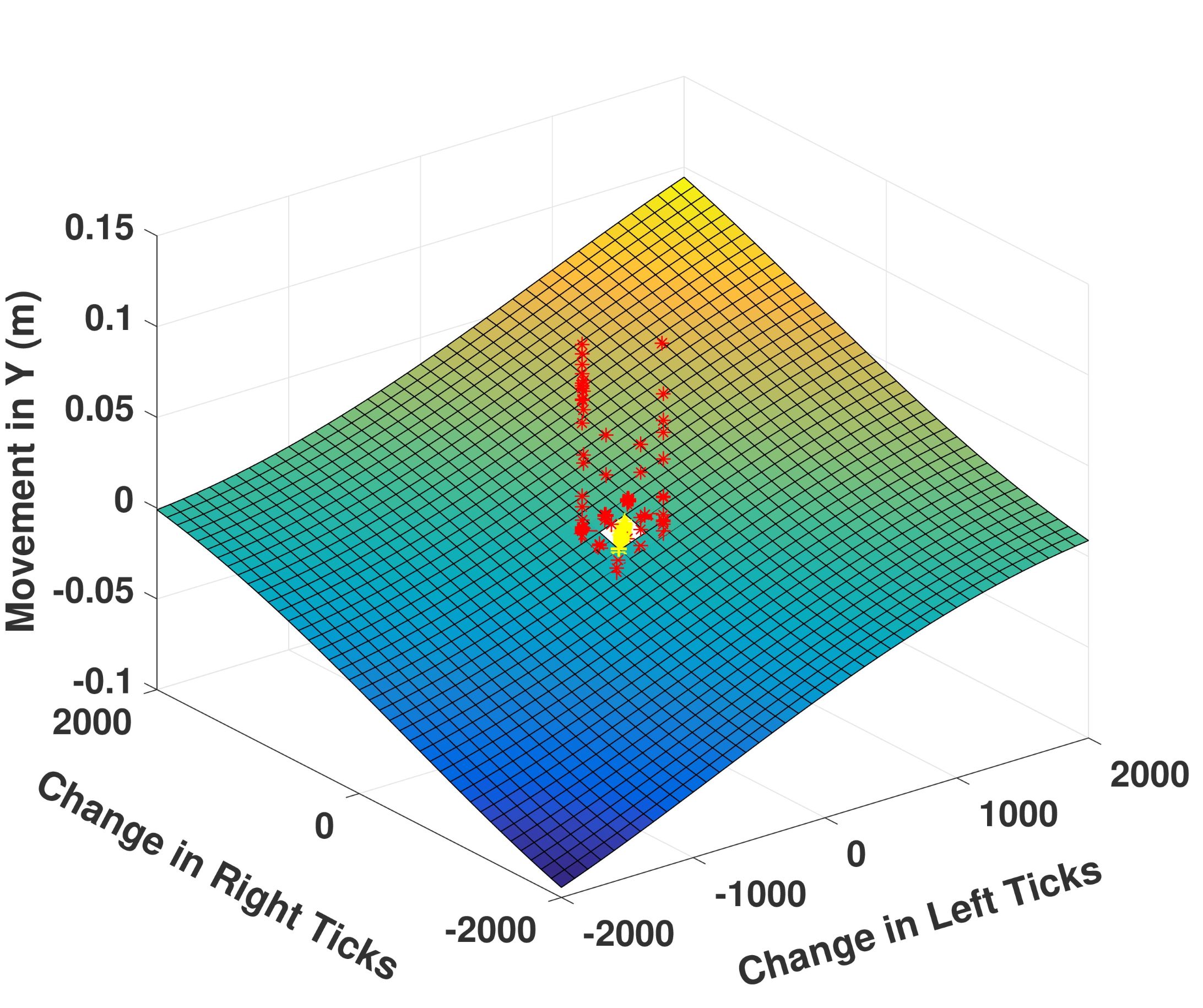}}
			\subfigure[Theta]{\includegraphics[width=0.31\linewidth]{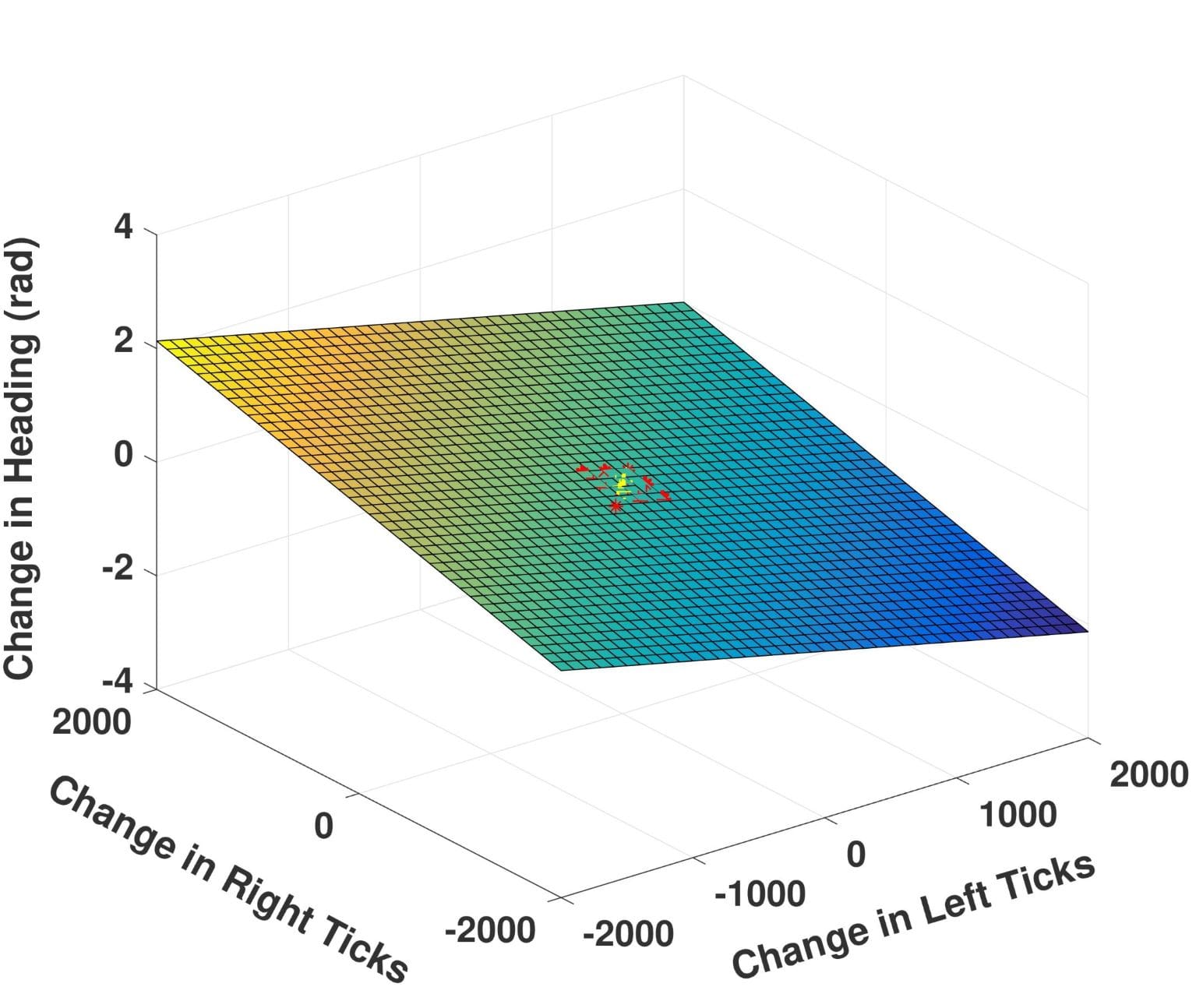}}
			\subfigure[X]{\includegraphics[width=0.33\linewidth]{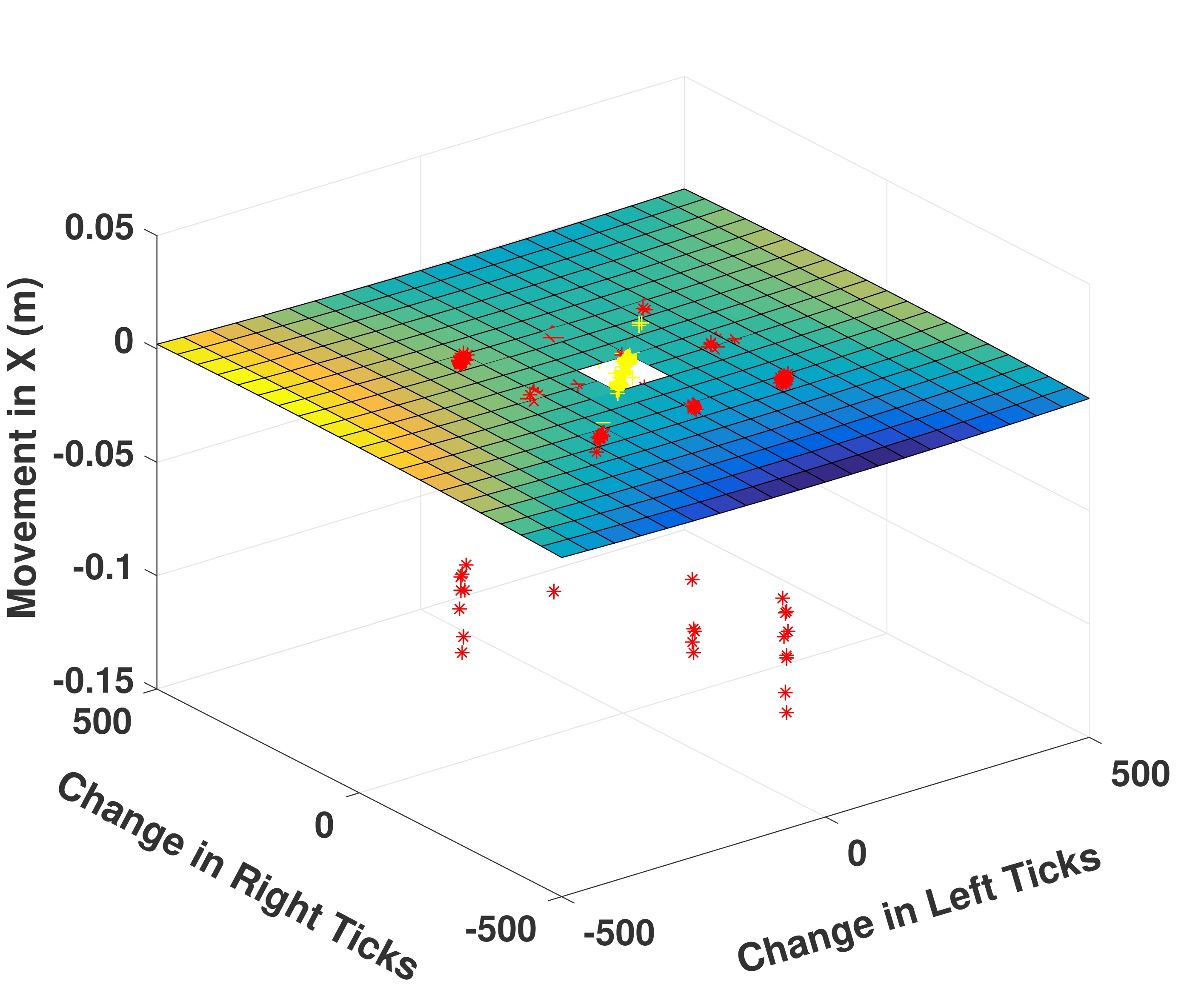}}
			\subfigure[Y]{\includegraphics[width=0.33\linewidth]{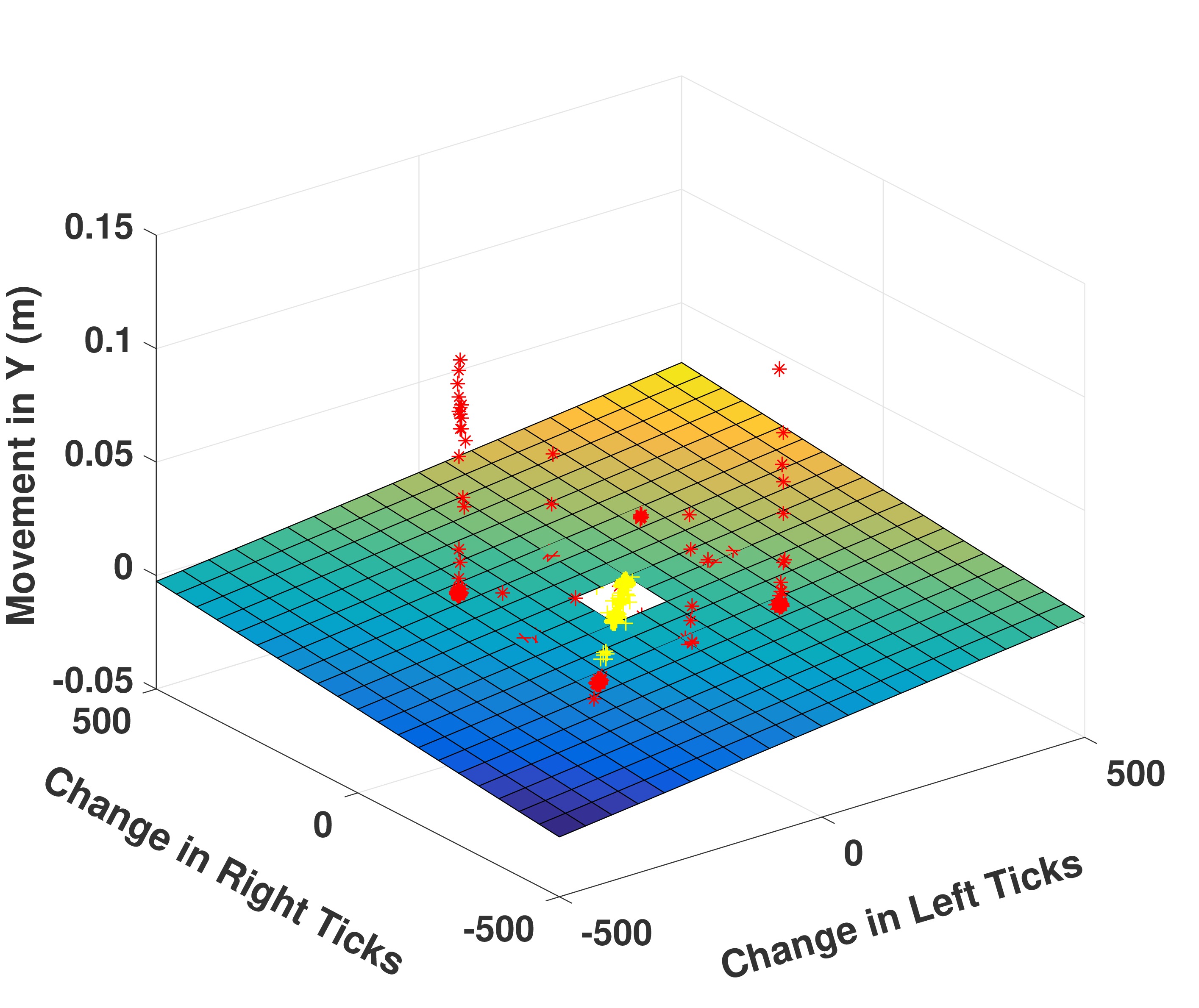}}
			\subfigure[Theta]{\includegraphics[width=0.33\linewidth]{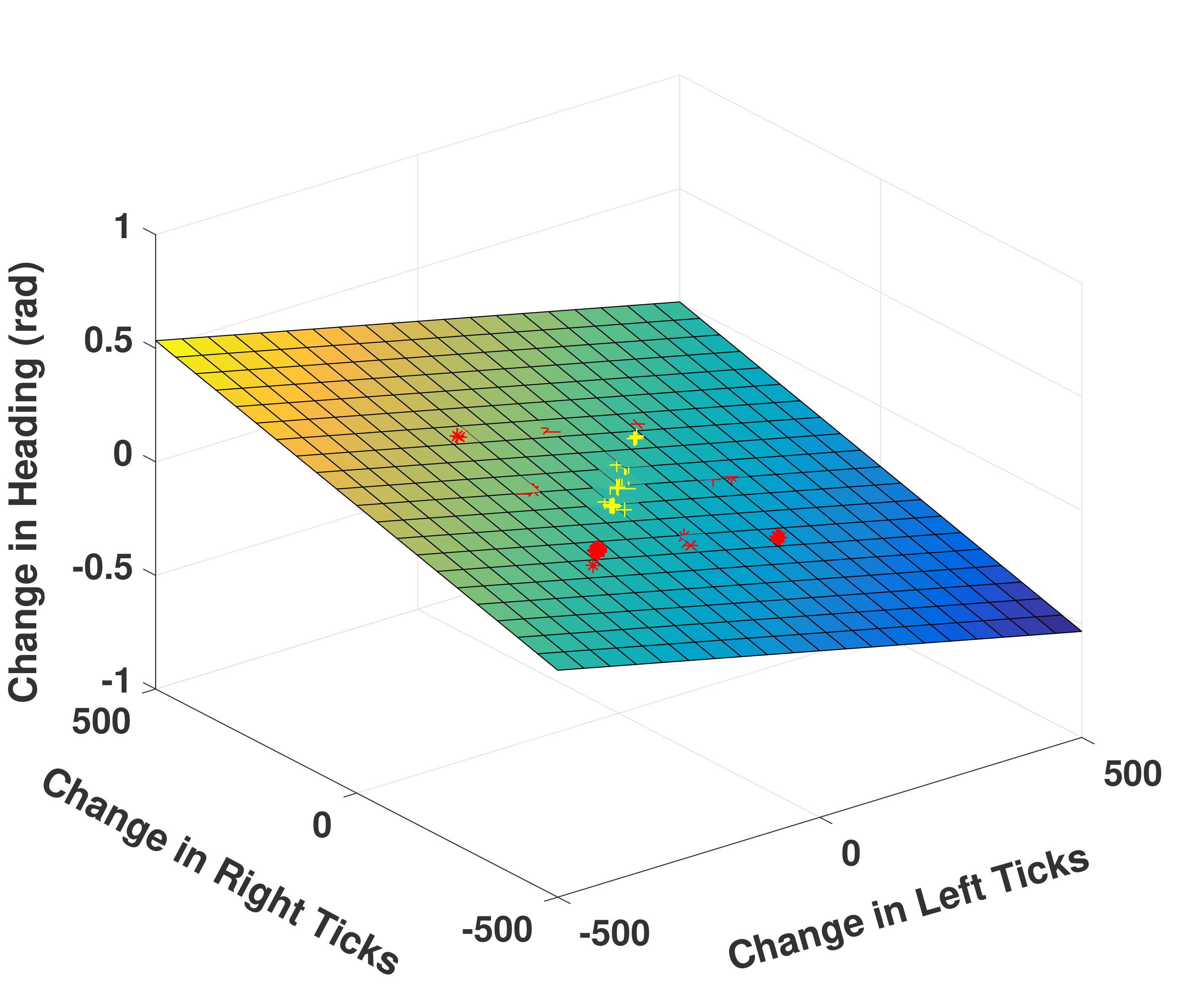}}
			\captionsetup{font=scriptsize}
			\caption{ (a),(b),(c) illustrates the movement of the sensor frame in $x,y,\theta$, respectively w.r.t change in left and right wheel ticks of a two wheel differential drive robot (i.e., $\f(\bullet) = \g(\bullet \ ;\ \hat{\p})$), overlaid with the corresponding sensor displacement measurements generated using raw data published at \cite{web} for a particular configuration. Note $\hat{\p}$ denotes parameter estimates found using CMLE \cite{censi}. Also in \cite{web}, since data from each configuration is divided into three subsets, we consider any two of them as training data and rest as test data. Points that are displayed in red and yellow color denote training and testing samples generated at selected scan instants respectively. Note: red points that are away from the 3D surface are outliers. (d)-(f) represent the truncated and enlarged versions of the same plots to expose the linearity.} \label{fig2}
		\end{figure*}
		
		\begin{algorithm} 
			\caption{CGP algorithm for simultaneous calibration of robot and sensor parameters} 
			\label{algo3}
			\begin{algorithmic}[1]
				\State  Collect measurements from sensors. 
				\State \textbf{Training Phase :}
				\State  Run sensor displacement algorithm for each selected interval, to get the estimates  $\{\hat{\s}^k\}$ and store them along with the corresponding angular velocities.
				\State Now pre-compute the following quantities :  
				\State \hspace{1cm}$(\boldsymbol{K}+\boldsymbol{\Sigma})^{-1}(\hat{\s}-\bar{\mub}) $ and $(\boldsymbol{K}+\boldsymbol{\Sigma})^{-1}$    
				\State \textbf{Testing Phase :}
				\State For every test input $\db_\star$, evaluate the following,    
				\State \hspace{1cm} $\hat{\mub}_{\star}  = \boldsymbol{k}_\star(\boldsymbol{K}+\boldsymbol{\Sigma})^{-1}(\hat{\s}-\bar{\mub}) + \mub(\db_\star)$ 
				\State \hspace{1cm} $\hat{\boldsymbol{\Sigma}}_\star = \boldsymbol{\kappa}(\db_\star,\db_\star) - \k_\star^T(\K+\Sig)^{-1}\k_\star $        
				\State Report $\hat{\s}_* $, where $p(\hat{\s}_{\star}) = \mathcal{N}(\hat{\s}_\star|\hat{\mub}_{\star},\hat{\boldsymbol{\Sigma}}_{\star})$
			\end{algorithmic}
		\end{algorithm}
		
		Note that in general, the choice of the mean and kernel functions is important and specific to the type of robot in use. In the present case, we use the linear mean function   
		\begin{equation}
		\mub(\textbf{x}) = \textbf{C} \textbf{x}
		\end{equation}
		where $\textbf{x} \in \Rn^m $ is the vector of wheel ticks recorded in a time interval and $\textbf{C} \in \Rn^{3\times m}$ is the associated hyper-parameter of the mean function. Recall that $m$ represents total number of wheels equipped with wheel encoders. Intuitively, the implication of this choice of linear mean function is that the relative position of the robot varies linearly with the wheel ticks recorded in the corresponding time interval. Such a relationship generally holds for arbitrary drive configurations if the time interval is sufficiently small.
		A widely used kernel function is the radial basis function as follows 
		\begin{equation}\label{covse}
		[\boldsymbol{\kappa}_{rbf}(\textbf{x},\textbf{x}')]_{i,i'} = \sigma_{i,i'}^2 \exp\left(-\frac{1}{2}(\textbf{x}-\textbf{x}')^T \textbf{B}_{i,i'}^{-1}(\textbf{x}-\textbf{x}')\right)
		\end{equation}
		where $\textbf{x},\textbf{x}'$ $\in \mathbb{R}^m$ are the data inputs with hyper-parameters $\Xi = [\sigma_{i,i'},\textbf{B}_{i,i'}]$, here $i,i' = 1,2,3$. It will be shown in section \ref{model_free} that for the two-wheel differential drive robot in use here, the squared exponential kernel \eqref{covse} with the linear mean function yielded better results than others. On the other hand for four-wheel Mecanum drive in use here, the inner product kernel, which amounts to a linear transformation of the feature space, 
		\begin{equation}
		[\boldsymbol{\kappa}_{lin}(\textbf{x},\textbf{x}')]_{i,i'} = \langle\ \textbf{x},\textbf{x}'\rangle
		\end{equation}
		performed better. We remark here that for our experiments we have assumed $\boldsymbol{\kappa}(\textbf{x},\textbf{x}') = diag([\boldsymbol{\kappa}(\textbf{x},\textbf{x}')]_{i,i})$. In general, the choice of the mean and kernel functions and that of the associated hyper-parameters is made a priori. For our experiments we infer the hyper-parameters by optimizing the corresponding log marginal likelihood. However, they may also be determined during the calibration phase via cross-validation.  
		
		\subsubsection{Calibration via approximate linear motion model}\label{linear_model} As an alternative to the general and flexible CGP approach that is applicable to any robot, we also put forth a computationally simple approach that relies on a linear approximation of $\f$. Specifically, if $\Delta t_{jk}$ is sufficiently small, so are elements of $\db_{jk}$. Therefore, it follows from the first order Taylor's series expansion, that $\f$ is approximately linear. This assertion if further verified empirically for the two-wheel differential drive. As evident from Fig. \ref{fig2}, for $\Delta t_{jk}$ sufficiently small, the elements of $\db_{jk}$ are concentrated around zero and the surface fitting them is indeed approximately linear. Motivated by the observation in Fig. \ref{fig2}, we let $\f(\db_{jk}) = \W\db_{jk}$, where $\W \in \Rn^{3 \times m}$ is the unknown weight matrix. The following robust linear regression problem can subsequently be solved to yield the weights:
		\begin{align}\label{rlr}
		\widehat{\W} = \arg\min_{\W} \sum_{(j,k)\in\E}\sum_{i\in\{x,y,\theta\}} \rho_c\left(\frac{\hat{\s}_{jk}^i - [\W\db_{jk}]_i}{\sigma_{jk}^i}\right) 
		\end{align}
		where $\rho_c$ is the Huber loss function \cite{huber}. Here, \eqref{rlr} is a convex optimization problem and can be solved efficiently with complexity $\O(n^3)$. Note that the entries of $\W$ do not have any physical significance and cannot generally be related to the intrinsic or extrinsic robot parameters, especially after  wheel deformation. Also, the computational cost incurred in making prediction is $\mathcal{O}(m)$ for the linear model but $\mathcal{O}(n^2)$ for the CGP algorithm.

		\section{Experiments on Autonomous Calibration of Mobile Robots} \label{expsec}
		This section details the experiments carried out to test the various calibration algorithms proposed in the paper. For all experiments, we made use of two different two-wheel differential drive and a four-wheel Mecanum drive robots, each equipped with wheel encoders and laser range finder. We begin with detailing the performance metrics used for evaluation followed by details regarding the experimental setup and results. 
		\subsection{Performance Metrics}
		The various calibration algorithms detailed in the paper output a robot/sensor motion model $\hat{\f}$, and the goal of this section is to evaluate the efficacy of the learned model. {In the absence of wheel slippages, it is remarked that the accuracy of motion model is quantified by the closeness of the robot trajectory estimate obtained from odometry to the ground truth trajectory.} Since even small errors in the model accumulate over time, the overall trajectory may deviate significantly over a longer interval. Therefore in practical settings, odometry data must generally be augmented or fused with data from other sensors. 
		\begin{figure}
			\centering  
			\subfigure{\includegraphics[width=0.85\linewidth,trim={0cm 0cm 0cm 0cm},clip]{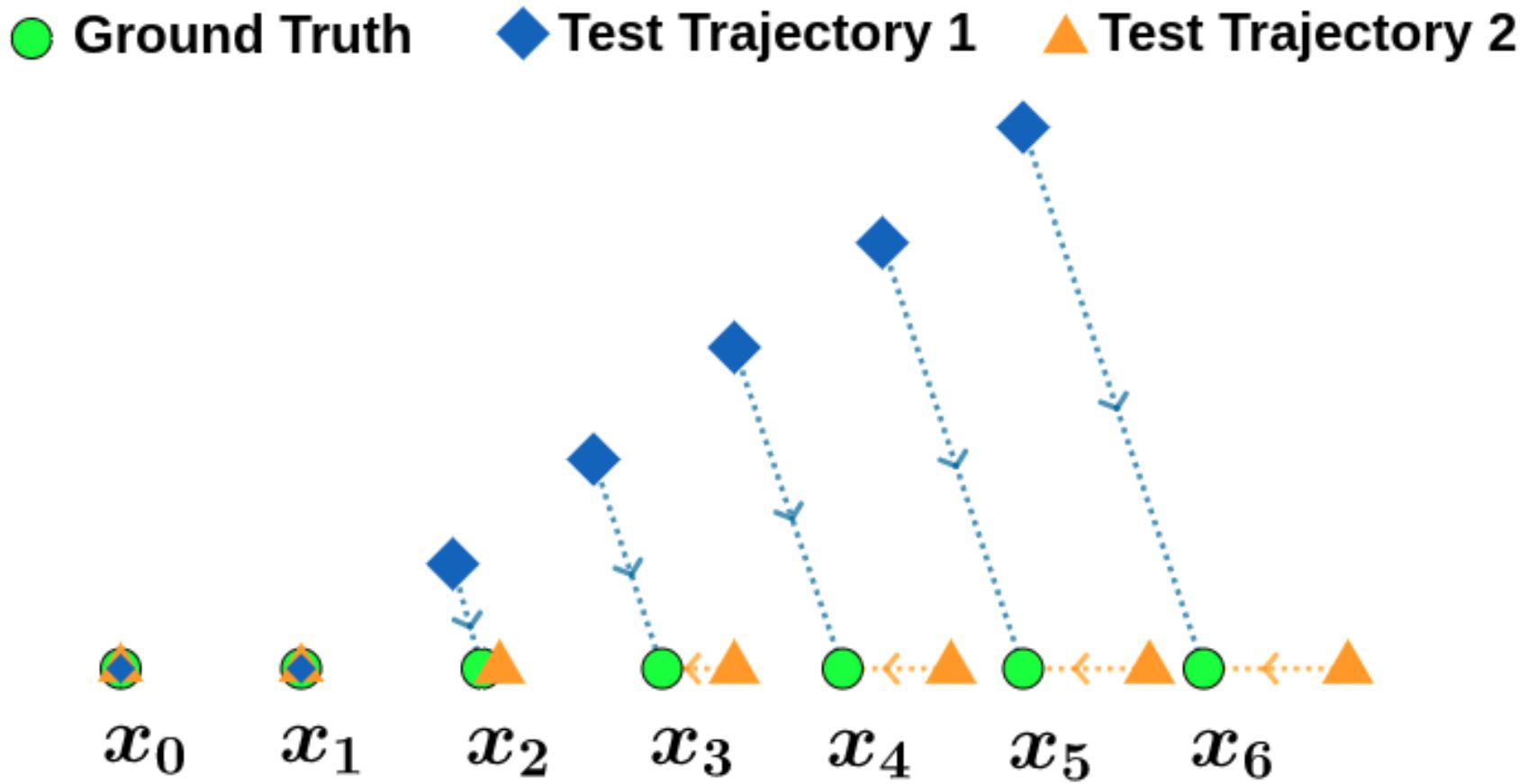}}
			\captionsetup{font=scriptsize} 
			\caption{An example scenario describing loose relation between the error metrics ATE \& RPE. $\left \{ \x_0,\x_1,\dots,\x_6 \right \}$ are the poses recorded at time instants $t \in \mathcal{T}=\left\{t_0,t_1,\dots t_6 \right\}$ respectively. Note that dotted lines with arrows represents correspondences of test trajectory poses with that of ground truth poses. } \label{fig_error}
		\end{figure}
		
		Since ground truth data was not available for the experiments, we instead used a SLAM algorithm to localize the sensor and build a map of the environment. While SLAM output would itself be inaccurate as compared to the ground truth, some of them \cite{slam} do not require odometry measurements and consequently serves as a benchmark for all calibration algorithms. Specifically, the \textit{google cartographer} algorithm, which leverages a robust scan to sub-map joining routine, is used for generating the trajectory and the map \cite{slam} of the environments. Note that in the absence of extrinsic calibration parameters, SLAM outputs only the sensor trajectory (and not the robot trajectory), which is subsequently used for comparisons. Further, estimating the sensor trajectory from odometry measurements requires the sensor motion model $\f$.

		Various sensor trajectory estimates are compared on the basis of Relative Pose Error (RPE)  and the Absolute Trajectory Error (ATE) motivated from \cite{benchmark}. The RPE measures the local accuracy of the trajectory, and is indicative of the drift in the estimated trajectory as compared to the ground truth. At any time $t_k \in \mathcal{T}$ ,  let the odometry and SLAM pose estimates be denoted by $\hat{\x}_k$ and $\x_k$, respectively. Then, relative pose change between times $t_k$ and $t_{k+1}$ estimated via odometry and SLAM are given by $\circleddash\ \hat{\x}_k \oplus \hat{\x}_{k+1}$ and $\circleddash\ \x_k \oplus \x_{k+1}$, respectively. Defining $\e^r_k := \circleddash\ (\circleddash\ \hat{\x}_k \oplus \hat{\x}_{k+1})\oplus (\circleddash\ \x_k \oplus \x_{k+1})$, the RPE is defined as the root mean square of the translational components of $\{\e^r_k\}_{k=1}^{n-1}$, i.e.,
		\begin{equation}
		\text{RPE} := \left ( \frac{1}{n-1} \sum\limits_{k=1}^{n-1} \left \|\text{trans}(\e^{r}_k)  \right \|^2 \right )^{1/2}
		\end{equation}
		where trans$(\textbf{e}_k)$ refers to the translational components of $\textbf{e}_k$. In contrast, the ATE measures the global (in)consistency of the estimated trajectory and is indicative of the absolute distance between the poses estimated by odometry and SLAM at any time $t_k$. Defining the absolute pose error at time $t_k$ as $\e^a_k:=\circleddash\ \hat{\x}_k \oplus \x_k$,  the ATE is evaluated as the root mean square of the pose errors for all times $t_k\in \mathcal{T}$, i.e., 
		\begin{equation}
		\text{ATE}  := \left ( \frac{1}{n} \sum\limits_{k=1}^{n}  \left \|\text{trans}(\e^a_k)  \right \|^2 \right )^{1/2}.
		\end{equation}
		We remark here that both relative rotational and translational errors in the robot trajectory contribute to the RPE. In contrast, the ATE only considers the absolute translational errors. A robot is said to be calibrated if the trajectory obtained from odometry is close to that obtained from SLAM under both the measures. Note that both the error metrics are loosely related to each other; in most cases, RPE and ATE can either be both low or high. However, situations exist when the same is not true. For instance, consider the case described in Fig.\ref{fig_error}, where two test trajectories are compared against ground truth. The RPE of Test trajectory 1 is caused due to relative rotation error occurring only in time segment $t_{12}:= t_2- t_1$. In contrast, test trajectory 2 incurs more RPE than that of test trajectory 1 due to low translational errors present between most of the time segments. Since the test trajectory 1 drifts more globally, it has a higher ATE than the test trajectory 2. This test case would be helpful in analyzing some of the results obtained in the later sub-sections.
		
		\subsection{Experimental conditions for autonomous calibration of wheeled robots} 
		Next, we detail the experimental setup used to test the different calibration algorithms.

		\subsubsection{Robots and sensors setup}  To perform experimental evaluations we used two two-wheel differential drive robots \emph{iClebo Kobuki} and \emph{FireBird VI} (see Fig.\ref{fig7}) and a four-wheel Mecanum drive \emph{Turtlebot3} robot (see Fig. \ref{meca}). Note that both \emph{iClebo Kobuki} and \emph{FireBird VI} robots have different sets of intrinsic parameters \cite{kobuki_web,firebird_web}. \emph{Kobuki} is a low-cost research robot with a diameter of 351.5 mm, weight of 2.35 kilograms, and a maximum translational velocity of 0.7 m/s. The wheel encoders provide data at the rate of 50 Hz and with a resolution of 2578.33 ticks per revolution. It is also equipped with RPLidar A1 2D laser scanner having $360^\circ$ field of view, with a detection range of 6 meters and a distance resolution less than 0.5 m and the operating frequency of 5.5 Hz. The \emph{Fire Bird VI} is primarily a research robot with diameter 280 mm, weight of 12 kilograms, and maximum translational velocity of 1.28 m/s. All Fire Bird encoders publish data at the rate of 10 Hz with a resolution of 3840 ticks per revolution. \emph{Turtlebot3 Mecanum} is from the \emph{Robotis} group with all wheels diameter of 60 mm. It weighs 1.8 kilograms, and maximum translational velocity is 0.26m/s. The dynamixels used, publish data at 10 Hz with an approximate resolution of 4096 ticks per revolution. Both \emph{Firebird} VI and \emph{Turtlebot3 Mecanum} robots were mounted with the RPLidar A2 laser scanner, which comes with adjustable frequency in the range of 5 to 15 Hz. In the experiments with RPLidar A2, the frequency of 10 Hz was used resulting in an angular resolution of 0.9$^\circ$, while the detection range and distance resolution were being same as those of RPlidar A1 laser scanner. 
		Note that, for all our experiments, we made use of an onboard computer (with an i5 processor of 8GB RAM, running ROS kinetic) for processing the data from laser range finder and wheel encoders, performing SLAM for validation, and running the calibration algorithms.
		
		To demonstrate the non-availability of the robot model, one of the wheels of the \emph{Firebird }VI robot is deformed with a thick tape (see Fig.\ref{fig_def}). Care was taken to ensure that the deformation was not too large, to avoid wobbling of the robot and the scan plane of the Lidar. In the case of  \emph{Turtlebot3} robot two different configurations are constructed (see Fig. \ref{meca}), by changing the position of the wheels from the regular configuration. We will see further that the amount of deformation in tilted wheel configuration (as in Fig. \ref{meca}(b)) is more as opposed to unaligned wheels configuration (as in Fig. \ref{meca}(a)).

		\subsubsection{Robot Configurations} Experiments, comprising of training and test phases, were carried out for various configurations of the setup used. It is remarked that test data for all configurations are collected for both, short and longer trajectories with simple and varied robot motions. Each experiment is labelled for reference, with details provided as shown in Table \ref{table1}. For example, setting \textbf{A} refers to the experiment done using \emph{Kobuki} robot with configuration \textbf{K1}, where training and test data are collected in \emph{WSN} and \emph{MiPS} labs respectively.

		\subsubsection{Scan Matching}  Since the CAM algorithm is a one-step method as opposed to other techniques discussed in the paper, it does not require a scan matching algorithm to be run beforehand. To demonstrate other proposed approaches we used point-to-line ICP (PLICP) variant \cite{P2LICP} to estimate the sensor displacements $\hat{\s}_{jk}$. Moreover, ICP-like methods also output the corresponding covariance value in closed-form \cite{covariance} that can be used by the IRLS algorithm.
		
		\subsubsection{Data Processing} For the experiments, we ensure that scans are collected at times spaced $T$ seconds apart. The choice of $T$ is not trivial. For instance, choosing a small $T$ often makes the algorithm too sensitive to un-modelled effects arising due to synchronization of sensors, robot's dynamics. It is lucrative to choose far scan pairs as more information is captured about the parameters; however, both the scan matching output as well as the motion model become inaccurate when $T$ is large. For the experiments, we chose the largest value of $T$ that yielded a reliable scan matching output in the form of sensor motion, resulting in $T=0.7$ seconds for \emph{Kobuki}, $T = 0.3$ and $0.6$ seconds for the \emph{Firebird} VI and \emph{Turtlebot3 Mecanum} robots respectively. These values are chosen based on maximum wheel speeds such that slippages are minimized during experimentation. Note that since the odometry readings are acquired at a rate, higher than the scans, temporally closest odometry reading is associated with a given scan. With the chosen $T$ the robots would move a maximum displacement of 15cm in x and y and 8$^\circ$ in yaw, under such conditions PLICP achieves 99.51 $\%$ accuracy \cite{P2LICP}.
		\begin{table}
			\fontsize{6pt}{8pt}\selectfont
			\captionsetup{font=scriptsize}
			\caption{ List of experimental configurations with labels and corresponding locations at which training and test data are collected } \label{table1}
			\centering
			\begin{tabular}{|c|c|c|c|c|}
				\hline
				\textbf{Setting} & \textbf{Robot}                       & \textbf{Configuration}       & \textbf{\begin{tabular}[c]{@{}c@{}}Training\\    Data\end{tabular}}                 & \textbf{Test Data}                                                               \\ \hline \hline
				\textbf{A}       & \multirow{4}{*}{\textit{Kobuki}}     & \multirow{2}{*}{\textbf{K1}} & \multirow{2}{*}{WSN Lab}                                                            & \multirow{2}{*}{MiPS Lab}                                                        \\
				\textbf{B}       &                                      &                              &                                                                                     &                                                                                  \\ \cline{1-1} \cline{3-5} 
				\textbf{C}       &                                      & \multirow{2}{*}{\textbf{K2}} & \multirow{2}{*}{\begin{tabular}[c]{@{}c@{}}WSN Lab + \\ ACES Corridor\end{tabular}} & \multirow{2}{*}{\begin{tabular}[c]{@{}c@{}}KD building\\ 3rd Floor\end{tabular}} \\
				\textbf{D}       &                                      &                              &                                                                                     &                                                                                  \\ \hline
				\textbf{E}       & \textit{Turtlebot3}                  & \textbf{M1}                  & \begin{tabular}[c]{@{}c@{}}Helicopter \\ Building\end{tabular}                      & \begin{tabular}[c]{@{}c@{}}Helicopter \\ Building\end{tabular}                   \\ \hline
				\textbf{F}       & \textit{FireBird VI}                 & \textbf{F1}                  & WSN Lab                                                                             & \begin{tabular}[c]{@{}c@{}}Tomography \\ Lab\end{tabular}                        \\ \hline
				\textbf{G}       & \multirow{2}{*}{\textit{Turtlebot3}} & \textbf{T1}                  & \multirow{2}{*}{\begin{tabular}[c]{@{}c@{}}ACES \\ Library\end{tabular}}            & \multirow{2}{*}{\begin{tabular}[c]{@{}c@{}}ACES \\ Library\end{tabular}}         \\
				\textbf{H}       &                                      & \textbf{T2}                  &                                                                                     &                                                                                  \\ \hline
			\end{tabular}
		\end{table}
		\subsection{Experimental Results}
		Here we present calibration results for both model-based and model-free scenarios. In model-based scenario calibration entails learning various parameters associated with the motion model of the sensor in terms of odometric data. On the other hand, model-free based algorithms involves learning the non-parametric motion model of the sensor. Note that the proposed CAM and CIRLS algorithms are model-based and CGP is model-free.
		\subsubsection{Results for model-based Calibration} We first compare the performance of the proposed CAM and CIRLS algorithms with CMLE \cite{censi} against ground truth over various configurations (see Table \ref{table1}) in the context of two-wheel differential drive. Recall that the proposed CAM algorithm is specifically designed for calibrating a two-wheel differential robot with CMLE \cite{censi} as its counterpart existing in the literature. On the other hand, the proposed CIRLS algorithm is applicable for calibrating robots with arbitrary but known drive configurations, henceforth subsumes the differential drive case. To this end, we ran CMLE \cite{censi} algorithm for number of iterations N = 4 and N = 16 discarding one percent of samples with higher order residual terms in each iteration, over configurations \textbf{K1} and \textbf{K2} respectively.  Table \ref{table2} displays the corresponding estimated calibration parameters. Specifically for CMLE, CIRLS and CIRLS CF,  3-sigma confidence intervals for the estimated parameters are also displayed. 
		Recall that CIRLS CF is a special case of CIRLS applicable to two-wheel differential drive where the optimization problem in step 7 of CIRLS (see Algorithm \ref{alg2}) is solved in closed forms. Note that the closed forms are derived following the similar approach as described in \cite{censi}.  
		For CAM, an analytical expression for the covariance of the estimated parameters was not available and hence not shown here. Observe that the extrinsic laser parameters are different for configurations \textbf{K1} and \textbf{K2} since the laser sensor is mounted distinctly concerning the robot frame of reference. However, the robot intrinsic parameter estimates are still consistent, as expected. We can notice that the parameter estimates for CAM are slightly different than that of those estimated from CMLE, CIRLS, and CIRLS CF. Nevertheless, since the actual values of parameters are not known, their correctness should only be evaluated via performance metrics (RPE and ATE). 
		
		\begin{table*}
			\fontsize{8pt}{10pt}\selectfont 
			\centering
			\setlength{\tabcolsep}{4pt}
			\captionsetup{font=scriptsize}
			\caption{ List of estimated parameters via CMLE\cite{censi}, CIRLS, CIRLS CF, and CAM for configurations  \textbf{K1}, \textbf{K2} } \label{table2}
			\begin{tabular}{|c|c|c|ccc|ccc|}
				\hline
				\multirow{2}{*}{\textbf{\begin{tabular}[c]{@{}c@{}}Robot\\ Config.\end{tabular}}} & \multirow{2}{*}{\textbf{Method}} & \multirow{2}{*}{\textbf{N}} & \multicolumn{6}{c|}{\textbf{Estimated Parameters}} \\ \cline{4-9} 
				&  &  & \multicolumn{1}{c|}{$\hat{r}_L$ (mm)} & \multicolumn{1}{c|}{$\hat{r}_R$ (mm)} & \multicolumn{1}{c|}{$\hat{b}$ (mm)} & \multicolumn{1}{c|}{$\hat{l}_x$ (mm)} & \multicolumn{1}{c|}{$\hat{l}_y$ (mm)} & $\hat{l}_{\theta}$ (rad) \\ \hline \hline
				\multirow{4}{*}{\textbf{K1}} & CMLE\cite{censi} & 4 & \begin{tabular}[c]{@{}c@{}}35.04 $\vspace{-1mm}$ $\pm$ 0.22\end{tabular} & \begin{tabular}[c]{@{}c@{}}35.16 $\vspace{-1mm}$ $\pm$ 0.22\end{tabular} & \begin{tabular}[c]{@{}c@{}}238.42 $\vspace{-1mm}$ $\pm$ 1.84\end{tabular} & \begin{tabular}[c]{@{}c@{}}19.92 $\vspace{-1mm}$ $\pm$ 1.54\end{tabular} & \begin{tabular}[c]{@{}c@{}}47.01 $\vspace{-1mm}$ $\pm$ 2.89\end{tabular} & \begin{tabular}[c]{@{}c@{}}3.13 $\vspace{-1mm}$ $\pm $0.01\end{tabular} \\ \cline{4-9} 
				& CAM & - & 35.94 & 35.99 & 241.00 & 12.34 & 53.52 & 3.13 \\ \cline{4-9} 
				& CIRLS & - & \begin{tabular}[c]{@{}c@{}}35.16 $\vspace{-1mm}$ $\pm$ 0.19\end{tabular} & \begin{tabular}[c]{@{}c@{}}35.18$\vspace{-1mm}$  $\pm$ 0.19\end{tabular} & \begin{tabular}[c]{@{}c@{}}238.38 $\vspace{-1mm}$ $\pm$ 1.38\end{tabular} & \begin{tabular}[c]{@{}c@{}}19.81 $\vspace{-1mm}$ $\pm$ 2.38\end{tabular} & \begin{tabular}[c]{@{}c@{}}45.85 $\vspace{-1mm}$ $\pm$ 2.39\end{tabular} & \begin{tabular}[c]{@{}c@{}}3.13 $\vspace{-1mm}$  $\pm$ 0.01\end{tabular} \\ \cline{4-9} 
				& CIRLS CF & - & \begin{tabular}[c]{@{}c@{}}35.09 $\vspace{-1mm}$ $\pm$ 0.21\end{tabular} & \begin{tabular}[c]{@{}c@{}}35.12$\vspace{-1mm}$  $\pm$ 0.21\end{tabular} & \begin{tabular}[c]{@{}c@{}}237.76 $\vspace{-1mm}$ $\pm$ 1.59\end{tabular} & \begin{tabular}[c]{@{}c@{}}19.80 $\vspace{-1mm}$ $\pm$ 2.60\end{tabular} & \begin{tabular}[c]{@{}c@{}}46.15 $\vspace{-1mm}$ $\pm$ 2.61\end{tabular} & \begin{tabular}[c]{@{}c@{}}3.13 $\vspace{-1mm}$ $\pm$ 0.01\end{tabular} \\ \cline{1-1} \cline{4-9} 
				\multirow{4}{*}{\textbf{K2}} & CMLE \cite{censi} & 16 & \begin{tabular}[c]{@{}c@{}}34.92 $\vspace{-1mm}$ $\pm$ 0.36\end{tabular} & \begin{tabular}[c]{@{}c@{}}34.94 $\vspace{-1mm}$ $\pm$ 0.36\end{tabular} & \begin{tabular}[c]{@{}c@{}}231.26 $\vspace{-1mm}$ $\pm$ 4.83\end{tabular} & \begin{tabular}[c]{@{}c@{}}2.36 $\vspace{-1mm}$ $\pm$ 7.81\end{tabular} & \begin{tabular}[c]{@{}c@{}}-119.73 $\vspace{-1mm}$ $\pm$ 8.74\end{tabular} & \begin{tabular}[c]{@{}c@{}}1.00 $\vspace{-1mm}$ $\pm$ 0.02\end{tabular} \\ \cline{4-9} 
				& CAM & - & 35.98 & 35.98 & 242.08 & -0.64 & -128.79 & 1.00 \\ \cline{4-9} 
				& CIRLS & - & \begin{tabular}[c]{@{}c@{}}34.88 $\vspace{-1mm}$ $\pm$ 0.25\end{tabular} & \begin{tabular}[c]{@{}c@{}}34.90 $\vspace{-1mm}$ $\pm$ 0.24\end{tabular} & \begin{tabular}[c]{@{}c@{}}231.92 $\vspace{-1mm}$ $\pm$ 1.71\end{tabular} & \begin{tabular}[c]{@{}c@{}}1.73 $\vspace{-1mm}$ $\pm$ 2.49\end{tabular} & \begin{tabular}[c]{@{}c@{}}-117.05$\vspace{-1mm}$  $\pm$ 2.33\end{tabular} & \begin{tabular}[c]{@{}c@{}}1.00 $\vspace{-1mm}$ $\pm$ 0.01\end{tabular} \\ \cline{4-9} 
				& CIRLS CF & - & \begin{tabular}[c]{@{}c@{}}34.96 $\vspace{-1mm}$ $\pm$ 0.23\end{tabular} & \begin{tabular}[c]{@{}c@{}}34.98 $\vspace{-1mm}$ $\pm$ 0.23\end{tabular} & \begin{tabular}[c]{@{}c@{}}232.96 $\vspace{-1mm}$ $\pm$ 1.63\end{tabular} & \begin{tabular}[c]{@{}c@{}}1.50 $\vspace{-1mm}$ $\pm$ 2.08\end{tabular} & \begin{tabular}[c]{@{}c@{}}-117.42 $\vspace{-1mm}$ $\pm$ 1.94\end{tabular} & \begin{tabular}[c]{@{}c@{}}1.00 $\vspace{-1mm}$ $\pm$ 0.01\end{tabular} \\ \hline
			\end{tabular}
		\end{table*}
		
\begin{figure}
			\centering  
			\subfigure[MiPS Lab, setting \textbf{A}]{\includegraphics[width=0.49\linewidth,trim={0cm 0cm 0cm 0cm},clip]{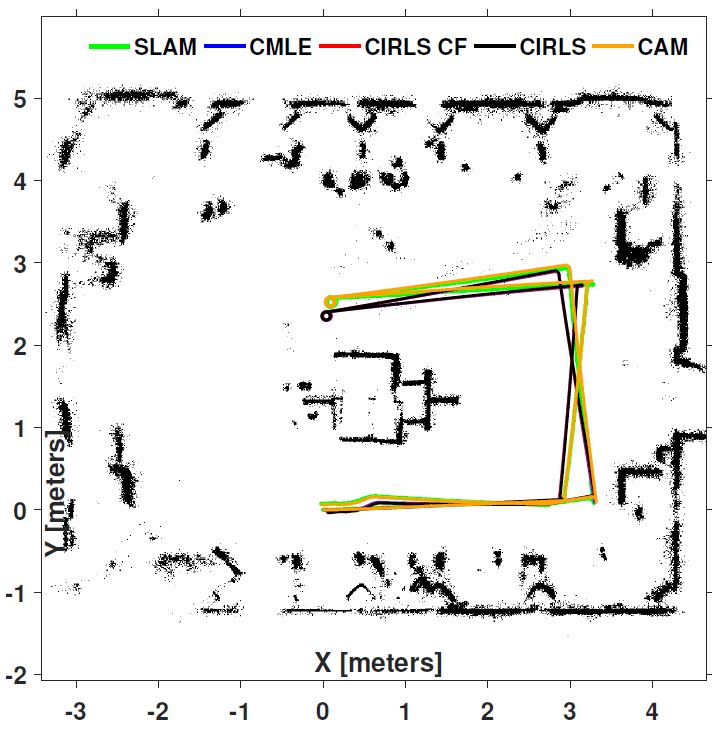}}
			\subfigure[MiPS Lab, setting \textbf{B}]{\includegraphics[width=0.49\linewidth,trim={0cm 0cm 0cm 0cm},clip ]{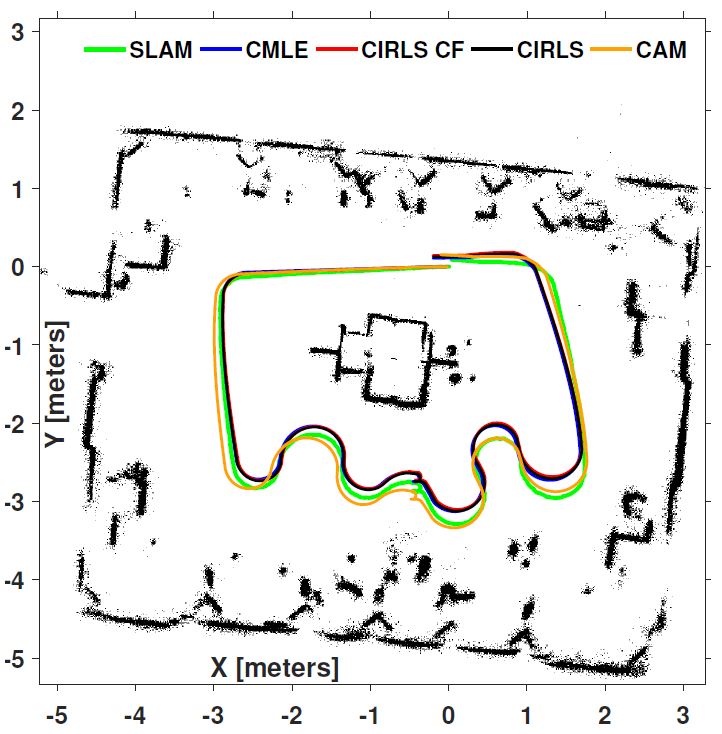}}
			\subfigure[4i Lab, setting \textbf{C}]{\includegraphics[width=0.49\linewidth,trim={2cm 1.5cm 2cm 1cm},clip]{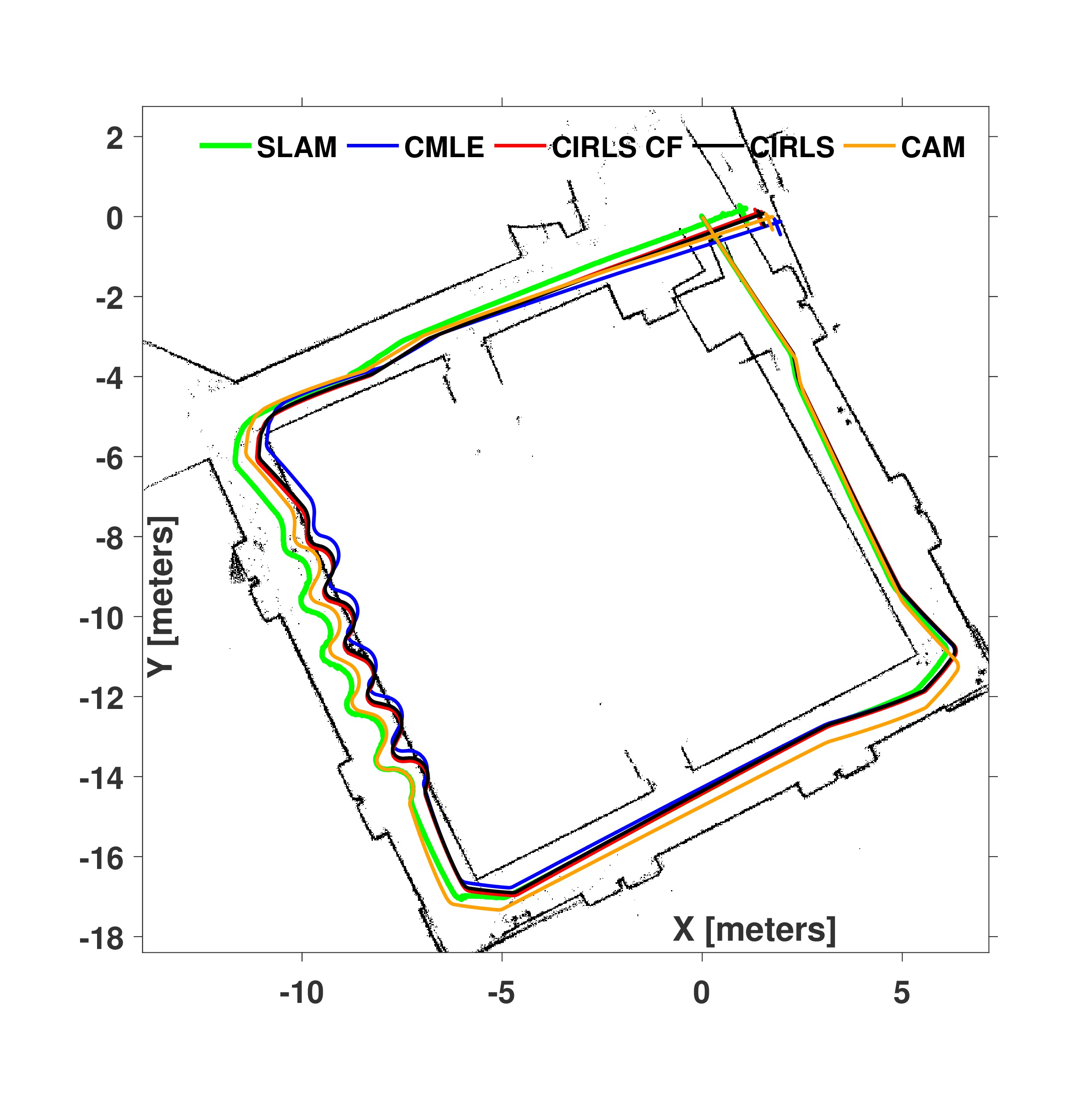}}
			\subfigure[KD building, setting \textbf{D}]{\includegraphics[width=0.49\linewidth,trim={2cm 1.5cm 2cm 1cm},clip]{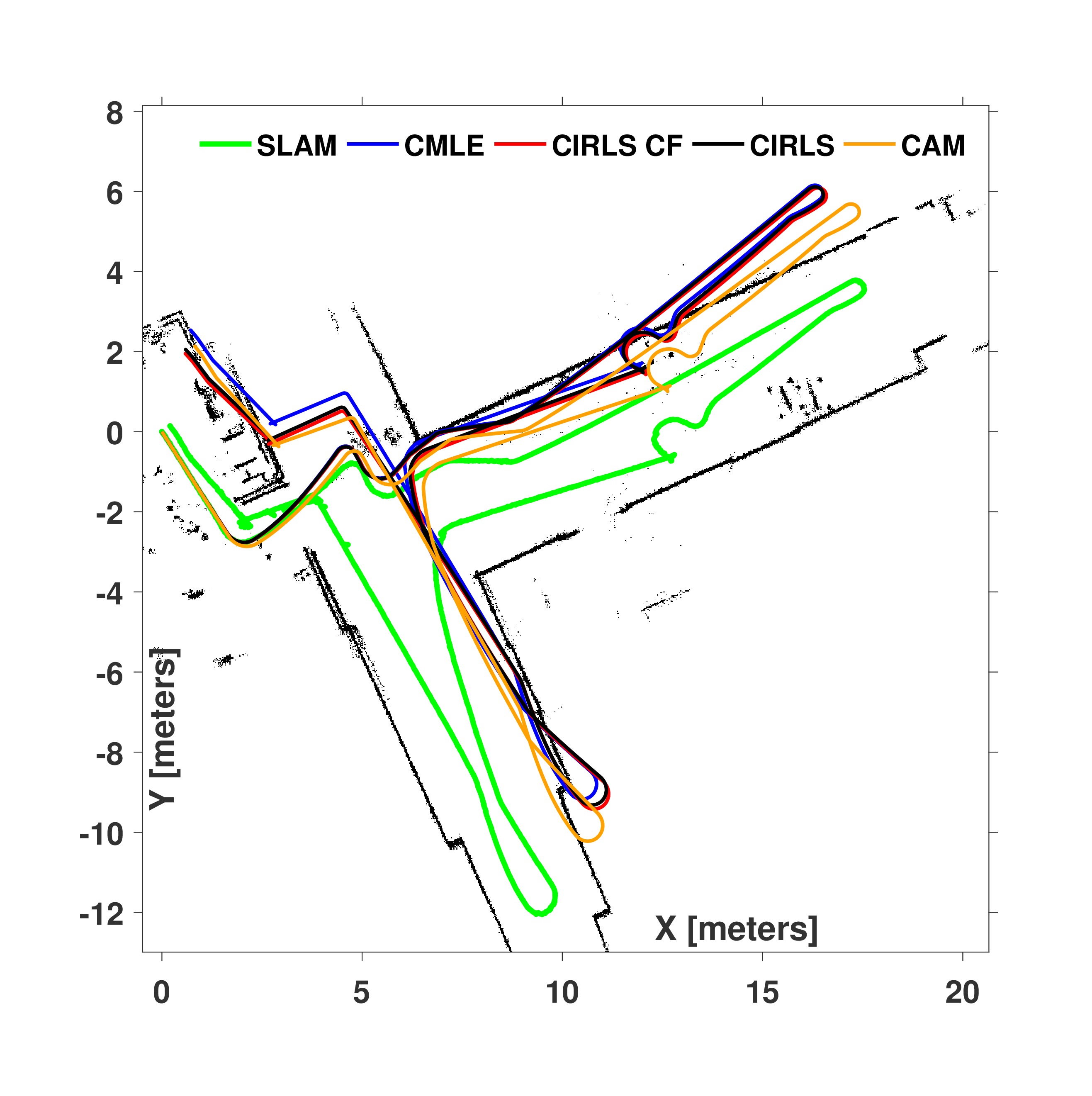}}
			\captionsetup{font=scriptsize}
			\caption{Trajectory comparisons against SLAM in various test environments. Here (a), (b) are the test environments for the configuration  {\textbf{K1}} where as (c), (d) are for the configuration  {\textbf{K2}}. (e) represents the test environment for configuration \textbf{M1}.}\label{fig3}
		\end{figure}

		To further compare the efficacy of the proposed algorithms, we generate the trajectory of the exteroceptive sensor on the corresponding test data using its inferred parametric motion model $\hat{\f}$. If the parameters are calibrated correctly, the generated trajectories should be close to the SLAM trajectory given the system is free from non-systematic errors like wheel slippages. Fig.\ref{fig3} displays the trajectories found using the parameters estimated from CAM, CIRLS, CIRLS CF, and CMLE \cite{censi} along with the map and trajectory generated using SLAM \cite{slam} for configurations \textbf{K1} and \textbf{K2}. Further, taking the SLAM trajectory as a reference, the RPE and ATE values for various algorithms are shown in Table \ref{table3}. From the table, it can be seen that almost all algorithms have similar RPE values, except for CAM for which they are slightly higher.
		\begin{table*}
			\fontsize{8pt}{10pt}\selectfont
			\centering
			\setlength{\tabcolsep}{3.3pt}
			\captionsetup{font=scriptsize}
			\caption{ Absolute trajectory error (ATE) and Relative pose error (RPE) for configurations \textbf{K1}, \textbf{K2} }\label{table3}
			\begin{tabular}{|c|cccc|cccc|}
				\hline
				\multirow{2}{*}{\textbf{Setting}} & \multicolumn{4}{c|}{\textbf{ATE} (cm)} & \multicolumn{4}{c|}{\textbf{RPE} (mm)} \\ \cline{2-9} 
				& \multicolumn{1}{c|}{CMLE \cite{censi}} & \multicolumn{1}{c|}{CIRLS} & \multicolumn{1}{c|}{CIRLS CF} & CAM & \multicolumn{1}{c|}{CMLE \cite{censi}} & \multicolumn{1}{c|}{CIRLS} & \multicolumn{1}{c|}{CIRLS CF} & CAM \\ \hline \hline
				\textbf{A} & 11.49 & 11.72 & 11.65 & 2.72 & 6.40 & 6.40 & 6.40 & 6.46 \\
				\textbf{B} & 16.79 & 17.23 & 18.71 & 10.13 & 7.63 & 7.64 & 7.63 & 7.72 \\
				\textbf{C} & 63.30 & 43.74 & 39.59 & 40.81 & 15.03 & 15.02 & 15.03 & 15.19 \\
				\textbf{D} & 217.84 & 202.30 & 196.21 & 162.69 & 27.63 & 27.62 & 27.63 & 27.75 \\ \hline
			\end{tabular}
		\end{table*}
		In contrast, however, the CAM algorithm exhibits the lowest ATE values suggesting that CAM parameters accurately predict relative heading of the robot over relative translation. It is to be noted that slight inaccuracy in relative heading estimates makes the global trajectory estimate drift heavily from ground truth over time. Interestingly, for settings \textbf{C} and \textbf{D} which contain outliers, the ATE of the CMLE approach \cite{censi} is relatively high. We remark here from Fig. \ref{fig5} that the CMLE algorithm incurs higher ATE values than CAM even for the optimal choice of $N$. Since all algorithms have similar and low RPE values, it can be concluded that the shape of the estimated trajectories is close to that of the ground truth. Recall that for testing purposes, the SLAM trajectory constitutes the ground truth. 
		\paragraph{Robustness to outliers}
		To further understand the improvements obtained from the proposed algorithms, we take a detailed look at the process of outlier removal in various methods. Unlike factory settings where a separate calibration phase and environment may be utilized, this paper advocates carrying out the calibration during the operational phase itself. As a result, however, outliers in the training phase are inevitable. To further test the effect of such outliers, we consider a corridor environment which is relatively featureless, resulting in a large number of scan matching failures that manifest themselves as outliers. Our experiments with configuration \textbf{K2} include such a scenario. Recall that the proposed algorithms are designed to handle outliers, thanks to the use of the Huber function as well as the trimming procedure (CAM) or the weight pruning algorithm (CIRLS). With the settings detailed in Sec. \ref{sec4}, no further parameter tuning is required, and the algorithm works well regardless of the number of outliers in the data. On the other hand, an iterative manual outlier rejection method is detailed in \cite{censi} amounting to extensive parameter tuning. The idea in \cite{censi} is to run the algorithm N iterations, while eliminating a fixed percentage $\alpha$ of outliers at every iteration. It is also required that the residual distribution should 'look' Gaussian, and the algorithm continues to run till that is the case. While it may be possible to utilize various statistical tests to discern the normality of a given distribution, the whole process is still manual and does not translate well to an automated calibration setting considered here. Indeed, it is evident from Table \ref{table4} that the choice of N is also critical, as the parameter estimates differ significantly over the range of N values. 
		\begin{table*}
			\centering
			\fontsize{8pt}{10pt}\selectfont
			\setlength{\tabcolsep}{6pt}
			\captionsetup{font=scriptsize}
			\caption{Variation in the parameter estimates found via CMLE algorithm \cite{censi}  for various values of N, for configuration \textbf{K2} } \label{table4}
			\begin{tabular}{|c|c|ccc|ccc|}
				\hline
				\multirow{2}{*}{\textbf{\begin{tabular}[c]{@{}c@{}}Robot\\ Config.\end{tabular}}} & \multirow{2}{*}{\textbf{N}} & \multicolumn{6}{c|}{\textbf{Estimated Parameters}} \\ \cline{3-8} 
				&  & \multicolumn{1}{c|}{$\hat{r}_L$ (mm)} & \multicolumn{1}{c|}{$\hat{r}_R$ (mm)} & \multicolumn{1}{c|}{$\hat{b}$ (mm)} & \multicolumn{1}{c|}{$\hat{l}_x$ (mm)} & \multicolumn{1}{c|}{$\hat{l}_y$ (mm)} & $\hat{l}_{\theta}$ (rad) \\ \hline \hline
				\multirow{7}{*}{\textbf{K2}} & 1 & \begin{tabular}[c]{@{}c@{}}28.56 $\vspace{-1mm}$ $\pm$ 0.27\end{tabular} & \begin{tabular}[c]{@{}c@{}}28.58$\vspace{-1mm}$ $\pm$ 0.27\end{tabular} & \begin{tabular}[c]{@{}c@{}}194.14$\vspace{-1mm}$ $\pm$ 3.76\end{tabular} & \begin{tabular}[c]{@{}c@{}}2.48 $\vspace{-1mm}$ $\pm$ 6.10\end{tabular} & \begin{tabular}[c]{@{}c@{}}-119.72 $\vspace{-1mm}$ $\pm$ 6.74\end{tabular} & \begin{tabular}[c]{@{}c@{}}1.01 $\vspace{-1mm}$ $\pm$ 0.02\end{tabular} \\ \cline{2-8} 
				& 7 & \begin{tabular}[c]{@{}c@{}}31.68 $\vspace{-1mm}$ $\pm$ 0.30\end{tabular} & \begin{tabular}[c]{@{}c@{}}31.68 $\vspace{-1mm}$ $\pm$ 0.30\end{tabular} & \begin{tabular}[c]{@{}c@{}}212.19 $\vspace{-1mm}$ $\pm$ 4.13\end{tabular} & \begin{tabular}[c]{@{}c@{}}2.59 $\vspace{-1mm}$ $\pm$ 6.67\end{tabular} & \begin{tabular}[c]{@{}c@{}}-120 $\vspace{-1mm}$ $\pm$ 7.40\end{tabular} & \begin{tabular}[c]{@{}c@{}}1.00 $\vspace{-1mm}$ $\pm$ 0.02\end{tabular} \\ \cline{2-8} 
				& 12 & \begin{tabular}[c]{@{}c@{}}34.13 $\vspace{-1mm}$ $\pm$ 0.34\end{tabular} & \begin{tabular}[c]{@{}c@{}}34.13 $\vspace{-1mm}$ $\pm$ 0.34\end{tabular} & \begin{tabular}[c]{@{}c@{}}226.59 $\vspace{-1mm}$ $\pm$ 4.59\end{tabular} & \begin{tabular}[c]{@{}c@{}}2.51 $\vspace{-1mm}$ $\pm$ 7.37\end{tabular} & \begin{tabular}[c]{@{}c@{}}-119.53 $\vspace{-1mm}$ $\pm$ 8.23\end{tabular} & \begin{tabular}[c]{@{}c@{}}1.00 $\vspace{-1mm}$ $\pm$ 0.02\end{tabular} \\ \cline{2-8} 
				& 16 & \begin{tabular}[c]{@{}c@{}}34.92 $\vspace{-1mm}$ $\pm$ 0.36\end{tabular} & \begin{tabular}[c]{@{}c@{}}34.94 $\vspace{-1mm}$ $\pm$ 0.36\end{tabular} & \begin{tabular}[c]{@{}c@{}}231.26 $\vspace{-1mm}$ $\pm$ 4.83\end{tabular} & \begin{tabular}[c]{@{}c@{}}2.36 $\vspace{-1mm}$ $\pm$ 7.81\end{tabular} & \begin{tabular}[c]{@{}c@{}}-119.73 $\vspace{-1mm}$ $\pm$ 8.74\end{tabular} & \begin{tabular}[c]{@{}c@{}}1.00 $\vspace{-1mm}$ $\pm$ 0.02\end{tabular} \\ \cline{2-8} 
				& 18 & \begin{tabular}[c]{@{}c@{}}35.06 $\vspace{-1mm}$ $\pm$ 0.37\end{tabular} & \begin{tabular}[c]{@{}c@{}}35.08 $\vspace{-1mm}$ $\pm$ 0.37\end{tabular} & \begin{tabular}[c]{@{}c@{}}232.04 $\vspace{-1mm}$ $\pm$ 4.95\end{tabular} & \begin{tabular}[c]{@{}c@{}}2.25 $\vspace{-1mm}$ $\pm$ 8.05\end{tabular} & \begin{tabular}[c]{@{}c@{}}-119.58 $\vspace{-1mm}$ $\pm$ 9.01\end{tabular} & \begin{tabular}[c]{@{}c@{}}1.00 $\vspace{-1mm}$ $\pm$ 0.02\end{tabular} \\ \cline{2-8} 
				& 20 & \begin{tabular}[c]{@{}c@{}}35.13 $\vspace{-1mm}$ $\pm$ 0.38\end{tabular} & \begin{tabular}[c]{@{}c@{}}35.14 $\vspace{-1mm}$ $\pm$ 0.38\end{tabular} & \begin{tabular}[c]{@{}c@{}}232.40 $\vspace{-1mm}$ $\pm$ 5.07\end{tabular} & \begin{tabular}[c]{@{}c@{}}2.17 $\vspace{-1mm}$ $\pm$ 8.24\end{tabular} & \begin{tabular}[c]{@{}c@{}}-119.83 $\vspace{-1mm}$ $\pm$ 9.21\end{tabular} & \begin{tabular}[c]{@{}c@{}}1.00 $\vspace{-1mm}$ $\pm$ 0.02\end{tabular} \\ \cline{2-8} 
				& 25 & \begin{tabular}[c]{@{}c@{}}35.27 $\vspace{-1mm}$ $\pm$ 0.40\end{tabular} & \begin{tabular}[c]{@{}c@{}}35.29 $\vspace{-1mm}$ $\pm$ 0.40\end{tabular} & \begin{tabular}[c]{@{}c@{}}233.35 $\vspace{-1mm}$ $\pm$ 5.35\end{tabular} & \begin{tabular}[c]{@{}c@{}}2.09 $\vspace{-1mm}$ $\pm$ 8.70\end{tabular} & \begin{tabular}[c]{@{}c@{}}-119.91 $\vspace{-1mm}$ $\pm$ 9.77\end{tabular} & \begin{tabular}[c]{@{}c@{}}1.00 $\vspace{-1mm}$ $\pm$ 0.02\end{tabular} \\ \hline
			\end{tabular}
		\end{table*}
		
		\begin{figure}
			\centering  
			\subfigure[]{\includegraphics[width=0.24\linewidth]{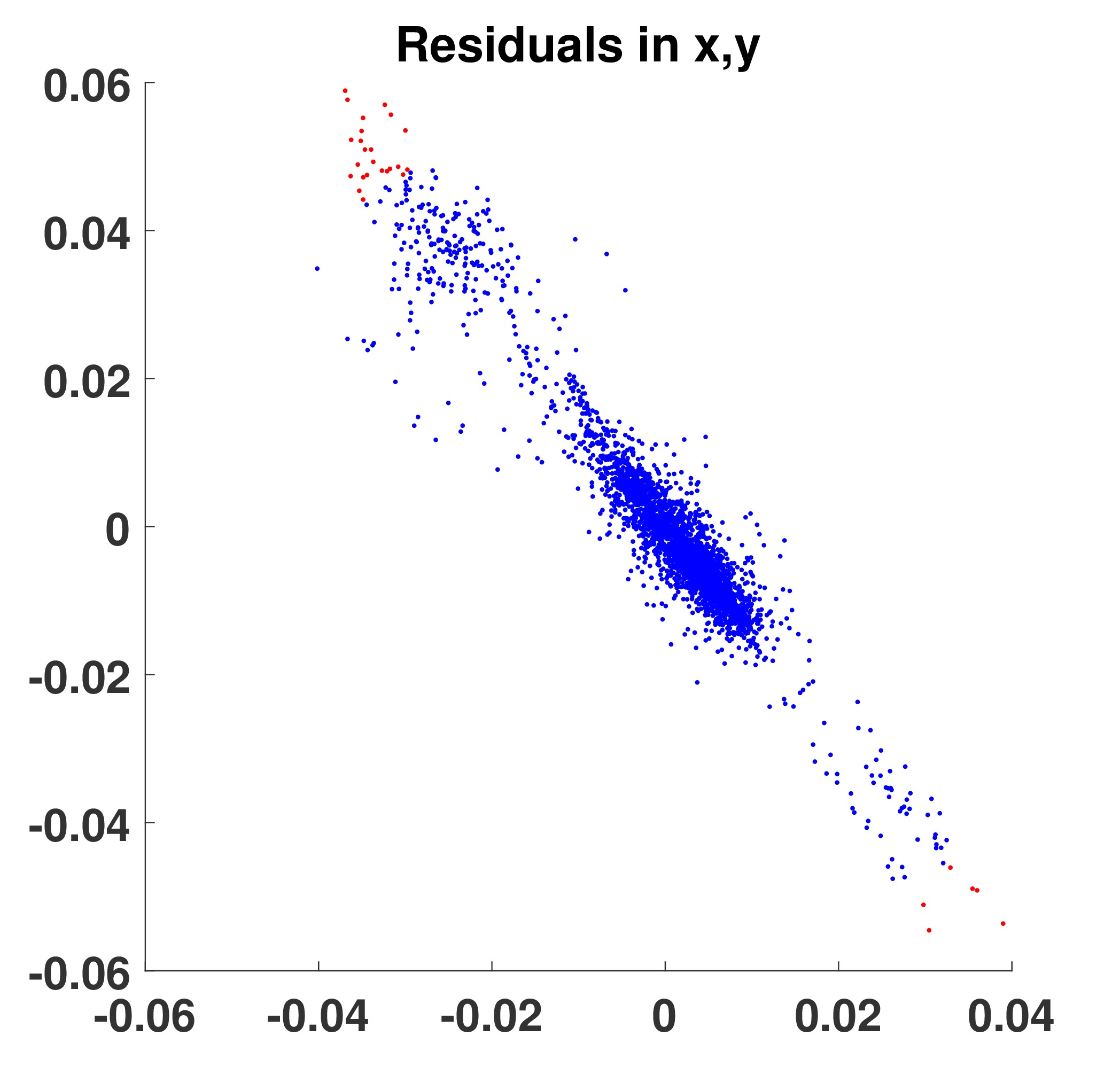} \includegraphics[width=0.24\linewidth]{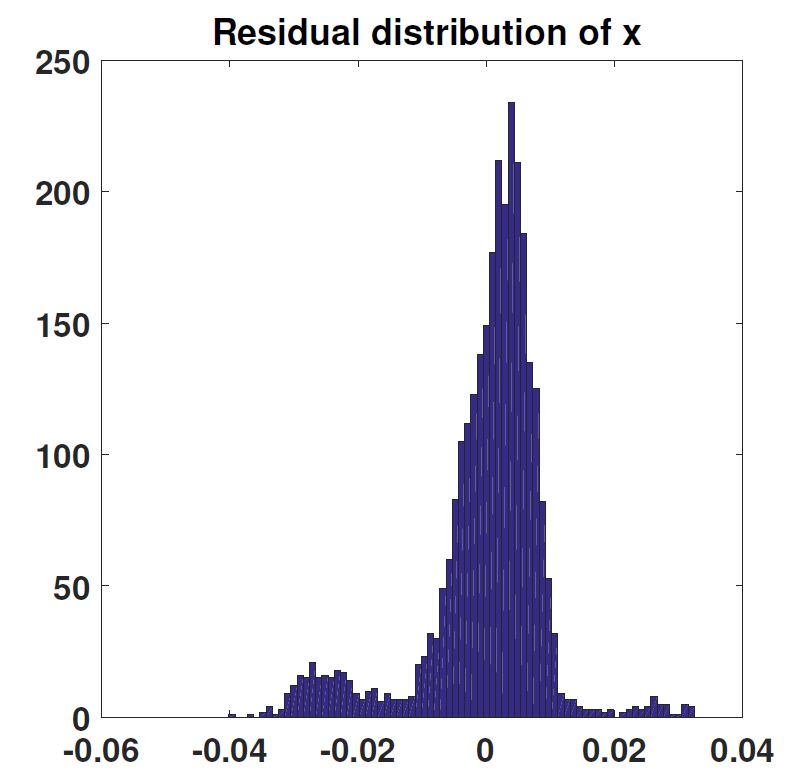}
				\includegraphics[width=0.24\linewidth]{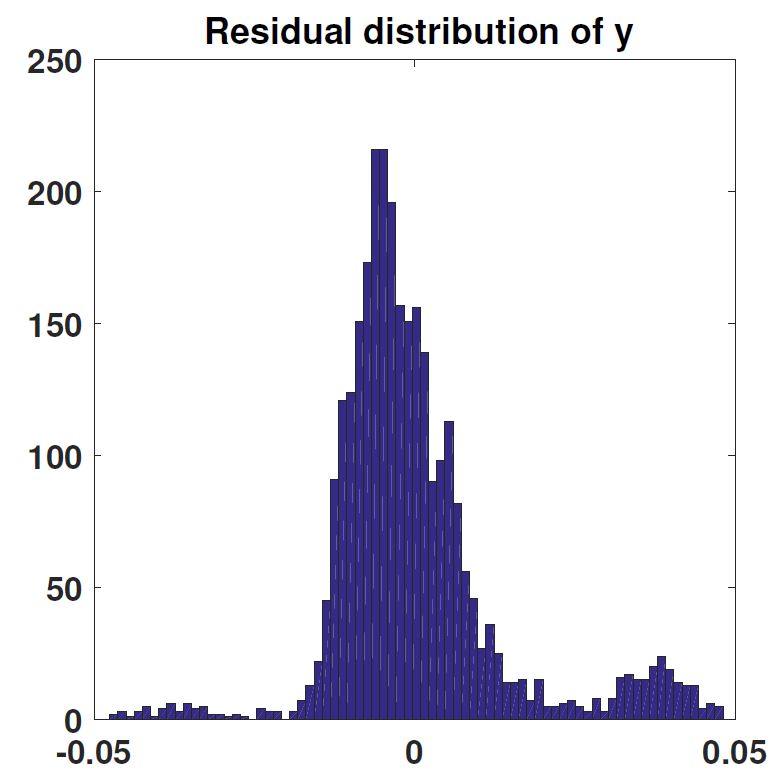}
				\includegraphics[width=0.24\linewidth]{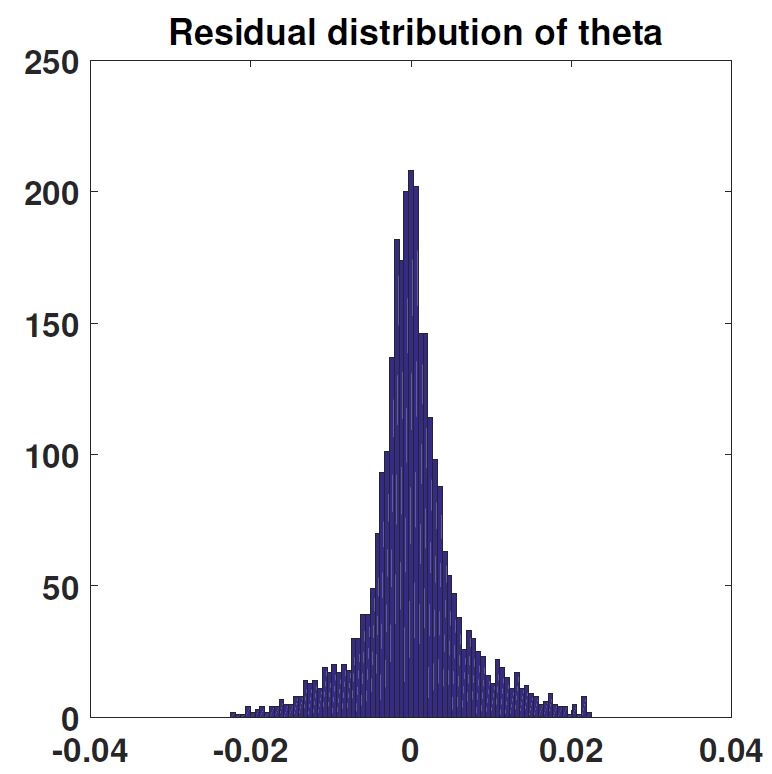}}
			
			\subfigure[]{\includegraphics[width=0.24\linewidth]{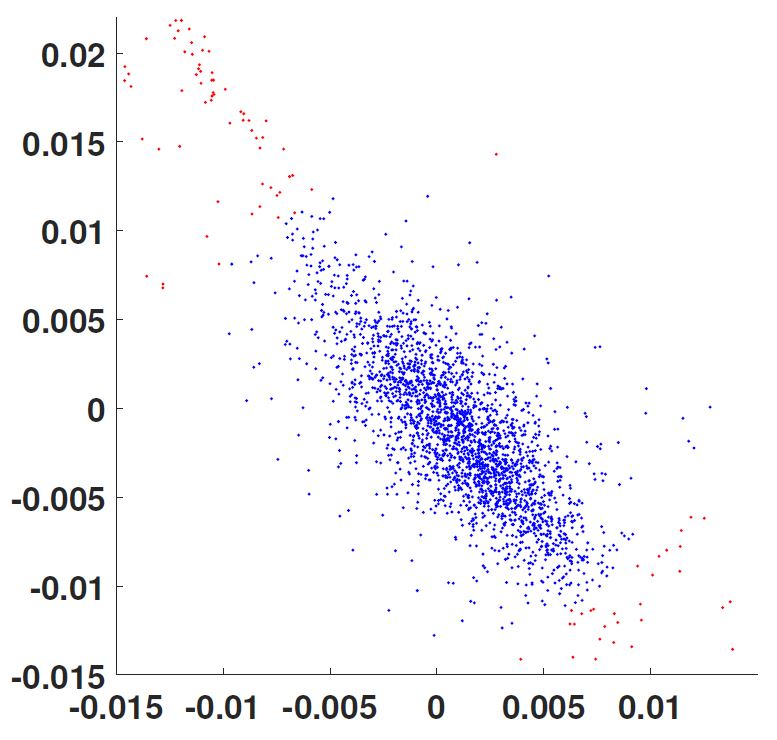}
				\includegraphics[width=0.24\linewidth]{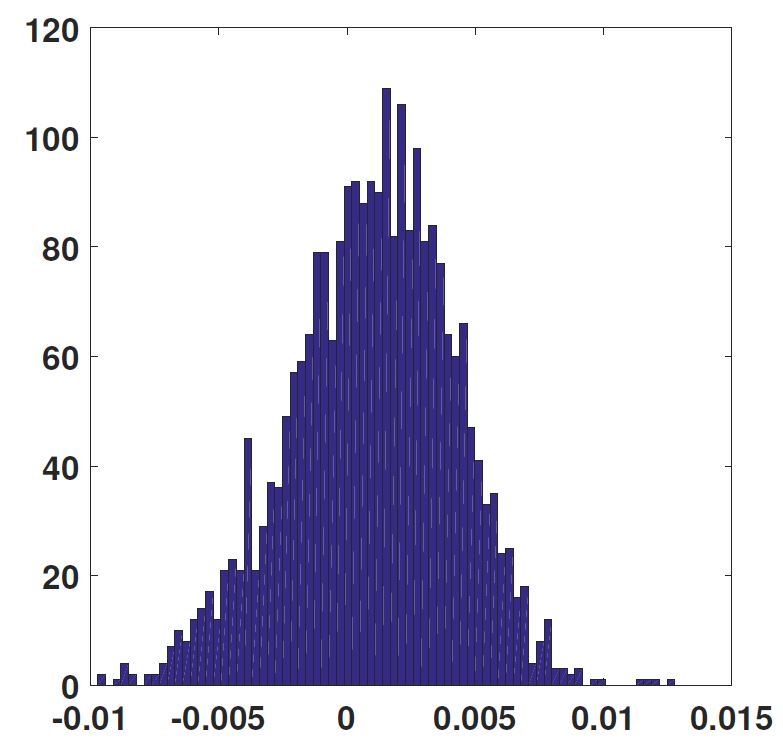}
				\includegraphics[width=0.24\linewidth]{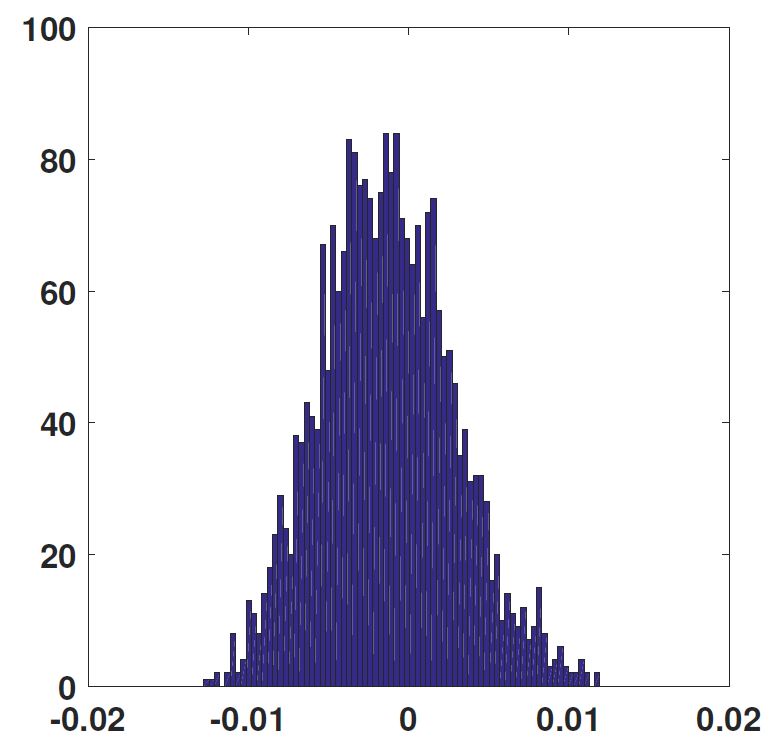}
				\includegraphics[width=0.24\linewidth]{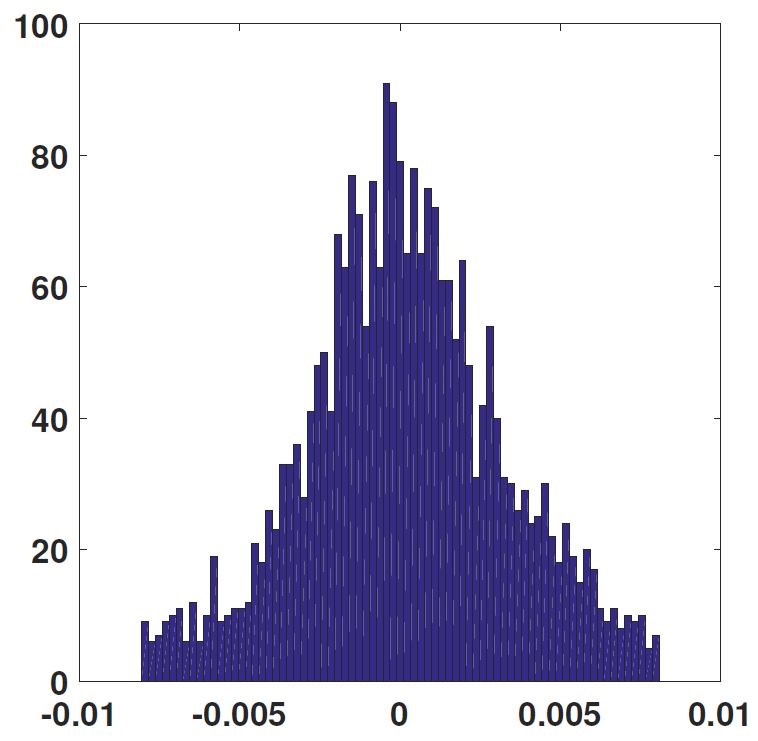}}
			
			\subfigure[]{\includegraphics[width=0.24\linewidth]{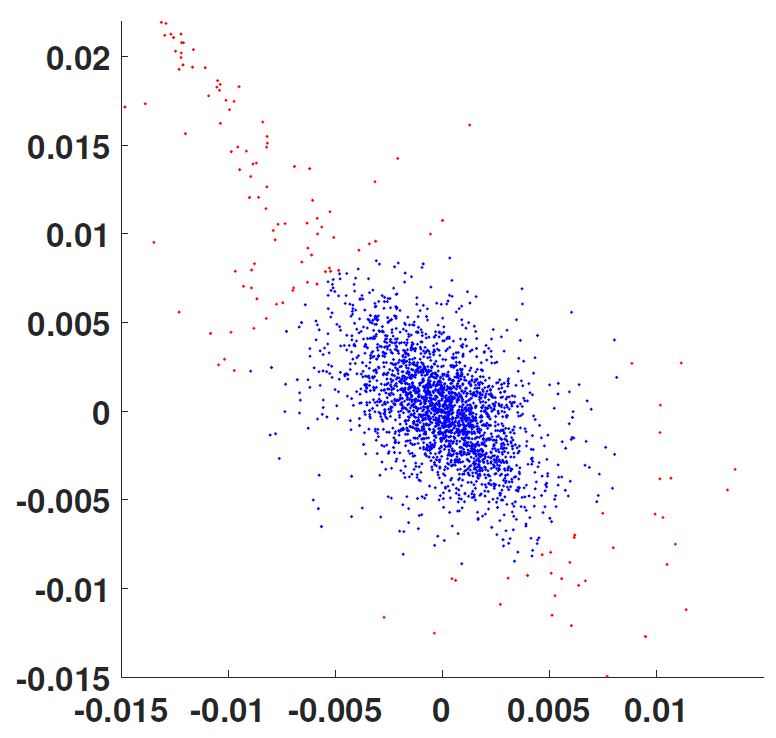}
				\includegraphics[width=0.24\linewidth]{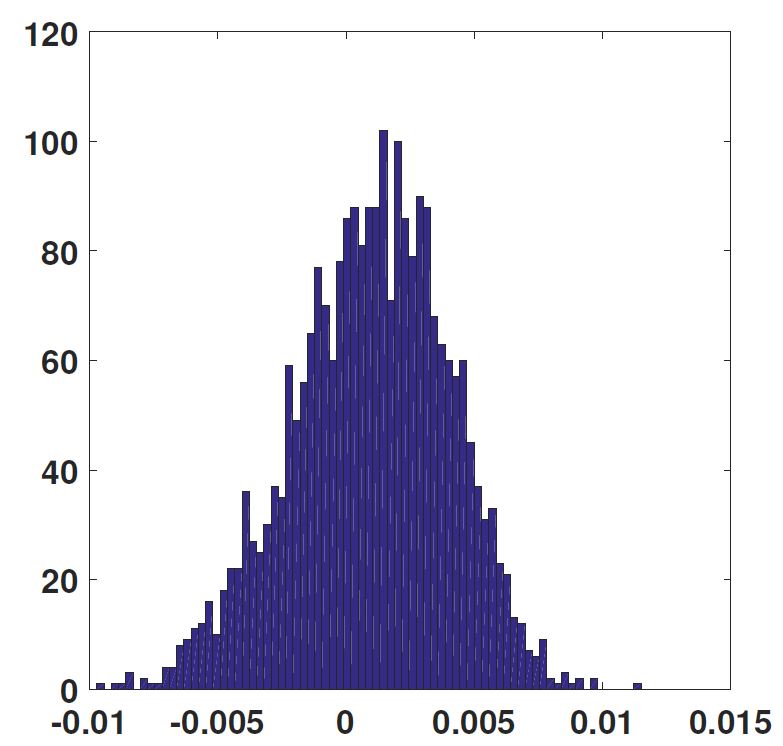}
				\includegraphics[width=0.24\linewidth]{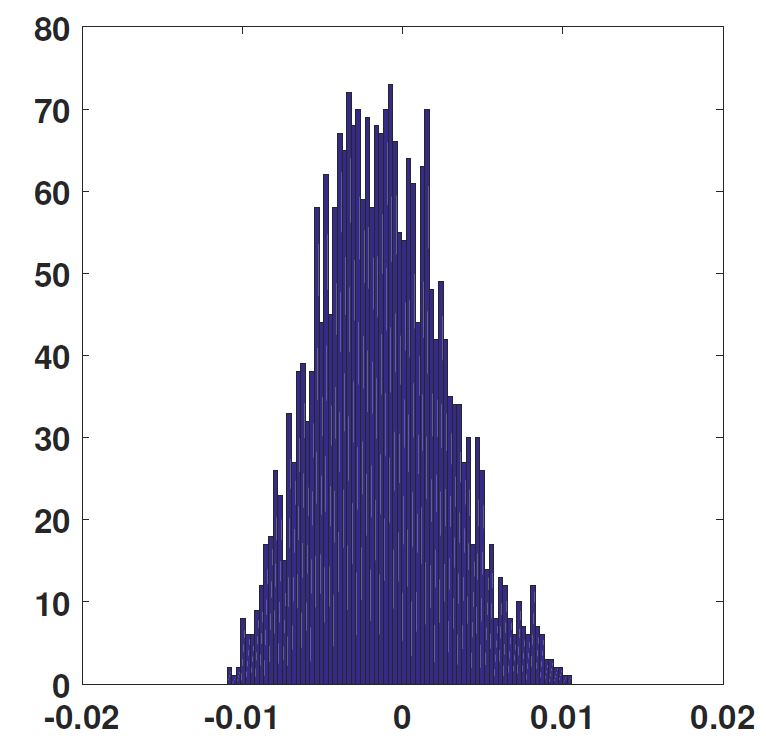}
				\includegraphics[width=0.24\linewidth]{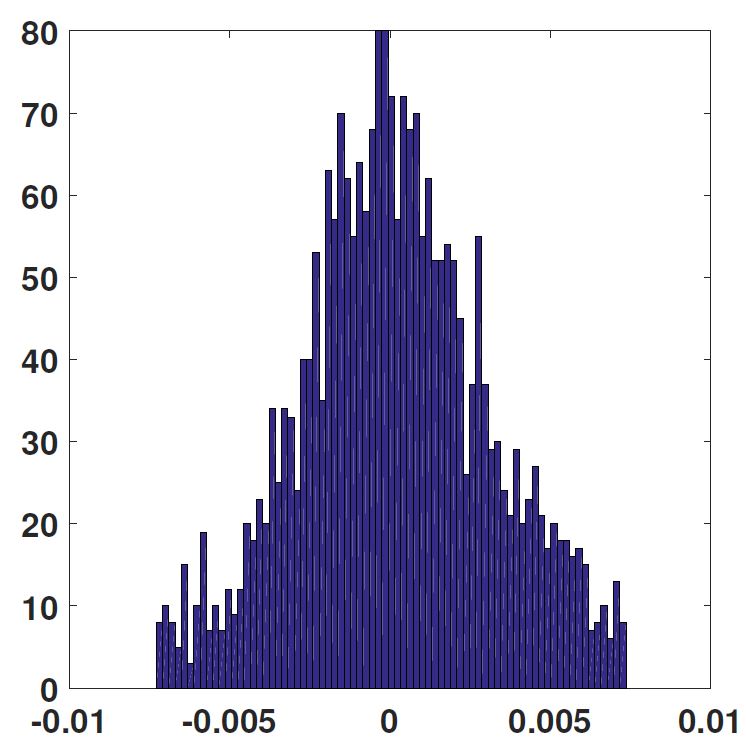}}
			
			\subfigure[]{\includegraphics[width=0.24\linewidth]{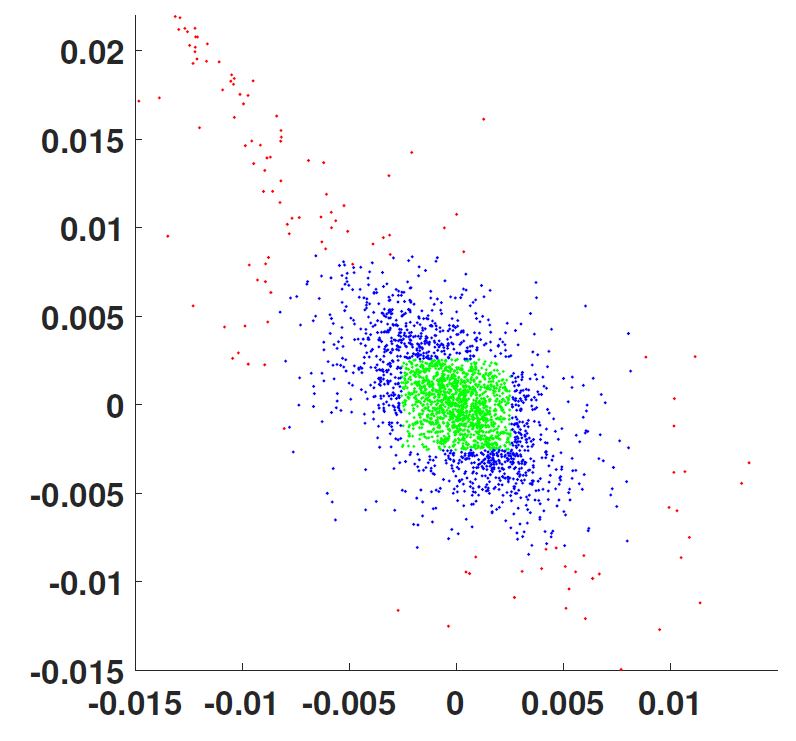}
				\includegraphics[width=0.24\linewidth]{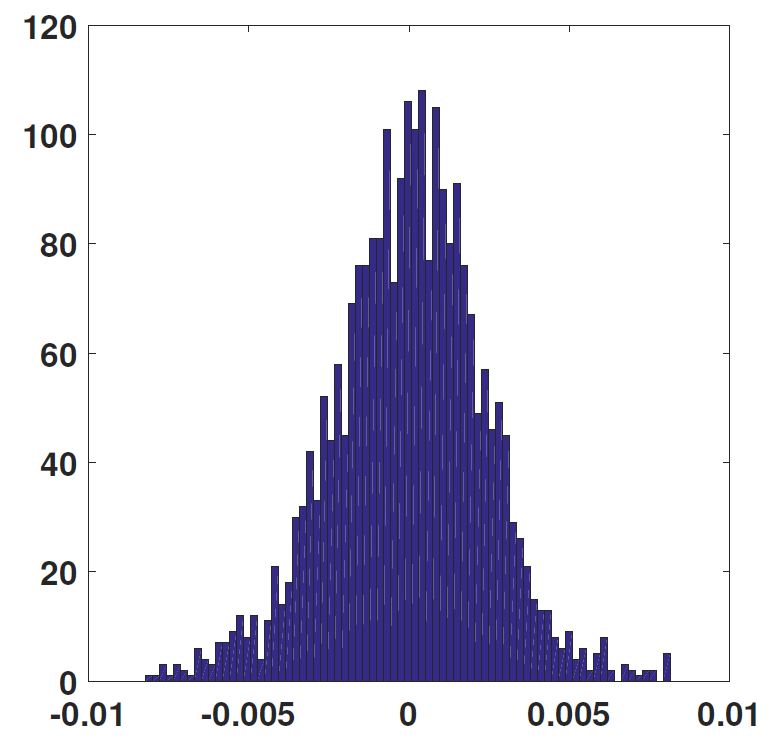}
				\includegraphics[width=0.24\linewidth]{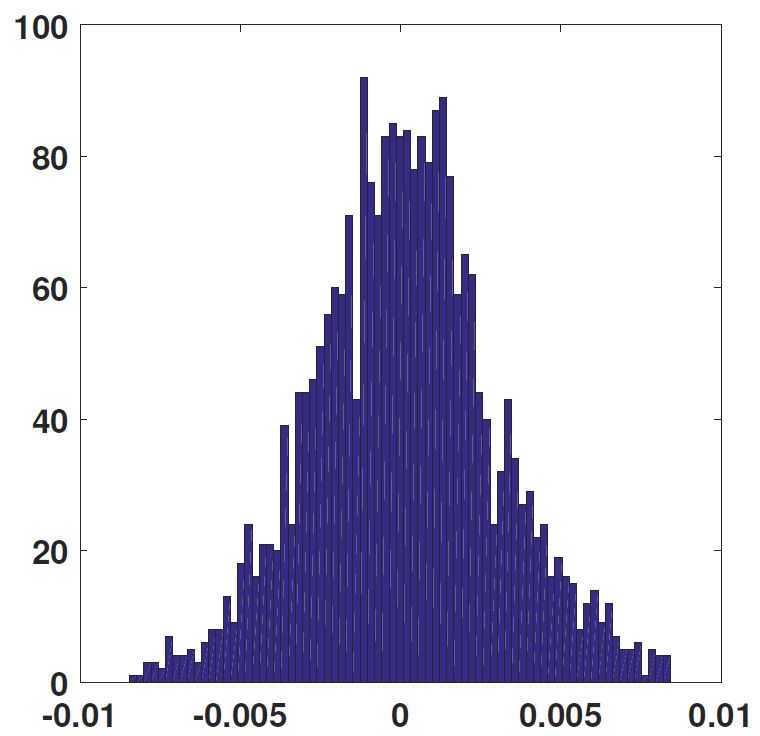}
				\includegraphics[width=0.24\linewidth]{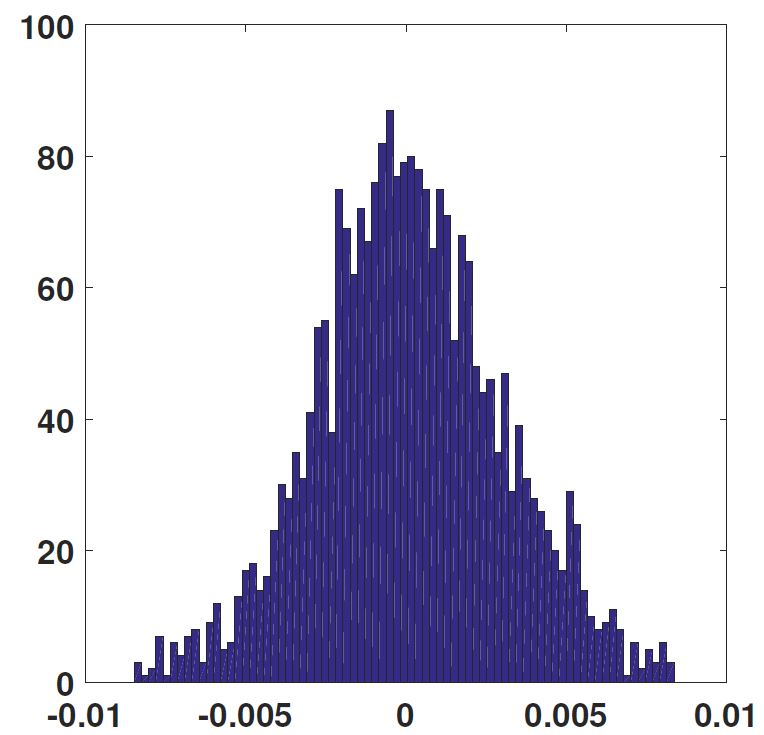}}
			\captionsetup{font=scriptsize}
			\caption{Residual distribution plots (for configuration {\textbf{K2}}) for number of iterations N=1, 16, 18 in case of CMLE \cite{censi} are respectively shown in (a), (b), (c), where points in red indicates the higher order residual terms trimmed as per \cite[Sec. V-C]{censi} while the inliers are colored blue. Plots in (d) display the residual distribution in the case of CIRLS CF at the convergence, where green points indicate those terms whose weights are exactly equal to one, where as points in blue have weights in the range $(\gamma, 1)$. It can be observed that the CIRLS CF automatically eliminates the outliers and yields a residual distribution that is close to Gaussian without any manual intervention or parameter tuning. }\label{fig4}
		\end{figure}
		
		The difficulty of selecting the correct value of N using such a process is depicted through the scatter plots provided in Fig.\ref{fig4}, which shows the residual distributions for different values of N. It can be seen that the residuals for N=16 and N=18, both look Gaussian. On the other hand the ATE values for the two values of N are quite different, as can be seen from Fig.\ref{fig5}. It can also be seen that one cannot simply take a sufficiently large value of N, as excessive removal of outliers leads to performance degradation. In contrast, the proposed methods do not suffer from such an issue as the weights are automatically tuned according to the number of outliers. Fig.\ref{fig4} also includes the residual distribution plots of the proposed CIRLS CF algorithm at convergence. 
		\begin{figure}
			\centering  
			\subfigure[ATE vs N]{\includegraphics[width=0.49\linewidth]{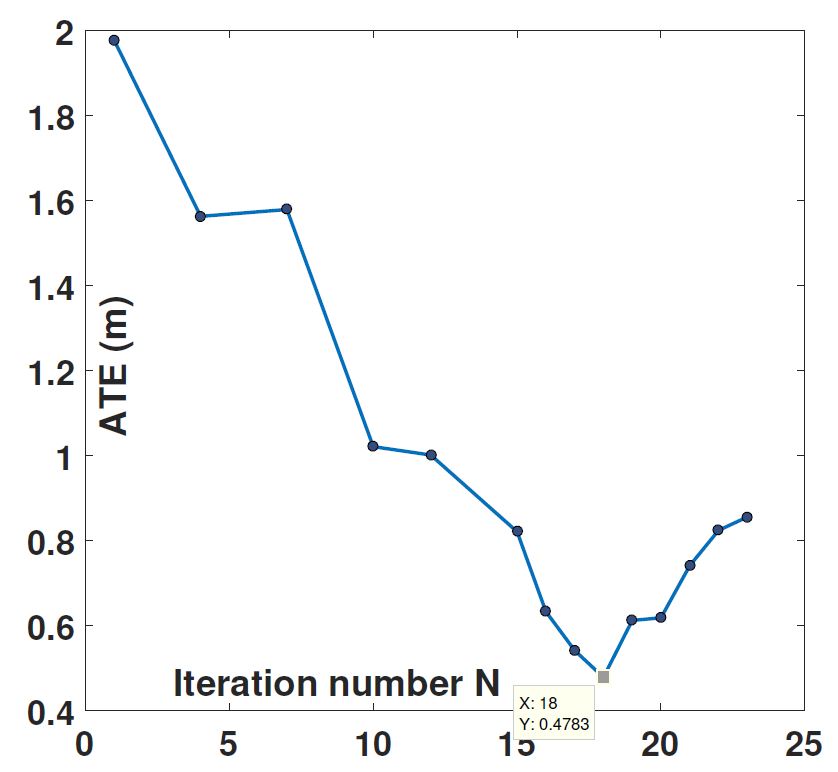}}
			\subfigure[RPE vs N]{\includegraphics[width=0.49\linewidth]{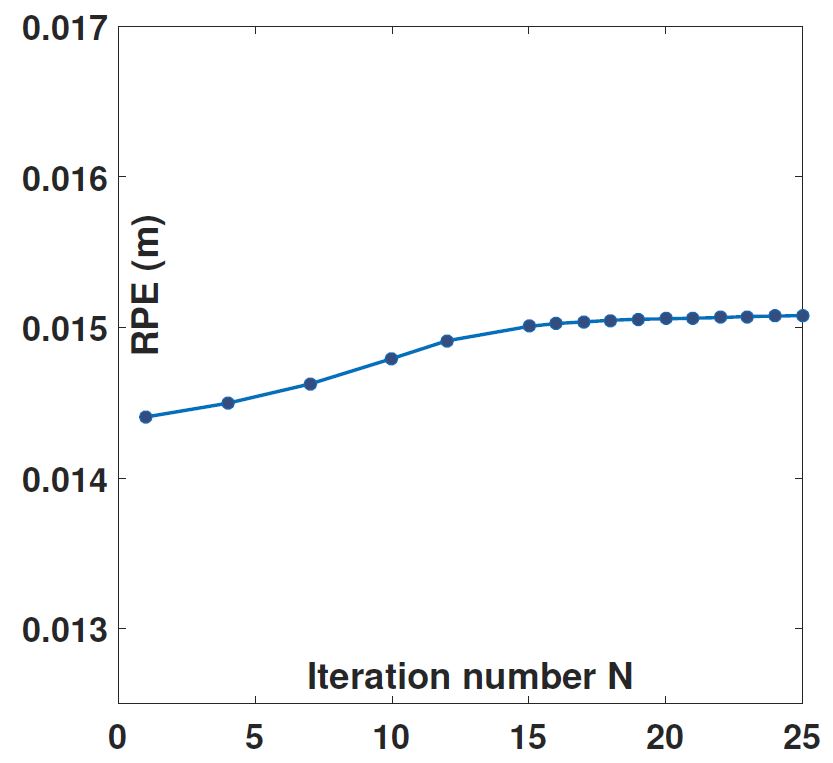}}
			\captionsetup{font=scriptsize}
			\caption{Trajectory error plots for CMLE algorithm \cite{censi} with configuration {\textbf{K2}} under setting {\textbf{D}}, against number of iterations N, while  discarding a fraction of higher order residual terms in each iteration.} \label{fig5}
		\end{figure} 
		
		\begin{figure}
			\centering  
			\subfigure[Helicopter Building, setting \textbf{E}]{\includegraphics[width=0.6\linewidth,trim={0cm 0cm 0cm 0cm},clip]{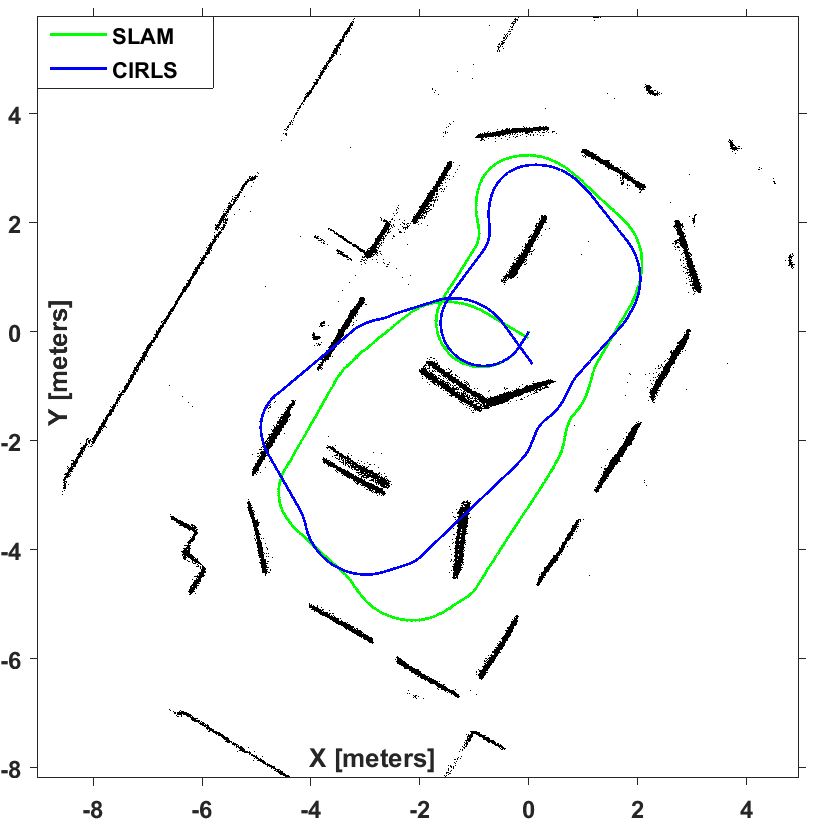}}
			\captionsetup{font=scriptsize}
			\caption{Trajectory comparisons against SLAM in the test environment. Here (a) represents the test environment for configuration \textbf{M1}.}\label{mecanum_}
		\end{figure}

		\begin{table*}
			\centering
			\fontsize{8pt}{10pt}\selectfont
			\setlength{\tabcolsep}{3.0pt}
			\captionsetup{font=scriptsize}
			\caption{List of estimated parameters via CIRLS and corresponding ATE and RPE for configuration \textbf{M1}} \label{table6}
			\begin{tabular}{|c|c|ccc|ccc|cc|}
				\hline
				\multirow{2}{*}{\textbf{\begin{tabular}[c]{@{}c@{}}Robot\\ Config.\end{tabular}}} & \multirow{2}{*}{\textbf{Method}} & \multicolumn{6}{c|}{\textbf{Estimated Parameters}}                                                                                                                                                                                                                                                                                                                                                                                                                         & \multicolumn{1}{c|}{\multirow{2}{*}{\begin{tabular}[c]{@{}c@{}}\textbf{ATE} \\ (cm)\end{tabular}}} & \multirow{2}{*}{\begin{tabular}[c]{@{}c@{}}\textbf{RPE} \\ (mm)\end{tabular}} \\ \cline{3-8}
				&                                  & \multicolumn{1}{c|}{\textit{$\hat{\ell}_x$}(mm)}                            & \multicolumn{1}{c|}{\textit{$\hat{\ell}_y$}(mm)}                            & \textit{$\hat{\ell}_{\theta}$}(rad)                                      & \multicolumn{1}{c|}{\textit{$r$}(mm)}                                     & \multicolumn{1}{c|}{\textit{$L_x$}(mm)}                                   & \textit{$L_y$}(mm)                                                          & \multicolumn{1}{c|}{}                                   &                                    \\ \hline \hline
				\textbf{M1}                                                                       & CIRLS                            & \begin{tabular}[c]{@{}c@{}}-32.60 $\vspace{-1mm}$ $\pm$ 3.50\end{tabular} & \begin{tabular}[c]{@{}c@{}}-25.30 $\vspace{-1mm}$ $\pm$ 3.60\end{tabular} & \begin{tabular}[c]{@{}c@{}}2.14 $\vspace{-1mm}$ $\pm$0.01\end{tabular} & \begin{tabular}[c]{@{}c@{}}30.40 $\vspace{-1mm}$ $\pm$0.20\end{tabular} & \begin{tabular}[c]{@{}c@{}}82.50 $\vspace{-1mm}$ $\pm$1.10\end{tabular} & \begin{tabular}[c]{@{}c@{}}162.50 $\vspace{-1mm}$ $\pm$ 1.10\end{tabular} & 81.90                                                   & 8.38                               \\ \hline
			\end{tabular}
		\end{table*}
		
		\begin{figure*}
			\centering  
			\captionsetup{font=scriptsize}
			\subfigure[Tomography Lab, Configuratoin \textbf{F1}]{\includegraphics[width=0.31\linewidth,trim={0cm 0cm 0cm 0cm},clip ]{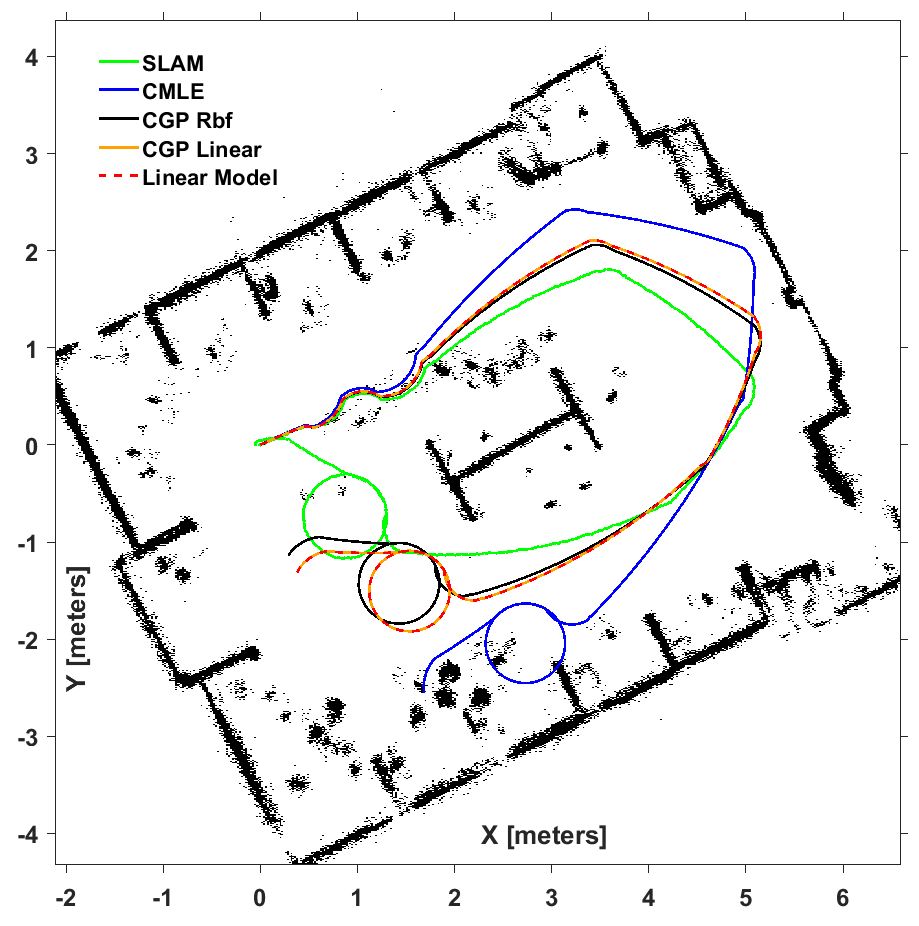}} 
			\subfigure[ACES Library, Configuration \textbf{T1}]{\includegraphics[width=0.31\linewidth,trim={0cm 0cm 0cm 0cm},clip]{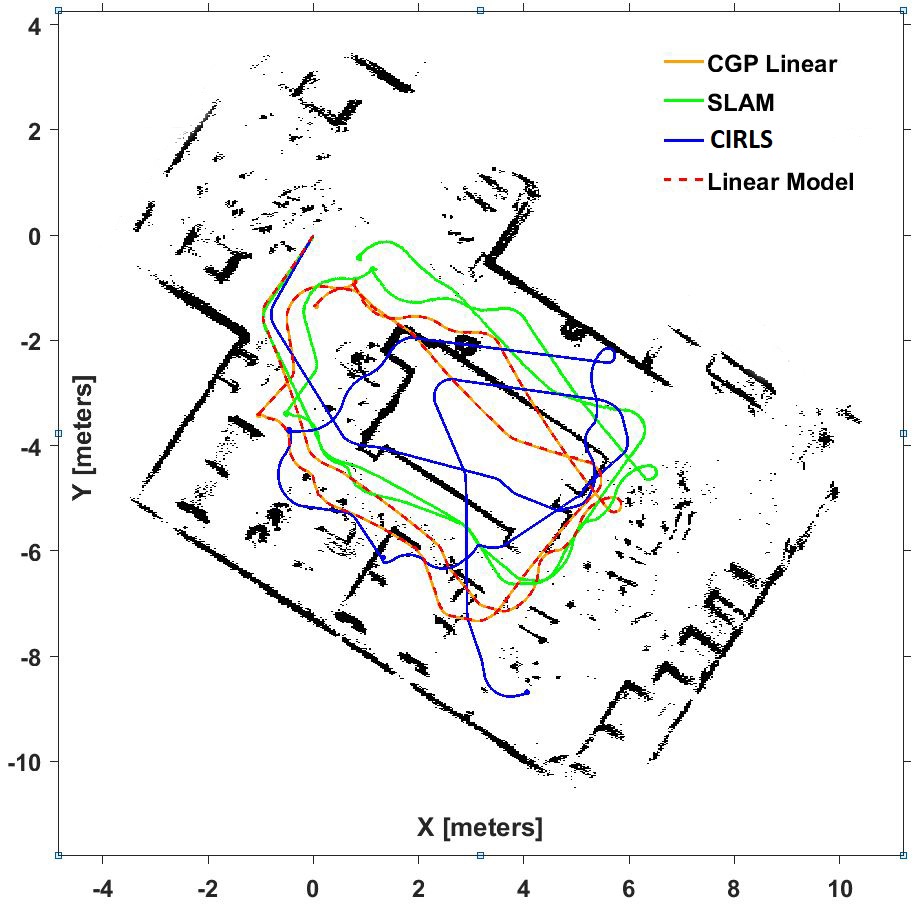}}
			\subfigure[ACES Library, Configuration \textbf{T2}]{\includegraphics[width=0.31\linewidth,trim={0cm 0cm 0cm 0cm},clip]{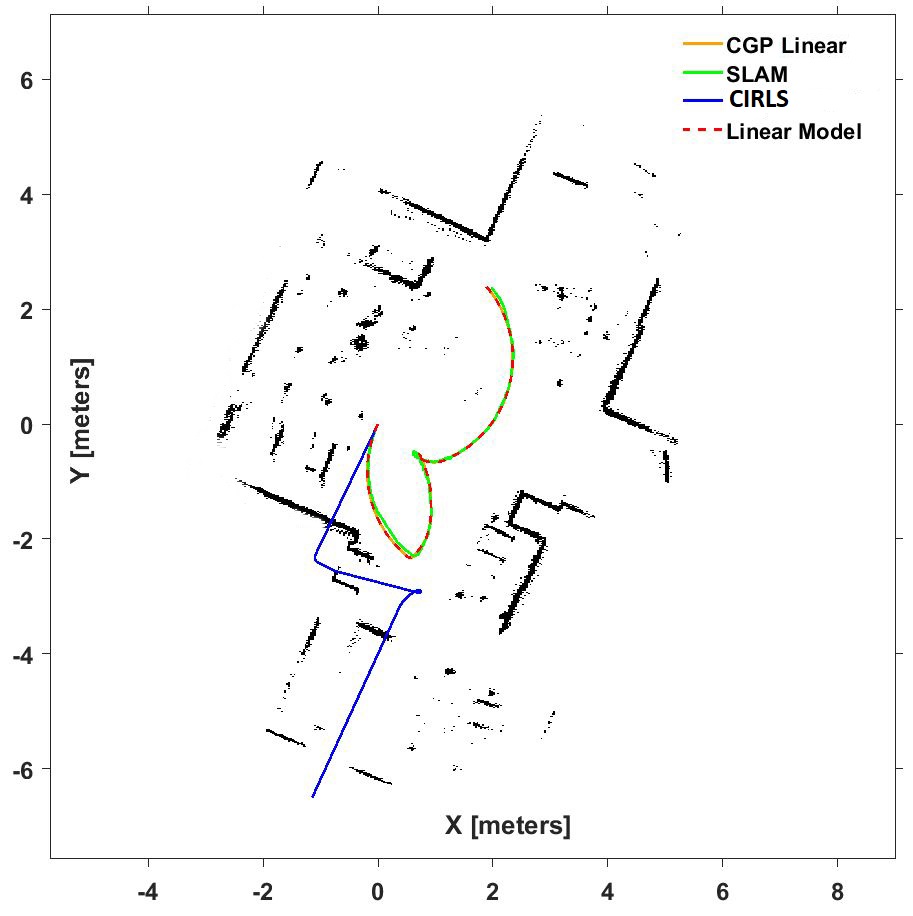}}
			\captionsetup{font=scriptsize}
			\caption{Trajectory comparison against SLAM for different robot configurations. (a) is the test environment for the configuration \textbf{F1} where as (b) and (c) are for configurations \textbf{T1} and \textbf{T2} respectively.
			}
			\vspace{-0.5cm} \label{traj_all} 
		\end{figure*}
		\begin{table*}
			\centering
			\fontsize{8pt}{12pt}\selectfont
			\setlength{\tabcolsep}{3.3pt}
			\captionsetup{font=scriptsize}
			\caption{List of estimated parameters for configuration \textbf{F1} }\label{table5}
			\begin{tabular}{|c|c|c|ccc|ccc|}
				\hline
				\multirow{2}{*}{\textbf{\begin{tabular}[c]{@{}c@{}}Robot\\ Config.\end{tabular}}} & \multirow{2}{*}{\textbf{Method}} & \multirow{2}{*}{\textbf{N}} & \multicolumn{6}{c|}{\textbf{Estimated Parameters}}                                                                                                                                                                                                                                                                                                                                                                                                                         \\ \cline{4-9} 
				&                                  &                             & \multicolumn{1}{c|}{\textit{$\hat{r}_L$ }(mm)}                             & \multicolumn{1}{c|}{\textit{$\hat{r}_R$ }(mm)}                             & \textit{$\hat{b}$ }(mm)                                                    & \multicolumn{1}{c|}{\textit{$\hat{\ell}_x$ }(mm)}                         & \multicolumn{1}{c|}{\textit{$\hat{\ell}_y$ }(mm)}                          & \textit{$\hat{\ell}_{\theta}$}(rad)                                        \\ \hline \hline
				\textbf{F1}                                                                       & CMLE \cite{censi}              & 4                           & \begin{tabular}[c]{@{}c@{}}52.22 $\vspace{-1mm}$ $\pm$ 0.54\end{tabular} & \begin{tabular}[c]{@{}c@{}}50.42 $\vspace{-1mm}$ $\pm$ 0.48\end{tabular} & \begin{tabular}[c]{@{}c@{}}282.67 $\vspace{-1mm}$ $\pm$3.61\end{tabular} & \begin{tabular}[c]{@{}c@{}}90.30 $\vspace{-1mm}$ $\pm$2.75\end{tabular} & \begin{tabular}[c]{@{}c@{}}-38.31 $\vspace{-1mm}$ $\pm$3.28\end{tabular} & \begin{tabular}[c]{@{}c@{}}-0.46 $\vspace{-1mm}$ $\pm$ 0.01\end{tabular} \\ \hline
			\end{tabular}
		\end{table*} 
		\begin{table}
			\centering
			\captionsetup{justification=centering}
			\captionsetup{font=scriptsize}
			\caption{ ATE and RPE for Configurations \textbf{F1}}
			\label{2wheel}
			\begin{tabular}{|c|c|c|c|}
				\hline
				\multicolumn{2}{|c|}{\textbf{GP}}                                       & \multicolumn{2}{c|}{Configuration \textbf{F1}}                                               \\ \hline
				\textbf{Mean fn} & \textbf{Kernel fn} & \multicolumn{1}{c|}{\textbf{ATE} (m)} & \textbf{RPE} (mm) \\ \hline \hline
				Zero                              & RBF                                 & 6.273                                                  & 9.634                              \\ \cline{1-2}
				Linear                            & RBF                                 & \textbf{0.592}                                                  & \textbf{9.367}                              \\ \cline{1-2}
				Zero                              & Linear                              & 0.687                        & 9.367    \\ \cline{1-2}
				Zero                              & RBF + Linear                        & 0.716                                                  & 9.34                               \\ \cline{1-2}
				Linear                            & RBF + Linear                        & 0.732                                                  & 9.343                              \\ \cline{1-2}
				\multicolumn{2}{|c|}{Linear Model}                                      & 0.687                                                  & 9.367                              \\ \cline{1-2}
				\multicolumn{2}{|c|}{CMLE \cite{censi}}                                       & 1.546                                                  & 9.361                              \\ \hline
			\end{tabular}
		\end{table}

		\begin{table}
			\fontsize{10pt}{12pt}\selectfont
			\captionsetup{justification=centering}
			\captionsetup{font=scriptsize}
			\caption{ ATE and RPE for Configurations \textbf{T1} and \textbf{T2}}
			\label{erro_tur}
			\resizebox{\columnwidth}{!}{%
				\begin{tabular}{|c|c|cc|cc|}
					\hline
					\multicolumn{2}{|c|}{\textbf{GP}}         & \multicolumn{2}{c|}{\textbf{Configuration T2}}                & \multicolumn{2}{c|}{\textbf{Configuration T1}}                \\ \hline
					$\textbf{Mean fn}$ & $\textbf{Kernel fn}$ & \multicolumn{1}{c|}{$\textbf{ATE}$ (m)} & $\textbf{RPE}$ (mm) & \multicolumn{1}{c|}{$\textbf{ATE}$ (m)} & $\textbf{RPE}$ (mm) \\ \hline \hline
					Zero               & RBF                  & 2.49                                    & 5.21                & 6.523                                   & 5.466               \\ \cline{1-2}
					Linear             & RBF                  & 0.161                                   & 8.324               & 5.006                                   & 5.573               \\ \cline{1-2}
					Zero               & Linear               & $\textbf{0.068}$                        & $\textbf{5.108}$    & $\textbf{0.87}$                         & $\textbf{5.457}$    \\ \cline{1-2}
					Zero               & RBF + Linear         & 3.798                                   & 5.121               & 0.87                                    & 5.455               \\ \cline{1-2}
					Linear             & RBF + Linear         & 1.193                                   & 5.49                & 1.849                                   & 5.519               \\ \cline{1-2}
					\multicolumn{2}{|c|}{Linear Model}        & 0.068                                   & 5.108               & 0.869                                   & 5.458               \\ \cline{1-2}
					\multicolumn{2}{|c|}{CIRLS}       & 4.24                                    & 5.112               & 3.647                                   & 5.517               \\ \hline
			\end{tabular}}
		\end{table}
		
		To further demonstrate the generality of the proposed CIRLS algorithm, we consider the joint calibration problem for the four-wheeled Mecanum drive robot (see Fig. \ref{meca}) mounted with a lidar sensor under setting \textbf{E} (see Table \ref{table1}). To this end, we perform calibration routine using CIRLS algorithm over configuration \textbf{M1}, and the corresponding estimated calibration parameters are displayed in Table \ref{table6}. Next, we generate the trajectory of the exteroceptive sensor mounted on the Mecanum robot using its inferred (using estimated parameters) motion model $\hat{\f}$ as displayed in Fig. \ref{mecanum_}(a). Observe that the predicted trajectory is close to ground truth and the corresponding ATE, and RPE values are also displayed in Table \ref{table6}. We remark here that, to the best of our knowledge, the proposed algorithm is the first of its kind capable of simultaneously calibrating complicated drives such as the four-wheel Mecanum drive in a robust manner.

		\subsubsection{Results for model-free Calibration}\label{model_free} For the case when the robot kinematic model is not known, ensured by deforming the left wheel of \emph{FireBird VI} (see Fig.\ref{fig_def}), we perform calibration of such a robot using model-based CMLE \cite{censi} algorithm along with the proposed model-free CGP algorithm, for configuration \textbf{F1} under setting \textbf{F}. With regards to the CMLE algorithm, Table \ref{table5} displays the corresponding estimated parameters. Observe that the CMLE algorithm predicts the radius of the left wheel to be slightly more than that of the right wheel, but recall that the left wheel is deformed in such a way that it loses its notion of circularity. Unlike CMLE \cite{censi}, the proposed CGP algorithm conducts a pre-selection phase involving testing over the various kernel and mean functions, using collected data.

		Once the model is learned in both parametric and non-parametric forms, predictions are made on the test data. The predicted trajectories are then compared with SLAM trajectory as the reference. Error metrics for the resulting trajectories are generated and displayed in Table \ref{2wheel}. 
		It is observed that CGP with squared exponential kernel function and linear mean function outperforms other trained models, including CMLE. We remark here that although CMLE predicts the radius of the left wheel to be slightly more than that of the right wheel, the predictions are worse since the original kinematic model is no longer applicable.  Observe that CGP with linear kernel function is comparable to the best case. Next, trajectories are generated for CMLE \cite{censi}, CGP with linear kernel and SLAM \cite{slam} and displayed in Fig. \ref{traj_all}(a). It is also evident that the proposed CGP with linear kernel predicts the test trajectory that is close to SLAM.

		Similar tests are conducted over four-wheeled Turtlebot3 robot with configurations \textbf{T1} and \textbf{T2}. Note that since the analysis of CMLE \cite{censi} is restricted to two-wheel differential drive robots, we use parametric motion model of four-wheel Mecanum drive robot \cite{mecanum} with manufacturer specified parameters for robot intrinsics, and nominal hand measured parameters for lidar extrinsics. Table \ref{erro_tur} displays error metrics evaluated for parametric and various learned non-parametric models. It is also observed here that the proposed CGP algorithm with linear kernel function outperforms other learned models. The corresponding test trajectories for configurations \textbf{T1} and \textbf{T2} are displayed in Fig. \ref{traj_all}(b) and Fig. \ref{traj_all}(c) respectively.

		Interestingly it can be observed from Table.\ref{2wheel} and Table \ref{erro_tur} that the linear model approximation is sufficient to explain the motion model with the set deformities in all configurations. Here we notice that learning a linear approximation of $\f$ is sufficient to predict robot odometry accurately; this is in lines with our discussion in Sec. \ref{linear_model}.

		\section{Conclusion} \label{consec}
		We develop generalized odometry and sensor calibration framework applicable to wheeled robots equipped with one or more exteroceptive sensor(s). The idea is to utilize the ego-motion estimates from the exteroceptive sensors to estimate the motion model of the sensor/robot. Three different algorithms, all capable of handling outliers in an automated manner, are proposed. The first algorithm pertains to a two-wheel differential drive equipped with a Lidar. A calibration via alternating minimization (CAM) approach is proposed that can be used to estimate the intrinsic and extrinsic parameters of the robot robustly. The proposed approach not only has superior performance as compared to the state-of-the-art approaches but also does not require any manual trimming of outliers. A more general scheme applicable to robots with arbitrary drive models is subsequently proposed that utilizes the iteratively reweighted least squares (IRLS) framework for estimation of the motion parameters. The resulting IRLS approach can be used to calibrate arbitrary robots with known motion models and is again shown to provide good calibration performance while handling outliers. Finally, for robots whose motion model is not known or too complicated, we advocate a non-parametric Gaussian process regression-based approach that directly learns the relationship between the wheel odometry and the sensor motion. The model-free calibration approach is tested on a robot with a deformed wheel and is shown to outperform all other techniques that make assumptions regarding the motion model of the robot. Multiple experiments are carried, and the efficacy of the proposed techniques is evaluated using the performance metrics.

		\appendices
		\section{Solutions to \eqref{hminl}-\eqref{hminr}}\label{appA}
		This section will detail the techniques required to solve \eqref{hminl} and \eqref{hminr}. Specifically, the extrinsic calibration problem in \eqref{hminl} will be solved in closed-form while a low-complexity grid search algorithm will be provided to solve \eqref{hminr}.

		\subsubsection{Extrinsic calibration}
		For brevity, let us denote $\mathbf{t}_\l := (\ell_x, \ell_y)^T$ and $\mathbf{t}_{jk} :=(q_{jk}^x,q_{jk}^y)^T$. Given $\r = \r'$, the objective function $h(\r',\l)$ can be written as (see \eqref{fobj}):
		
		\begin{align}\label{eq13}
		h(\r',\l) &= \sum_{(j,k) \in \E} 
		\Bigg( \sum_{i}  \Big\|  \mathbf{R}(\ell_\theta)\hspace{1mm}\bigg(\z_j^{(i)}-\mathbf{R}(q_{jk}^\theta)\ \z_k^{(i)}\bigg) \hspace{1mm} \nonumber\\ 
		&\hspace{1cm}+\bigg(\mathbf{I}_{2\times2}-\mathbf{R}(q_{jk}^\theta)\bigg)\hspace{1mm}\mathbf{t}_{\l} -\mathbf{t}_{jk} \Big \|_2^2 \Bigg) 
		\end{align}
		where $q_{jk}^\theta$ and $\mathbf{t}_{jk}$ are functions of $\r'$. Here, we have used the fact that the 2D rotational matrices commute. Let $\z_{jk}^{(i)}=\z_j^{(i)}-\mathbf{R}(q_{jk}^\theta)\ \z_k^{(i)}$ and $\mathbf{R}_{jk}=\mathbf{I}_{2\times2}-\mathbf{R}(q_{jk}^\theta)$ so that the extrinsic calibration problem may be written as 
		\begin{equation}\label{eq14}
		\l^* = \arg\min_{(\ell_\theta, \mathbf{t}_\l)} h(\r',\ell_\theta, \mathbf{t}_\l)
		\end{equation}
		where 
		\begin{equation}\label{eq15}
		h(\r',\ell_\theta, \mathbf{t}_\l) = \sum_{(j,k) \in \E} \left( \sum_{i}  \left \|  \mathbf{R}(\ell_\theta)\ \z_{jk}^{(i)}
		+ \mathbf{R}_{jk}\ \mathbf{t}_{\l} -\mathbf{t}_{jk} )\right \|_2^2 \right) 
		\end{equation}
		In order to solve \eqref{eq14} in closed-form, we expand \eqref{eq15} and collecting all constant terms into $c$, we obtain :   
		\begin{equation} \label{eq16}
		\begin{split}
		h(\r',\ell_\theta, \mathbf{t}_\l) = \sum_{(j,k) \in \E} & \Big(  2\ \z_{jk}^T\mathbf{R}(\ell_\theta)^T\mathbf{R}_{jk}\mathbf{t}_\l  -2\ \z_{jk}^T\mathbf{R}(\ell_\theta)^T\mathbf{t}_{jk}\\  & - 2\eta\ \mathbf{t}_{jk}^T\mathbf{R}_{jk}\mathbf{t}_\l + \eta\ \mathbf{t}_\l^T\mathbf{R}_{jk}^T\mathbf{R}_{jk}\mathbf{t}_\l \Big) + c 
		\end{split}
		\end{equation}
		where $\z_{jk} = \sum_{i}\z_{jk}^{(i)}$ , $\eta =(\sum_{i}^{\zeta.\left|\mathcal{Z}(t_j)\right|}1) $ denoting total number of scan points in the overlapping region. Now let
		\begin{equation}\label{eq17}
		\begin{split}
		\z_{jk}^T \mathbf{R}(\ell_\theta)^T &=  \begin{pmatrix}
		\z_{jk}^x &  \z_{jk}^y
		\end{pmatrix}\begin{pmatrix}
		cos \ \ell_\theta  & sin \ \ell_\theta\\ 
		-sin \ \ell_\theta & cos \ \ell_\theta 
		\end{pmatrix} \\
		&=   \underbrace{\begin{pmatrix}
			cos \ \ell_\theta  &  sin \ \ell_\theta
			\end{pmatrix}}_{\mathbf{x}^T} \hspace{3mm} \underbrace{\begin{pmatrix}
			\z_{jk}^x & \z_{jk}^y\\ 
			-\z_{jk}^y & \z_{jk}^x
			\end{pmatrix}}_{\mathbf{Z}_{jk}} 
		\end{split}
		\end{equation} 
		Hence using \eqref{eq17} we get the modified new function in terms of $\mathbf{x},\mathbf{t}$ and ignoring the terms independent to the optimization problem, with added constraint  as follows :
		\begin{equation}\label{eq18}
		\begin{split}
		\tilde{h}(\mathbf{x,t_\l}) = \sum_{(j,k) \in \E} & \Big(  2\ \mathbf{x}^T\mathbf{Z}_{jk}\mathbf{R}_{jk}\mathbf{t}_\l  -2\ \mathbf{x}^T\mathbf{Z}_{jk}\mathbf{t}_{jk}\\  & -2\eta\ \mathbf{t}_{jk}^T\mathbf{R}_{jk}\mathbf{t}_\l + \eta\ \mathbf{t}_\l^T\mathbf{R}_{jk}^T\mathbf{R}_{jk}\mathbf{t}_\l \Big)
		\end{split}
		\end{equation}
		By writing \eqref{eq18} in matrix form the problem becomes  
		\begin{equation}\label{eq19}
		\begin{split}
		\min_{\mathbf{x},\mathbf{t}_\l}  & \hspace{2mm}\mathbf{t}_\l^T \mathbf{Q}\mathbf{t}_\l + \mathbf{x}^T \mathbf{M}\mathbf{t}_\l + \mathbf{x}^T\mathbf{g} + \mathbf{t}_\l^T\mathbf{d} \\
		& s.t \hspace{2mm} \mathbf{x}^T\mathbf{x}=1
		\end{split}
		\end{equation}  
		where 
		\begin{equation}\label{eq20}
		\begin{split}
		\mathbf{Q} &= \eta \sum_{(j,k) \in \E}\mathbf{R}_{jk}^T\mathbf{R}_{jk},\  \mathbf{M} = 2\sum_{(j,k) \in \E}\mathbf{Z}_{jk}\mathbf{R}_{jk}  \\\mathbf{g} &= -2\sum_{(j,k) \in \E}\mathbf{Z}_{jk}\mathbf{t}_{jk},\  \mathbf{d}=-2\eta \sum_{(j,k) \in \E}\mathbf{R}_{jk}^T\mathbf{t}_{jk} \\ 
		\end{split}
		\end{equation}
		Since there is no constraint on $\mathbf{t}_\l$, we can solve \eqref{eq19} for $\mathbf{t}_\l$ as shown below,
		\begin{equation}\label{eq21}
		\begin{split}
		\nabla_{\mathbf{t}_\l}& \ \tilde{h}(\mathbf{x,t_\l}) = 0 
		\Rightarrow 2\mathbf{Q}\mathbf{t}_\l + \mathbf{M}^T\mathbf{x}+\mathbf{d}=0 \\
		&\Rightarrow \mathbf{\hat{t}_\l}  = -\frac{1}{2}\mathbf{Q}^\dagger(\mathbf{M}^T\mathbf{x}+\mathbf{d})
		\end{split}
		\end{equation}   
		where $\mathbf{Q}^\dagger$ is the pseudo inverse of $\mathbf{Q}$. Substituting for $\mathbf{t}_\l$ in \eqref{eq19} and ignoring constant terms yields the following in terms of  $\mathbf{x}$ : 
		\begin{equation}\label{eq22}
		\begin{split}
		\min_{\mathbf{x}} &\hspace{2mm}\mathbf{x}^T\tilde{\mathbf{M}}\mathbf{x} + \mathbf{x}^T\tilde{\mathbf{g}}\\
		& s.t \hspace{2mm} \mathbf{x}^T\mathbf{x}=1
		\end{split}
		\end{equation} 
		where $\tilde{\mathbf{M}} = -\frac{1}{4}\hspace{1mm}\mathbf{M}\hspace{1mm}\mathbf{Q}^\dagger\hspace{1mm}\mathbf{M}^T$ and $\tilde{\mathbf{g}} =\mathbf{g} - \frac{1}{2}\hspace{1mm}\mathbf{M}\hspace{1mm}\mathbf{Q}^\dagger\hspace{1mm}\mathbf{d}$.
		\par We solve the above problem using method of Lagrange multipliers. Specifically, the Lagrangian is given by
		\begin{equation}\label{eq23}
		\mathcal{L}(\mathbf{x}) = \mathbf{x}^T \tilde{\mathbf{M}}\mathbf{x} + \mathbf{x}^T\tilde{\mathbf{g}} + \lambda (\mathbf{x}^T\mathbf{x}-1) 
		\end{equation}   
		The necessary condition for optimality is $\nabla_{\mathbf{x}}\mathcal{L}(\mathbf{x}) =\mathbf{0}$ i.e.,
		\begin{equation}\label{eq24}
		\begin{split}
		& \hspace{4mm}2\tilde{\mathbf{M}}\mathbf{x} + \tilde{\mathbf{g}}+2\lambda\mathbf{x}=\mathbf{0}\\
		&\Rightarrow \mathbf{\hat{x}} = -\frac{1}{2}(\tilde{\mathbf{M}} + \lambda \mathbf{I}_{2\times2})^{-1}\tilde{\mathbf{g}}
		\end{split}
		\end{equation}    
		To solve for $\lambda$ substitute $\mathbf{\hat{x}}$ in the constraint $\mathbf{x}^T\mathbf{x}=1$, which results in a fourth order polynomial in $\lambda$, whose roots can be readily found. Having determined $\lambda$, solutions to \eqref{eq19}, can be found subsequently in closed forms and therefore $\l^*$ can be recovered as follows,
		
		\begin{equation}\label{eq25}
		\ell_\theta^* = \arctan \big(\mathbf{\hat{x}}(2),\mathbf{\hat{x}}(1) \big),  \hspace{5mm} (\ell_x^*,\ell_y^*)^T = \mathbf{\hat{t}}_\l 
		\end{equation}
		Next, we detail the 
		\textit{{proof of existence of pseudo inverse of \textbf{Q}}}.

		As $\mathbf{Q} = \eta \sum_{(j,k) \in
			\E}\mathbf{R}_{jk}^T\mathbf{R}_{jk}$, let us consider one term
		$(j,k)^{th}$ of the summation.
		
		\begin{equation}
		\begin{aligned}
		\mathbf{R}_{jk}^T\mathbf{R}_{jk} &=
		(\mathbf{I}-\mathbf{R}(q_{jk}^\theta))^T
		(\mathbf{I}-\mathbf{R}(q_{jk}^\theta)) \\
		&= 2\mathbf{I} - \mathbf{R}(q_{jk}^\theta)^T - \mathbf{R}(q_{jk}^\theta)
		\\
		&= 2
		\begin{bmatrix}
		1 - \cos(q_{jk}^\theta) & 0 \\
		0 & 1 - \cos(q_{jk}^\theta)
		\end{bmatrix} \succcurlyeq 0
		\end{aligned}
		\end{equation}
		
		Since, $\eta > 0$ as it is the total number of scan points in the
		overlapping region. Therefore, $\mathbf{Q}$ can be written as a sum of
		positive semidefinite matrices multiplied by a non-negative intiger,
		hence positive semidefinite. Pseudo inverse will exist if any one of the
		matrix in the sum is positive definite. This will happen if for any of
		the terms $q_{jk}^\theta \neq 0$, which holds for exciting trajectories
		not having all pure translations.
		Therefore pseudo inverse of $\mathbf{Q}$ i.e. $\mathbf{Q}^\dagger$
		always exist for sufficiently exciting trajectories.

		\subsubsection{Intrinsic calibration}\label{intr} Here the goal is to solve for intrinsic parameters in closed forms, given extrinsic parameters.  To this end, expanding \eqref{fobj} given $\l = \l'$, we obtain 
		\begin{equation}\label{eq26}
		\begin{split}
		h(\r,\l') = \sum_{(j,k) \in \E} &
		\Bigg( \sum_{i}  \Big\| \tilde{\z}_j^{(i)}  -\bigg( \mathbf{R}(q_{jk}^\theta)\tilde{\z}_k^{(i)} + \mathbf{t}_{jk} \bigg) \hspace{1mm} \Big\|_2^2 \Bigg) 
		\end{split}
		\end{equation}
		where $\tilde{\z}_j^{(i)} = \l' \oplus \z_j^{(i)} $ and $\tilde{\z}_k^{(i)} = \l' \oplus \z_k^{(i)} $. The intrinsic calibration problem is therefore posed as 
		\begin{equation}\label{eq27}
		\mathbf{r}^* = \arg \min_{\mathbf{r}} h(\r,\l')
		\end{equation} 
		To solve the above problem we first find an equivalent problem in terms of new set of variables $\tilde{\r} = (\tilde{\r}_L,\tilde{\r}_R,b)$, where $\tilde{\r}_L = -\mathbf{J}_\r^{21}$, $\tilde{\r}_R = \mathbf{J}_\r^{22}$ and remember the definition of $\mathbf{J}_\r$ from equation \eqref{eq3}. Rewriting \eqref{qjk2w} with this new set of variables, we found that $\mathbf{t}_{jk} = b \ \mathbf{\tilde{t}}_{jk} $ where $\mathbf{\tilde{t}_{jk}}$ is a vector function parameterized by  $\tilde{\r}_L $ and $\tilde{\r}_R $. Therefore equation \eqref{eq26} becomes 
		\begin{equation}\label{eq26_}
		\begin{split}
		h(\tilde{\r},\l') = \sum_{(j,k) \in \E} &
		\Bigg( \sum_{i}  \Big\| \tilde{\z}_j^{(i)}  -\bigg( \mathbf{R}(q_{jk}^\theta)\tilde{\z}_k^{(i)} + b\ \mathbf{\tilde{t}}_{jk} \bigg) \hspace{1mm} \Big\|_2^2 \Bigg) 
		\end{split}
		\end{equation}
		Note here that $\q_{jk}^\theta$ is a scalar function parameterized by $\tilde{\r}_L $ and $\tilde{\r}_R $. Hence one can solve \eqref{eq26_} for $b$ in closed form and is given by 
		\begin{equation}\label{eqb_48}
		\hat{b} = \frac{\sum_{(j,k) \in \E} \sum_i \mathbf{\tilde{t}}_{jk}^T R(\q_{jk}^\theta) \tilde{\z}_k^{(i)} - \mathbf{\tilde{t}}_{jk}^T \tilde{\z}_j^{(i)}} {\eta \sum_{(j,k) \in \E}\|\mathbf{\tilde{t}}_{jk} \|_2^2}
		\end{equation}   
		where $\eta = \sum_{i}1$. Now substituting for $b$ in \eqref{eq26} using closed form \eqref{eqb_48} yields an objective function $\tilde{h}(\tilde{r}_L, \tilde{r}_R)$. Since approximate values of $\tilde{r}_L$ and $\tilde{r}_R$ are known from the original robot specifications or through hand-held measurements, a simple grid search around these values yields the optimum values that minimize $\tilde{h}(\tilde{r}_L, \tilde{r}_R)$. Alternatively, any two-dimensional search algorithm may be utilized \cite{grid}. The complexity of such a search is low since a good initialization is always available. 
		
		To handle outliers due to non-systematic errors such as wheel slippages, the intrinsic calibration problem defined in \eqref{eq27} can be solved by incorporating Huber loss instead of squared loss. However, though closed forms for the parameter $b$ does not exists, the problem can be solved by employing numerical techniques detailed in \cite{numer}.

		\section{  Autonomous Calibration using PLICP Framework }\label{appB}
		PLICP uses point-to-line metric \cite{P2LICP} instead of point-to-point used by ICP. Proceeding in a manner similar to that in the PLICP algorithm \cite{P2LICP}, we formulate the objective function for simultaneous calibration using PLICP framework as follows : 
		\begin{equation} \label{apeq34}
		h(\l,\r) = \sum_{(j,k) \in \E} \sum_{i}\bigg( \boldsymbol{\eta}_{jk}^{(i)T} \Big[ \l \oplus \z_{j}^{(i)} - \left (\mathbf{q}_{jk} \oplus \l\oplus \z_k^{(i)}  \right ) \Big] \bigg)^2 
		\end{equation}    
		where $\boldsymbol{\eta}_{jk}^{(i)}$ is the normal vector. In a very compact form \eqref{apeq34} can be written as : 
		\begin{equation}\label{apeq35}
		h(\l,\r) = \sum_{(j,k) \in \E} \left( \sum_{i}  \left \| \l\oplus \z_j^{(i)} - \left (\mathbf{q}_{jk} \oplus \l\oplus \z_k^{(i)}  \right ) \right \|_{\mathbf{C}_{jk}^{(i)}}^2 \right) 
		\end{equation}   
		where $\mathbf{C}_{jk}^{(i)} =  \boldsymbol{\eta}_{jk}^{(i)} \boldsymbol{\eta}_{jk}^{(i)T} $.
		\subsubsection{Extrinsic calibration}
		Following the similar analysis done from \eqref{eq13} - \eqref{eq15}, here in this context we have the following subproblem 
		\begin{equation}\label{apeq36}
		\min_{(\ell_\theta,\mathbf{t}_\l)} \sum_{(j,k) \in \E}\left( \sum_{i}  \left \|  \mathbf{R}(\ell_\theta)\ \z_{jk}^{(i)}
		+ \mathbf{R}_{jk}\ \mathbf{t_\l} -\mathbf{t}_{jk} )\right \|_{\mathbf{C}_{jk}^{(i)}}^2  \right) 
		\end{equation}    
		By reducing \eqref{apeq36} to quadratic form in four dimensional space $\mathbf{x}=[x_1,x_2,x_3,x_4] \overset{\Delta}{=} [\ell_x,\ell_y,cos\ \ell_\theta, sin\ \ell_\theta]  $ and imposing additional constraint $x_3^2+x_4^2=1$, a closed form can be found as detailed in \cite{P2LICP}, i.e., by writing \eqref{apeq36} as follows,   
		\begin{equation}
		\begin{split}
		\min_{\mathbf{x}} &\hspace{2mm} \mathbf{x}^T\mathbf{M}\mathbf{x} + \mathbf{g}^T\mathbf{x}\\
		& s.t \hspace{2mm} \mathbf{x}^T\mathbf{W}\mathbf{x}=1
		\end{split}
		\end{equation}   
		
		where 
		\begin{equation}
		\begin{split}
		\mathbf{M} &= \sum_{(j,k) \in \E} \sum_{i} \mathbf{M}_{jk}^{(i)T}\mathbf{C}_{jk}^{(i)}\mathbf{M}_{jk}^{(i)} \\
		\mathbf{g} &= \sum_{(j,k) \in \E} \sum_{i}-2\mathbf{M}_{jk}^{(i)T}\mathbf{C}_{jk}^{(i)}\mathbf{t}_{jk}\\
		\mathbf{W} &=  \begin{pmatrix}
		\mathbf{0}_{2\times2} &\mathbf{0}_{2\times2} \\ 
		\mathbf{0}_{2\times2} & \mathbf{I}_{2\times2} 
		\end{pmatrix} 
		\end{split}
		\end{equation} and $\mathbf{M}_{jk}^{(i)} = [\mathbf{R}_{jk}\ \ \mathbf{Z}_{jk}^{(i)}]_{2\times4}$, \hspace{3mm}
		$ \mathbf{Z}_{jk}^{(i)}=\begin{pmatrix}
		\z_{jk}^{(i)x} &-\z_{jk}^{(i)y} \\ 
		\z_{jk}^{(i)y}&\z_{jk}^{(i)x} 
		\end{pmatrix}$.
		\subsubsection{Inrinsic Calibration }
		To find wheel odometry parameters in this framework, we have to solve the following subproblem, 
		\begin{equation}
		\begin{split}
		\min_{\r}\sum_{(j,k) \in \E} &
		\Bigg( \sum_{i}  \Big\| \tilde{\z}_j^{(i)}  -\bigg( \mathbf{R}(q_{jk}^\theta)\tilde{\z}_k^{(i)} + \mathbf{t}_{jk} \bigg) \hspace{1mm} \Big\|_{\mathbf{C}_{jk}^{(i)}}^2 \Bigg) 
		\end{split}
		\end{equation} 
		A similar procedure can be carried out as detailed in Appendix A-(2), to solve the aforementioned subproblem.
		
		\section{Proof of Proposition 1}\label{appC}  
		It can be seen from \eqref{fobj} that for pure rotation (i.e. when  $q_{jk}^{x} = q_{jk}^y = 0$), the function $h$ does not depend on intrinsic parameters $r_L$ and $r_R$. In other words,  $r_L$ and $r_R$ are not observable if all the scan pairs in $\E$ correspond only to pure rotations. Likewise, for pure translations (i.e. when  $q_{jk}^{\theta} = 0$), the function $h$ becomes independent of the extrinsic parameters $\ell_x$ and $\ell_y$, making them unobservable. In summary, at least one scan pair in $\E$ must correspond to translation and one to rotation.


		\footnotesize
		\bibliographystyle{IEEEtran}
		\bibliography{refer}

	\end{document}